%% file: arxiv.tex
\documentclass[11pt]{article}

\usepackage[final]{emnlp_2026}

\usepackage{times}
\usepackage{latexsym}

\usepackage[T1]{fontenc}

\usepackage[utf8]{inputenc}

\usepackage{microtype}

\usepackage{inconsolata}

\usepackage{graphicx}

\usepackage{algorithm}
\usepackage{algpseudocode}
\usepackage{adjustbox}
\usepackage{amsfonts}
\usepackage{amsmath}
\usepackage{amssymb}
\usepackage{amsthm}
\usepackage{array}
\usepackage{bm}
\usepackage{booktabs}
\usepackage{calc}
\usepackage{calculator}
\usepackage{colortbl}
\usepackage{enumitem}
\usepackage{etoc}
\usepackage{float}
\usepackage{fp}
\usepackage{framed}
\usepackage{graphicx}
\usepackage{hhline}
\usepackage{hyperref}
\usepackage{inconsolata}
\usepackage{latexsym}
\usepackage{listings}
\usepackage{makecell}
\usepackage{mathtools}
\usepackage{mdframed}
\usepackage{microtype}
\usepackage{multicol}
\usepackage{multirow}
\usepackage{pgf}
\usepackage{setspace}
\usepackage{subcaption}
\usepackage{tabularx}
\usepackage{tikz}
\usepackage{times}
\usepackage{titletoc}
\usepackage{wrapfig}
\usepackage{xcolor}
\usepackage{xspace}

\definecolor{mycitecolor}{HTML}{65A3FF}
\hypersetup{
    colorlinks=true,      
    linkcolor=mycitecolor,      
    citecolor=mycitecolor, 
    urlcolor=mycitecolor         
}

\usepackage[most]{tcolorbox}

\definecolor{blockbg}{HTML}{EDF6ED}
\definecolor{blockedge}{HTML}{8AC890}

\newenvironment{promptbox}[1]{%
  \def\boxtitle{#1}%
  \colorbox{blockedge}{%
    \parbox{\dimexpr\textwidth-2\fboxsep}{%
      \centering%
      \rule{0pt}{12pt}
      \bfseries\boxtitle%
      \rule[-16pt]{0pt}{12pt}
    }%
  }%
  \vspace{-\baselineskip}%
  \begin{mdframed}[
    backgroundcolor=blockbg,
    linecolor=blockedge,
    linewidth=1pt,
    roundcorner=3pt,
    innerleftmargin=6pt,
    innerrightmargin=6pt,
    innertopmargin=6pt,
    innerbottommargin=6pt,
    skipabove=0pt
  ]
}{%
  \end{mdframed}
}

\definecolor{modelbg}{HTML}{EDF5FE}
\definecolor{modeledge}{HTML}{82B5F4}

\newenvironment{agentbox}[1]{%
  \def\boxtitle{#1}%
  \colorbox{modeledge}{%
    \parbox{\dimexpr\textwidth-2\fboxsep}{%
      \centering%
      \rule{0pt}{12pt}
      \bfseries\boxtitle%
      \rule[-16pt]{0pt}{12pt}
    }%
  }%
  \vspace{-\baselineskip}%
  \begin{mdframed}[
    backgroundcolor=modelbg,
    linecolor=modeledge,
    linewidth=1pt,
    roundcorner=3pt,
    innerleftmargin=6pt,
    innerrightmargin=6pt,
    innertopmargin=6pt,
    innerbottommargin=6pt,
    skipabove=0pt
  ]
}{%
  \end{mdframed}
}

\definecolor{ERRORBG}{HTML}{FDB0B0}

\definecolor{examplebg}{HTML}{F5F5F5}
\definecolor{exampleedge}{HTML}{BBBBBB}

\newcommand{\ours}{\textbf{\textcolor[HTML]{548235}{\textsc{Anagent}}}\xspace}

\newcommand{\ourbench}{\textbf{\textcolor[HTML]{BC6666}{\textsc{AnaBench}}}\xspace}

\newcommand{\logo}{%
  \includegraphics[height=2.0ex]{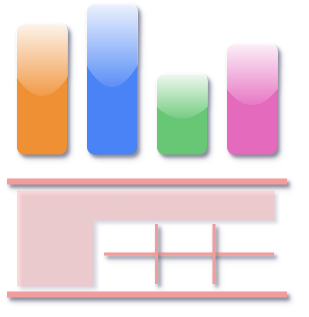}%
  \xspace%
}

\newcommand{\code}[1]{\texttt{#1}}

\definecolor{PlannerColor}{HTML}{EA9C20}
\definecolor{ExpertColor}{HTML}{578ED9}
\definecolor{SolverColor}{HTML}{75B675}
\definecolor{CriticColor}{HTML}{EC83CD}
\newcommand{\Planner}{\textcolor{PlannerColor}{\textbf{\textsc{Planner}}}\xspace}
\newcommand{\Expert}{\textcolor{ExpertColor}{\textbf{\textsc{Expert}}}\xspace}
\newcommand{\Solver}{\textcolor{SolverColor}{\textbf{\textsc{Solver}}}\xspace}
\newcommand{\Critic}{\textcolor{CriticColor}{\textbf{\textsc{Critic}}}\xspace}

\definecolor{CHECK_BG}{HTML}{D3F5D6}

\definecolor{GreenCheck}{HTML}{6AC566}
\definecolor{RedCross}{HTML}{EA7A7A}
\newcommand{\cmark}{\textcolor{GreenCheck}{\scalebox{1.8}{$\checkmark$}}}
\newcommand{\xmark}{\textcolor{RedCross}{\scalebox{1.8}{$\times$}}}

\newif\ifcomments
\commentstrue   

\definecolor{xuehangcolor}{HTML}{FCFFA9}

\definecolor{qingyuncolor}{HTML}{BFFFBF}

\definecolor{tomcolor}{HTML}{A9FFF8}

\newcommand{\blfootnote}[1]{%
  \begingroup
  \renewcommand\thefootnote{}\footnote{#1}%
  \addtocounter{footnote}{-1}%
  \endgroup
}

\author{
  Xuehang Guo\textsuperscript{1} \quad
  Zhiyong Lu\textsuperscript{2} \quad
  Tom Hope\textsuperscript{3} \quad
  Qingyun Wang\textsuperscript{1} \\
  \normalfont\normalsize
  \textsuperscript{1}William \& Mary \quad
  \textsuperscript{2}National Library of Medicine \quad
  \textsuperscript{3}The Allen Institute for AI
}

\title{\ours For Enhancing Scientific Table \& Figure Analysis}

\begin{document}

\maketitle

\blfootnote{%
  Correspondence to:
  Xuehang Guo \texttt{<xguo15@wm.edu>},
  Zhiyong Lu \texttt{<zhiyong.lu@nih.gov>},
  Tom Hope \texttt{<tomh@allenai.org>},
  Qingyun Wang \texttt{<qwang16@wm.edu>}.
}

\begin{abstract}

In scientific research, analysis requires accurately interpreting complex multimodal knowledge, integrating evidence from different sources, and drawing inferences grounded in domain-specific knowledge. However, current artificial intelligence (AI) systems struggle to consistently demonstrate such capabilities. The complexity and variability of scientific tables and figures, combined with heterogeneous structures and long-context requirements, pose fundamental obstacles to scientific table \& figure analysis.
To quantify these challenges, we introduce \ourbench, a large-scale benchmark featuring $63,178$ instances from nine scientific domains, systematically categorized along seven complexity dimensions.
To tackle these challenges, we propose \ours, a multi-agent framework for enhanced scientific table \& figure analysis through four specialized agents: \Planner decomposes tasks into actionable subtasks, \Expert retrieves task-specific information through targeted tool execution, \Solver synthesizes information to generate coherent analysis, and \Critic performs iterative refinement through five-dimensional quality assessment.
We further develop modular training strategies that leverage supervised finetuning and specialized reinforcement learning to optimize individual capabilities while maintaining effective collaboration. Comprehensive evaluation across 170 subdomains demonstrates that \ours achieves substantial improvements, up to $\uparrow 13.43\%$ in training-free settings and $\uparrow 42.12\%$ with finetuning, while revealing that task-oriented reasoning and context-aware problem-solving are essential for high-quality scientific table \& figure analysis. Our project page: \href{https://xhguo7.github.io/Anagent/}{https://xhguo7.github.io/Anagent/}.

\end{abstract}

\input{sections/1_intro}

\input{sections/2_anabench}

\input{tables/exp_main1_training_free}

\input{sections/3_anagent}

\input{tables/exp_main2_finetuned}

\input{sections/4_exps}
\input{sections/5_results}

\input{sections/7_conclusion}

\input{sections/acknowledge}

\bibliography{custom}

\newpage
\appendix
\onecolumn

\input{sections/appendix}

\end{document}

%% file: sections/1_intro.tex
\section{Introduction}
\label{sec:intro}

AI has made notable progress in assisting scientifists across diverse domains~\cite{Boiko2023AutonomousCRA,Gao2024EmpoweringBDA,wang2026apodex} and stages of the research lifecycle, such as hypothesis discovery~\cite{wang-etal-2024-scimon,garikaparthi_iris_2025}, literature review~\cite{Zhang2024FromRTA}, citation recommendation~\cite{Choi2025CiteGuardFCA,Press2024CiteMECLA}, etc.
With the growing trend of \textit{human-AI co-discovery} \cite{Gottweis2025TowardsAI}, these advances reveal AI's potential in serving as AI co-scientists to accelerate scientific discovery and improve research communication~\cite{Gridach2025AgenticAFA,zhang_survey_2024}.
%
%
%
%
%
%
However, to function effectively as AI co-scientists, AI systems draw on capabilities in \textit{multimodal reasoning}~\cite{bai_table_reasoning_2025,zhao_chartcoder_2025}, \textit{long-context comprehension}~\cite{Reddy2024TowardsSDA,Sundar2024cPAPERSADA}, and \textit{domain-specific understanding}, which remain challenging for current AI systems~\cite{Zhou2025FromHTA}.

A fundamental yet important task that reflects these capabilities is \textbf{scientific table \& figure analysis}, as tables and figures provide critical information that is often difficult to express through text alone in scientific papers.
Analyzing these artifacts requires AI systems to accurately: (1) interpret complex multimodal data across diverse layouts and formats (\textit{e.g.}, \texttt{LaTeX} tables, bar charts, architectural diagrams), (2) integrate evidence from multiple sources and lengthy contexts (\textit{e.g.}, captions, sections, citations), and (3) generate task-oriented insights grounded in specialized terminology, related contexts, and domain-specific knowledge. Despite recent advances in multimodal large language models (MLLMs), scientific table \& figure analysis remains challenging, particularly when handling the \textbf{\textit{heterogeneity}} of scientific literature across different \textit{authoring formats} (\textit{e.g.}, \texttt{LaTeX}, \texttt{XML}), \textit{rendered formats} (\textit{e.g.}, PDF, HTML), and \textit{dissemination platforms} (\textit{e.g.}, arXiv~\cite{arxiv}, PubMed~\cite{pubmed}).
This is further complicated by \textit{error propagation}~\cite{Gridach2025AgenticAFA}, as mistakes in structural parsing, numerical extraction, or contextual interpretation can cascade into factual incorrectness.

\textbf{Where existing benchmarks fall short?}
Several benchmarks have been proposed for scientific table \& figure understanding \cite{li2024m3sciqa,singh-etal-2024-scidqa,lu-etal-2023-scitab,zhang2025scitat,pramanick2024spiqa,lou2023s2abel,jin2019pubmedqa,liu2026wildsci}. However, those benchmarks primarily focus on narrowly defined tasks, such as \textit{question answering}, \textit{claim verification}, or \textit{caption generation}. As such, they fail to capture the full spectrum of challenges inherent in scientific analysis writing (Tab.~\ref{tab:benchmark_comparison}), including varying levels of analytical depth,
diverse reasoning requirements across scientific domains, 
and synthesis of information across multiple modalities and long contexts (Fig.~\ref{fig:preliminary}). Moreover, our preliminary exploration reveals that current MLLMs struggle significantly with scientific analysis (\S\ref{subsec:anabench:motivations}).
These limitations are particularly pronounced for scientific analysis tasks requiring complex reasoning across different scopes, depths, and objectives (Fig.~\ref{fig:preliminary-error-analysis}).


\textbf{Our Approach.}
To tackle these challenges, we introduce \ourbench, a scientific table \& figure analysis benchmark encompassing tables and figures from 9 scientific domains across 170 fine-grained disciplines, systematically categorized along seven complexity dimensions that capture multifaceted challenges of scientific analysis (Fig.~\ref{fig:preliminary}).
Building on insights from how human researchers approach scientific writing (Fig.~\ref{fig:teaser}), we propose \ours (Fig.~\ref{fig:anagent}), a multi-agent framework that decomposes scientific analysis into specialized subtasks handled by four collaborative agents: \Planner for \textit{task decomposition and planning}, \Expert for \textit{knowledge searching and retrieval}, \Solver for \textit{reasoning and generation}, and \Critic for \textit{reflection and refinement}.
To enhance agent-wise performance on their specialized tasks, we implement test-time optimization (\S\ref{subsec:anagent:infer}) and modular finetuning (\S\ref{subsec:anagent:train}) to enhance individual agent capabilities while maintaining effective collaboration.
Our main contributions are:

\begin{itemize}[leftmargin=*, topsep=0pt, itemsep=1pt, parsep=1pt]
    \item We introduce \ourbench (\S\ref{sec:anabench}), a benchmark consisting of $63,178$ instances for evaluating and advancing AI systems in scientific tale \& figure analysis, spanning seven data and analysis complexities (Fig.~\ref{fig:preliminary}).
    
    \item We propose \ours (\S\ref{sec:anagent}), a multi-agent framework for scientific table \& figure analysis writing, comprising four specialized agents equipped with specialized tools, enabling complex reasoning, systematic knowledge integration, and collaborative scientific analysis writing.
    
    \item By presenting specialized evaluation metrics for assessing scientific analysis quality (\S\ref{subsec:anabench:evaluation_metrics}), results in \S\ref{sec:results} demonstrate that \ours significantly improves scientific table \& figure analysis through test-time optimization ($\Delta_{\textit{rel}} \geq$ $\uparrow 13.43\%$) and modular training ($\Delta_{\textit{rel}} \geq$ $\uparrow 42.12\%$).
\end{itemize}


\input{figures/preliminary}

%% file: figures/preliminary.tex
\begin{figure}[!t]

\vspace{0pt}

\centering
\includegraphics[width=1.0\linewidth]{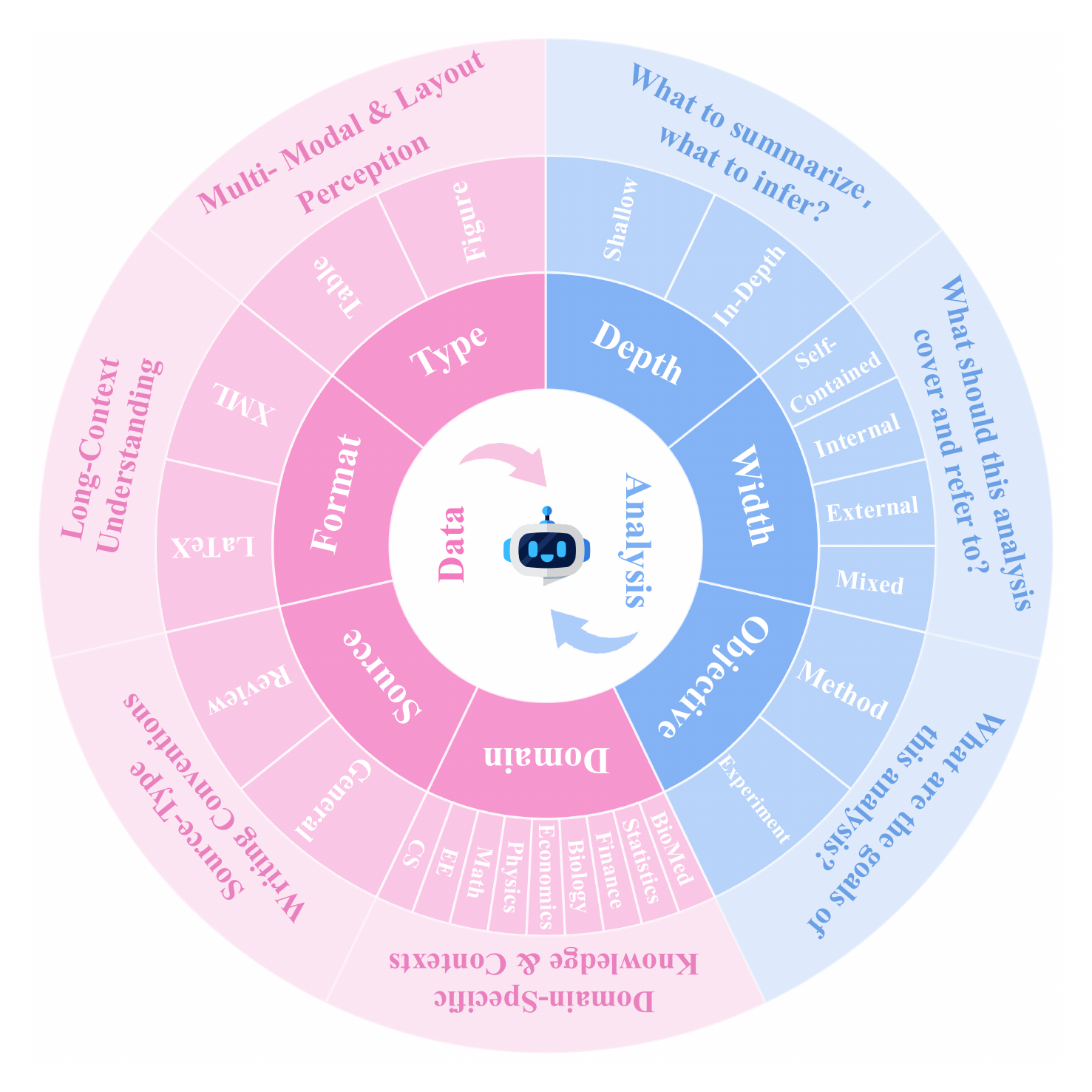}


\caption{\textbf{Challenges In Scientific Table \& Figure Analysis.} The heterogeneity of scientific literature presents great challenges for high-quality analysis of scientific tables and figures (Fig.~\ref{fig:preliminary-error-analysis}).}
\label{fig:preliminary}

\vspace{12pt}

\end{figure}

%% file: sections/2_anabench.tex
\section{\ourbench: Evaluating Scientific Analysis}
\label{sec:anabench}

\input{figures/anabench}


\subsection{Problem Formulation: Scientific Analysis}
\label{subsec:anabench:problem_formulation}

We formulate the task of \textit{scientific table \& figure analysis} as a context-aware generation problem:
Given an input $x$ comprising one or more tables $\{x_t\}$ and/or one or more figures $\{x_f\}$, together with their source information $s$ and input query $q$ that specifies the analysis requirements and objectives, the goal is to generate a well-written analysis $y$ that accurately interprets the provided tabular and visual data, integrates evidence across all available contexts, situates findings within the broader research, and delivers domain-specific insights.
Formally, the task can be expressed as:

\vspace{-16pt}
\begin{equation}
y = f(x, s, q) \quad \text{where } x \in \{\{x_t\}, \{x_f\}, \{x_t, x_f\}\}
\end{equation}
\vspace{-6pt}

\noindent
As such, this scientific analysis writing problem encompasses \textbf{\textit{multimodal long-context reasoning}} for input tables and figures with different formats and layouts.

\subsection{Preliminary: MLLM Agents In Scientific Analysis}
\label{subsec:anabench:motivations}

To empirically assess the challenges faced by MLLM agents in scientific analysis, we conduct a preliminary study (\S\ref{appendix:preliminary}) evaluating their performance across seven complexity dimensions (Fig.~\ref{fig:preliminary}). We randomly select 120 samples from \ourbench with all seven challenges evenly distributed, and employ \texttt{Qwen3-VL-8B} as MLLM agent backbone to generate scientific analysis. Performance is evaluated using SciBERT (Eq.~\ref{eq:scibert_score}).
With performance struggling to exceed 60\% across all metrics, Fig.~\ref{fig:preliminary-error-analysis} reveals pronounced difficulties in multimodal, multi-layout understanding and in-depth analysis that demand inferential generation. These findings highlight that \textit{MLLM agents face substantial challenges in interpreting complex heterogeneous scientific artifacts}.

\subsection{Benchmarking Scientific Analysis}
\label{subsec:anabench:benchmark_construction}



\textbf{Benchmark Construction.}
Our benchmark construction method comprises four stages (Fig.~\ref{fig:anabench}):
(1) \textit{Source collection}, which identifies and collects candidate source papers that satisfy predefined relevance and retrieval criteria.
(2) \textit{Data extraction}, which extracts tables, figures, and their associated contexts. A context retrieval depth $d$ controls the level of context referenced by each table or figure. Extracted data are augmented via two-level filtering: \emph{paper-level filtering} removes papers that fail to meet validity requirements, and \emph{data-level filtering} excludes tables and figures with formatting errors, missing information, or other quality issues.
(3) \textit{Instance construction}, which transforms each filtered data into a scientific analysis instance. Each instance consists of table and/or figure data, corresponding contexts, metadata, and the gold analysis. Resulting instances are further refined through a specialized \emph{data cleaning} step using configurable thresholds, including the maximum number of samples and the minimum length of ground truths.
(4) \textit{MLLM-assisted task classification}, which combines rule-based heuristics with MLLM classification (\S\ref{appendix:benchmark:curriculum:task_classification}) to categorize \ourbench along seven dimensions (\S\ref{appendix:benchmark:curriculum}). 
%
Through four-stage construction, \ourbench achieves large-scale coverage across seven complexity dimensions while faithfully reflecting real-world distributions of data characteristics and analytical challenges.

\noindent\textbf{Data Complexity.}
We consider four data complexity dimensions (\S\ref{appendix:benchmark:curriculum:data_complexity}):
(1) \textbf{\textit{Type}}: the type of analysis data (\textit{table}, \textit{figure}, or \textit{both});
(2) \textbf{\textit{Domain}}: domain disciplines that the source paper belongs to, with \ourbench spanning 9 broad domains across 170 disciplines;
(3) \textbf{\textit{Format}}: the format of analysis writing (\texttt{LaTeX} or \texttt{XML});
(4) \textbf{\textit{Source}}: the type of source paper (\textit{general} research papers or \textit{reviews \& surveys}).


\noindent\textbf{Analysis Complexity.}
We characterize analysis complexity along three complementary dimensions (\S\ref{appendix:benchmark:curriculum:analysis_complexity}):
(1) \textbf{\textit{Width}}: the reference scope of the analysis (\textit{self-contained}, \textit{internal}, \textit{external}, or \textit{mixed});
(2) \textbf{\textit{Depth}}: the level of analytical rigor (\textit{shallow} or \textit{in-depth});
(3) \textbf{\textit{Objective}}: the primary goal and focus of the analysis (\textit{methodology} or \textit{experiment}).


\input{figures/anagent}

\subsection{Evaluating Scientific Analysis}
\label{subsec:anabench:evaluation_metrics}


\textbf{Rule-Based Evaluation.}
Rule-based evaluation metrics cover both lexical and semantic assessment, where lexical evaluation include ROUGE-L (Eq.~\ref{eq:rouge_l}) \cite{lin2004rouge}, BLEU (Eq.~\ref{eq:bleu}) \cite{papineni2002bleu}, and word overlap (Eq.~\ref{eq:word_overlap}); while semantic assessment calculates similarity between model generated analysis $y$ and ground-truth analysis $y^*$ through cosine similarity (Eq.~\ref{eq:cosine_sim}), SciBERT (Eq.~\ref{eq:scibert_score}) \cite{Beltagy2019SciBERT}, and METEOR (Eq.~\ref{eq:meteor}) \cite{banerjee2005meteor} scores.

\noindent\textbf{MLLM-As-Judge.}
For more reliable evaluation, we implement MLLM-as-judge by leveraging Gemini-2.5-Flash and GPT-4.1-mini to grade each generated analysis across five dimensions (Fig.~\ref{fig:five_dimension_evaluation_prompt}, \S\ref{appendix:preliminary:method} \& \ref{appendix:evaluation}), including \textit{analysis consistency}, \textit{query-analysis alignment}, \textit{knowledge utilization}, \textit{format correctness}, and \textit{grounding accuracy}.


%% file: figures/anabench.tex
\begin{figure*}[!t]


\centering
\includegraphics[width=1.0\linewidth]{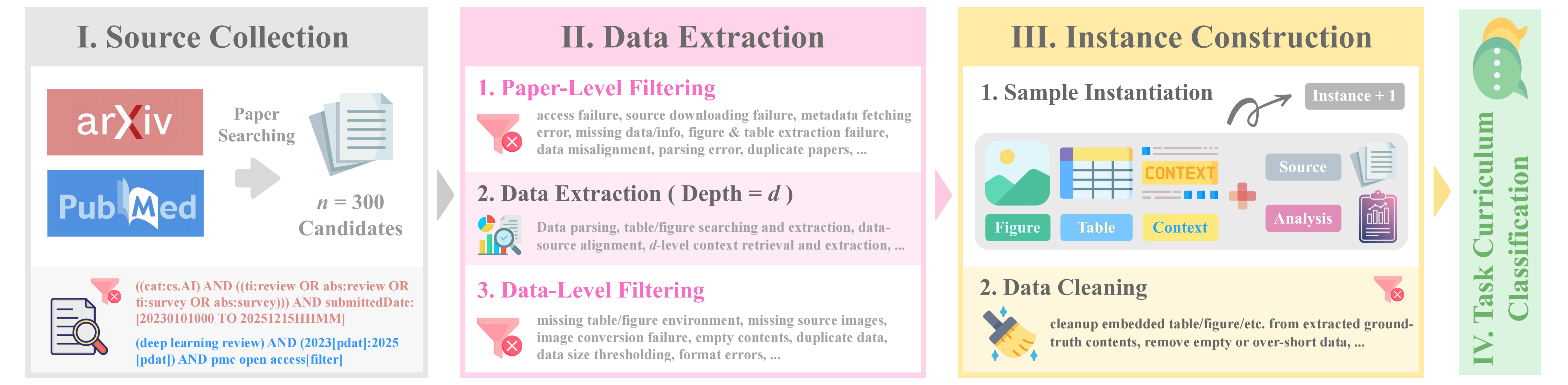}

\caption{\textbf{\ourbench For Evaluating Autonomous Scientific Analysis.} We implement four-stage benchmark construction method to build \ourbench, with multi-level filtering to enhance data quality.}
\label{fig:anabench}

\vspace{6pt}

\end{figure*}

%% file: figures/anagent.tex
\begin{figure*}[!t]


\centering
\includegraphics[width=1.0\linewidth]{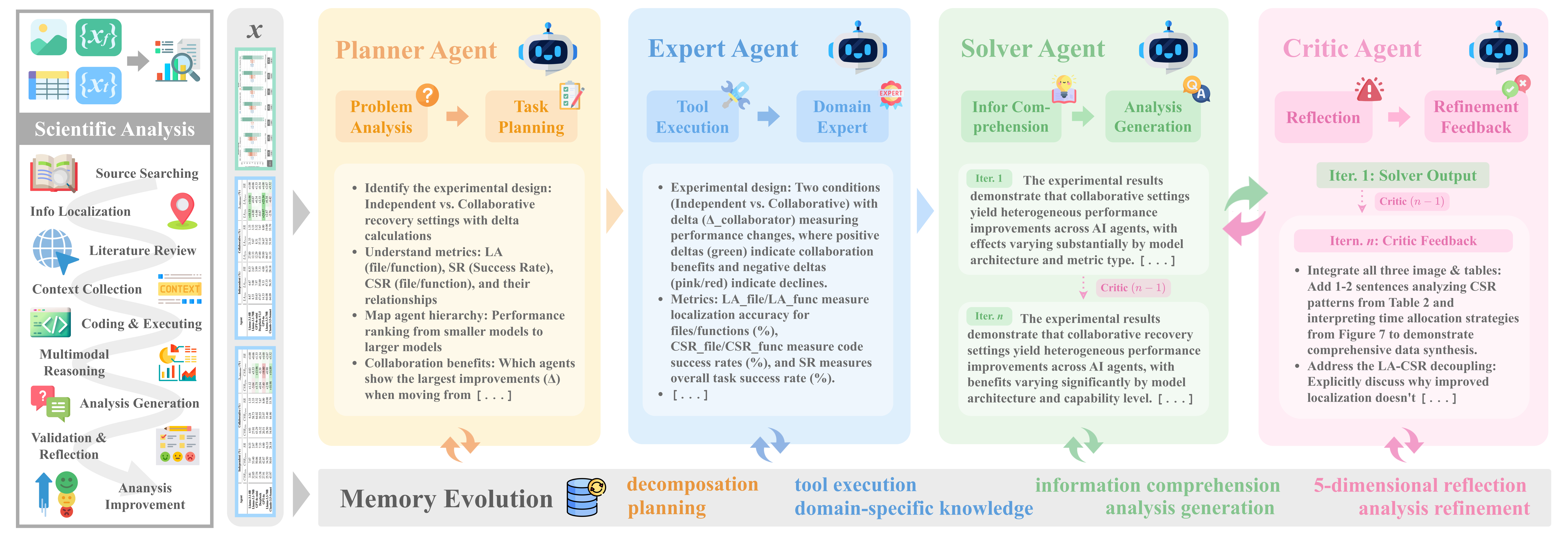}

\caption{\textbf{Multi-Agent Coordinative Scientific Analysis.} Our multi-Agent scientific analysis framework, \ours, is developed to cover various stages to analyze scientific tables and figures through four collaborative agents: \Planner, \Expert, \Solver, and \Critic. Some example details are omitted as \textbf{\texttt{[...]}} for clarity.}
\label{fig:anagent}

\vspace{6pt}

\end{figure*}

%% file: tables/exp_main1_training_free.tex
\definecolor{baselinebg}{HTML}{E6E6E6}
\definecolor{oursbaselinebg}{HTML}{DCE9FD}
\definecolor{ourstrainedbg}{HTML}{D9F4CF}

\definecolor{deltagreen}{HTML}{5CB24B}
\definecolor{deltared}{HTML}{ED7777}

\newcommand{\reldeltapctbelow}[2]{%
  \FPeval{\result}{round((#2-#1)/#1*100:2)}%
  \FPifpos\result%
    {\textbf{\textcolor{deltagreen}{↑\result\%}}}%
  \else%
    \FPeval{\absresult}{round(0-\result:2)}%
    {\textbf{\textcolor{deltared}{↓\absresult\%}}}%
  \fi%
}

\newcommand{\reldeltapct}[2]{%
  \FPeval{\result}{round((#2-#1)/#1*100:2)}%
  \FPifpos\result%
    {\textbf{\textcolor{deltagreen}{↑\result\%}}}%
  \else%
    \FPeval{\absresult}{round(0-\result:2)}%
    {\textbf{\textcolor{deltared}{↓\absresult\%}}}%
  \fi%
}

\newcommand{\reldeltapctbf}[2]{%
  \FPeval{\result}{round((#2-#1)/#1*100:2)}%
  \FPifpos\result%
    {\textbf{\textcolor{deltagreen}{↑\result\%}}}%
  \else%
    \FPeval{\absresult}{round(0-\result:2)}%
    {\textbf{\textcolor{deltared}{↓\absresult\%}}}%
  \fi%
}

\newcommand{\reldeltapctul}[2]{%
  \FPeval{\result}{round((#2-#1)/#1*100:2)}%
  \FPifpos\result%
    {\textbf{\textcolor{deltagreen}{↑\result\%}}}%
  \else%
    \FPeval{\absresult}{round(0-\result:2)}%
    {\textbf{\textcolor{deltared}{↓\absresult\%}}}%
  \fi%
}

\newcommand{\absdeltapct}[2]{%
  \FPeval{\result}{round(#2-#1:2)}%
  \FPifpos\result%
    {\textbf{\textcolor{deltagreen}{↑\result}}}%
  \else%
    \FPeval{\absresult}{round(0-\result:2)}%
    {\textbf{\textcolor{deltared}{↓\absresult}}}%
  \fi%
}

\newcommand{\absdeltapctbf}[2]{%
  \FPeval{\result}{round(#2-#1:2)}%
  \FPifpos\result%
    {\textbf{\textcolor{deltagreen}{↑\result}}}%
  \else%
    \FPeval{\absresult}{round(0-\result:2)}%
    {\textbf{\textcolor{deltared}{↓\absresult}}}%
  \fi%
}

\newcommand{\absdeltapctul}[2]{%
  \FPeval{\result}{round(#2-#1:2)}%
  \FPifpos\result%
    {\textbf{\textcolor{deltagreen}{↑\result}}}%
  \else%
    \FPeval{\absresult}{round(0-\result:2)}%
    {\textbf{\textcolor{deltared}{↓\absresult}}}%
  \fi%
}

\newcommand{\absdeltapctbful}[2]{%
  \FPeval{\result}{round(#2-#1:2)}%
  \FPifpos\result%
    {\textbf{\textcolor{deltagreen}{↑\result}}}%
  \else%
    \FPeval{\absresult}{round(0-\result:2)}%
    {\textbf{\textcolor{deltared}{↓\absresult}}}%
  \fi%
}


\begin{table*}[t]

\vspace{-12pt}

\small
\centering
\renewcommand{\arraystretch}{1.1}  
\setlength{\tabcolsep}{4pt}  

\begin{tabular*}{\textwidth}{@{\extracolsep{\fill}}l c ccc ccc ccc@{}}
\toprule
\multirow{2}{*}{\textbf{Model}} & \multirow{2}{*}{\textbf{Size}} & \multicolumn{3}{c}{\textbf{Semantic Accuracy (\%)}} & \multicolumn{3}{c}{\textbf{Lexical Accuracy (\%)}} & \multicolumn{3}{c}{\textbf{Overall Accuracy (\%)}} \\
\cmidrule(lr){3-5} \cmidrule(lr){6-8} \cmidrule(lr){9-11}
& & \textbf{\textsc{Cosine}} & \textbf{BERT} & \textbf{\textsc{Meteor}} & \textbf{\textsc{Rouge-L}} & \textbf{\textsc{Bleu}} & \textbf{\textsc{Word}} & $\bm{S}_{\textbf{\textsc{Sem}}}$ & $\bm{S}_{\textbf{\textsc{Lex}}}$ & $\bm{S}_{\textbf{\textsc{Avg}}}$ \\

\midrule

\multicolumn{11}{c}{\cellcolor{baselinebg}\textbf{Baselines}} \\
\addlinespace[3pt]

\textbf{GPT-4.1-mini} & \textbf{-} & \textbf{56.34} & \textbf{59.74} & \textbf{19.47} & 16.74 & \textbf{3.39} & \textbf{11.49} & \textbf{45.18} & \textbf{10.54} & \textbf{27.86} \\
\textbf{Gemini-2.5-Flash} & \textbf{-} & 52.41 & 55.99 & 19.01 & 14.90 & 2.76 & 9.95 & 42.47 & 9.20 & 25.84 \\
\multirow{2}{*}{\textbf{InternVL-3.5}} & \textbf{4B} & 54.38 & 58.19 & 18.76 & 15.67 & 2.66 & 9.80 & 43.78 & 9.37 & 26.58 \\
& \textbf{8B} & 55.73 & 59.10 & \underline{19.30} & \underline{16.80} & 2.86 & 10.28 & 44.71 & 9.98 & 27.34 \\
\multirow{2}{*}{\textbf{Qwen2.5-VL}} & \textbf{3B} & 54.74 & 58.49 & 17.82 & 15.89 & 2.56 & 10.02 & 43.68 & 9.49 & 26.59 \\
& \textbf{7B} & 55.65 & \underline{59.66} & 18.90 & 16.40 & 2.98 & 10.38 & 44.74 & 9.98 & 27.31 \\
\multirow{2}{*}{\textbf{Qwen3-VL}} & \textbf{4B} & 55.41 & 58.15 & 18.41 & 15.77 & 2.77 & 10.06 & 43.99 & 9.53 & 26.76 \\
& \textbf{8B} & \underline{55.94} & 59.11 & 19.16 & \textbf{17.06} & \underline{3.02} & \underline{10.39} & \underline{44.73} & \underline{10.16} & \underline{27.44} \\


\multicolumn{11}{c}{\cellcolor{oursbaselinebg}\textbf{\ours (Zero-Shot)}} \\
\addlinespace[3pt]

\textbf{GPT-4.1-mini} & \textbf{-} & \textbf{59.94} & \textbf{61.63} & \underline{22.75} & \textbf{18.19} & \underline{4.81} & \textbf{12.26} & \reldeltapctul{45.18}{48.11} & \reldeltapctbf{10.54}{11.75} & \reldeltapctbf{27.86}{29.93} \\

\textbf{Gemini-2.5-Flash} & \textbf{-} & 55.60 & 59.37 & 19.40 & 16.04 & 3.15 & 11.10 & \reldeltapct{42.47}{44.79} & \reldeltapct{9.20}{10.09} & \reldeltapct{25.84}{27.44} \\

\multirow{2}{*}{\textbf{InternVL-3.5}} & \textbf{4B} & 58.26 & 59.86 & 21.21 & 16.10 & 3.29 & 11.11 & \reldeltapct{43.78}{46.44} & \reldeltapct{9.37}{10.17} & \reldeltapct{26.58}{28.31} \\

& \textbf{8B} & 59.46 & 61.25 & 22.59 & 17.00 & 3.88 & 11.68 & \reldeltapct{44.71}{47.77} & \reldeltapct{9.98}{10.85} & \reldeltapct{27.34}{29.31} \\

\multirow{2}{*}{\textbf{Qwen2.5-VL}} & \textbf{3B} & 57.50 & 60.01 & 21.03 & 17.34 & 3.87 & 11.53 & \reldeltapct{43.68}{46.18} & \reldeltapct{9.49}{10.91} & \reldeltapct{26.59}{28.55} \\

& \textbf{7B} & 58.91 & 60.41 & 21.59 & 17.47 & 4.11 & 11.85 & \reldeltapct{44.74}{46.97} & \reldeltapct{9.98}{11.14} & \reldeltapct{27.31}{29.06} \\

\multirow{2}{*}{\textbf{Qwen3-VL}} & \textbf{4B} & 59.41 & 60.21 & 21.23 & 16.27 & 3.90 & 11.33 & \reldeltapct{43.99}{46.95} & \reldeltapct{9.53}{10.50} & \reldeltapct{26.76}{28.73} \\

& \textbf{8B} & \underline{59.76} & \underline{61.53 }& \textbf{23.07} & \underline{17.75} & \textbf{4.98} & \underline{12.20} & \reldeltapctbf{44.73}{48.12} & \reldeltapctul{10.16}{11.64} & \reldeltapctul{27.44}{29.88} \\


\multicolumn{11}{c}{\cellcolor{oursbaselinebg}\textbf{\ours (One-Shot)}} \\
\addlinespace[3pt]

\textbf{GPT-4.1-mini} & \textbf{-} & \underline{60.87} & \textbf{63.28} & \underline{24.26} & \textbf{20.65} & \underline{5.73} & 12.55 & \reldeltapctbf{45.18}{49.47} & \reldeltapctbf{10.54}{12.98} & \reldeltapctbf{27.86}{31.22} \\

\textbf{Gemini-2.5-Flash} & \textbf{-} & \textbf{61.06} & 61.34 & 20.52 & 17.40 & 4.06 & 11.47 & \reldeltapct{42.47}{47.64} & \reldeltapct{9.20}{10.98} & \reldeltapct{25.84}{29.31} \\

\multirow{2}{*}{\textbf{InternVL-3.5}} & \textbf{4B} & 59.11 & 60.52 & 22.60 & 18.04 & 3.82 & 11.50 & \reldeltapct{43.78}{47.41} & \reldeltapct{9.37}{11.12} & \reldeltapct{26.58}{29.27} \\

& \textbf{8B} & 60.26 & 62.12 & 23.18 & 19.14 & 4.56 & \textbf{12.97} & \reldeltapct{44.71}{48.52} & \reldeltapct{9.98}{12.22} & \reldeltapct{27.34}{30.37} \\

\multirow{2}{*}{\textbf{Qwen2.5-VL}} & \textbf{3B} & 58.89 & 60.70 & 22.19 & 18.41 & 3.99 & 11.54 & \reldeltapct{43.68}{47.26} & \reldeltapct{9.49}{11.31} & \reldeltapct{26.59}{29.29} \\

& \textbf{7B} & 60.24 & 61.00 & 23.41 & 19.41 & 4.98 & 12.47 & \reldeltapct{44.74}{48.22} & \reldeltapct{9.98}{12.29} & \reldeltapct{27.31}{30.25} \\

\multirow{2}{*}{\textbf{Qwen3-VL}} & \textbf{4B} & 59.64 & 60.61 & 22.42 & 18.05 & 4.03 & 11.51 & \reldeltapct{43.99}{47.55} & \reldeltapct{9.53}{11.20} & \reldeltapct{26.76}{29.38} \\

& \textbf{8B} & 60.55 & \underline{62.27} & \textbf{24.65} & \underline{20.06} & \textbf{5.92} & \underline{12.95} & \reldeltapctul{44.73}{49.15} & \reldeltapctul{10.16}{12.98} & \reldeltapctul{27.44}{31.07} \\

\bottomrule
\end{tabular*}

\caption{\textbf{Evaluation of Training-Free Agents.} Comparing between baselines and training-free \ours ($M_e=5$) on \ourbench (\S\ref{appendix:anabench:train_eval_data}), \textbf{\textit{relative differences}} (Eq.~\ref{eq:relative_delta}) are shown as \textit{positive} \textcolor{deltagreen}{$\uparrow \Delta_{\textit{rel}}\%$} or \textit{negative} \textcolor{deltared}{$\downarrow \Delta_{\textit{rel}}\%$}.}
\label{tab:experiment:training_free_main_results}

\vspace{12pt}

\end{table*}

%% file: sections/3_anagent.tex
\section{\ours: Multi-Agent Scientific Analysis}
\label{sec:anagent}

\subsection{\ours For Scientific Table \& Figure Analysis}
\label{subsec:anagent:anagent}

Facing challenges in both data and analysis levels (Fig.~\ref{fig:preliminary}), traditional approaches that directly map inputs to outputs struggle with varying task complexities due to their lack of systematic reasoning and knowledge retrieval capabilities.

\noindent\textbf{\textit{How do human scientists analyze tables and figures?}} Instead of simply describing what we observe, we engage in a deliberate process of understanding the research question, planning the problem-solving, gathering relevant domain knowledge, interpreting the data in context, and rigorously evaluating our findings and conclusions (Fig.~\ref{fig:teaser}).

\noindent\textbf{Our Approach.}
Inspired by human analysis workflow (Fig.~\ref{fig:teaser}), we propose \ours (Fig.~\ref{fig:anagent}), a multi-agent system for enhanced table \& figure analysis.
Given input $x$,
source $s$, and query $q$, \ours operates through four interactive stages (\S\ref{appendix:anagent}):

\noindent\textbf{Stage 1: Task Decomposition.}
\Planner analyzes the input and decomposes the complex task into actionable subtasks $\tau_i$ ($i=1,\ldots,M_p$):

\vspace{-16pt}
\begin{equation}
\Planner(x, s, q) = \{\tau_1, \tau_2, \ldots, \tau_{M_p}\}
\label{eq:planner}
\end{equation}
\vspace{-18pt}

\noindent\textbf{Stage 2: Task-Oriented Knowledge Retrieval.}
\Expert performs iterative knowledge acquisition through multi-turn tool executions. At each turn $e$, the knowledge base $\mathcal{K}_e$ is expanded by incorporating new knowledge retrieved based on subtask $\tau_e$ and previously accumulated knowledge $\mathcal{K}_{e-1}$:

\vspace{-20pt}
\begin{equation}
\resizebox{\linewidth}{!}{$\displaystyle
\mathcal{K}_e = \mathcal{K}_{e-1} \cup \Expert(\tau_e, \mathcal{K}_{e-1}), \quad e=1,\ldots,M_e
$}
\label{eq:expert}
\end{equation}
\vspace{-16pt}

\noindent\textbf{Stage 3: Solution Generation.}
\Solver synthesizes the accumulated knowledge $\mathcal{K}_n$ with the input to generate candidate analysis. At iteration $i$, it incorporates feedback $f_{i-1}$:

\vspace{-20pt}
\begin{equation}
\begin{aligned}
y_i &= \Solver(x, s, q, \mathcal{K}_n, f_{i-1}), \quad i=1,\ldots,M_s
\end{aligned}
\label{eq:solver}
\end{equation}
\vspace{-20pt}

\noindent\textbf{Stage 4: Reflective Refinement.}
\Critic assesses generated analysis through five-dimensional evaluation protocol (\S\ref{appendix:evaluation:five_dimension}) and provides feedback for iterative improvement:

\vspace{-16pt}
\begin{equation}
f_i = \Critic(y_i, x, s, q, \mathcal{K}_n), \quad i=1,\ldots,M_c
\label{eq:critic}
\end{equation}
\vspace{-18pt}

The interactive refinement between \Solver (Eq.~\ref{eq:solver}) and \Critic (Eq.~\ref{eq:critic}) produces the final analysis $y = y_M$.
%
%
%
%
%
%






\subsection{Scientific ToolKits}
\label{subsec:anagent:tool}

To facilitate complex scientific analysis spanning multiple stages from source searching to analysis writing (Fig.~\ref{fig:anagent}), we develop 5 scientific toolkits with 16 specialized tools (Tab.~\ref{tab:five_toolkits}) to enable efficient scientific analysis with improved accuracy and comprehensiveness (\S\ref{appendix:anagent:tool}).

\subsection{Multi-Agent Optimization}
\label{subsec:anagent:infer}

\textbf{Few-Shot Optimization.}
To enhance the adaptability of individual agents, we include few-shot learning by providing each agent with $k$-shot exemplars. These examples guide agents to perform specialized tasks effectively, enabling test-time adaptation without extensive task-specific training.

\noindent\textbf{Critic-Guided Reflective Optimization.}
To further improve collaborative performance, we incorporate a dedicated \Critic that assesses and optimizes \Solver's analysis solutions.
Through five-dimensional protocol (\S\ref{appendix:evaluation:five_dimension}),
\Critic provides targeted feedback to guide \Solver optimizing analysis solution, reducing errors, improving logical consistency, and mitigating hallucinations.

\noindent\textbf{Agent-Level Capability Augmentation.}
In multi-agent systems, overall performance is significantly influenced by individual agents' capabilities. To this end, we introduce agent-level capability augmentation, a strategy in which individual agents can be independently enhanced by more capable models to improve system-level outcomes, enabling selective upgrades at test time. 


\subsection{Modular Training}
\label{subsec:anagent:train}

\textit{How to train \ours to enhance individual agent capabilities while maintaining effective global collaboration?}
We develop a modular training paradigm that aligns with the functional decomposition of \ours. Each agent is first initialized via supervised finetuning (SFT)
to establish analysis and reasoning foundations, followed by agent-specific reinforcement learning (RL) to optimize specialized behaviors and capabilities.

\noindent\textbf{Supervised Finetuning.}
All agents in \ours are initialized through the SFT phase on the scientific analysis writing training set (Tab.~\ref{tab:sft_data_size}) randomly sampled from \ourbench (\S\ref{sec:anabench}). Each training instance consists of the multimodal input $x \in \{x_t, x_f\}$, source information $s$, query $q$, and the corresponding ground-truth analysis $y^\ast$.
Let $\theta$ denote the shared model parameters. The SFT objective (Eq.~\ref{eq:sft_objective}) is to minimize the token-level negative log-likelihood of the reference analysis conditioned on the input (\S\ref{subsec:anagent:train}):

\vspace{-12pt}
\begin{equation}
\small
\mathcal{L}_{\mathrm{SFT}}(\theta)
= 
\mathbb{E}_{(x,s,q,y^\ast)}
\left[
- \sum_{t=1}^{|y^\ast|}
\log p_\theta \!\left( y_t^\ast \mid y_{<t}^\ast, x, s, q \right)
\right]
\label{eq:sft_objective}
\end{equation}
\vspace{-6pt}


\noindent\textbf{RL Optimization.}
After SFT initialization, each agent is further optimized via RL with Group Relative Policy Optimization (GRPO) \cite{Shao2024DeepSeekMathPT}.
For each agent $a \in \{\Planner, \Expert, \Solver, \Critic\}$, we define an agent-specific policy $\pi_{\theta_a}$ derived from SFT initialization and optimized on specialized RL datasets (\S\ref{appendix:anagent:rl_training}).
Given an input state $\xi_a$ and a sampled action $z_a \sim \pi_{\theta_a}(\cdot \mid \xi_a)$, GRPO maximizes the expected relative advantage within a sampled group $\mathcal{G}_a = \{z_a^{(1)}, \dots, z_a^{(K)}\}$:

\vspace{-10pt}
\begin{equation}
\small
\mathcal{L}_{\mathrm{RL}}^{(a)}(\theta_a)
=
- \mathbb{E}_{\xi_a}
\left[
\frac{1}{K}
\sum_{k=1}^{K}
\hat{A}_a^{(k)}
\log \pi_{\theta_a}\!\left(z_a^{(k)} \mid \xi_a\right)
\right]
\label{eq:grpo_objective}
\end{equation}
\vspace{-6pt}


\noindent where $K$ represents the number of sampled candidate actions in each GRPO group, 
$\hat{A}_a^{(k)}$ denotes the normalized advantage computed from relative rewards within the group $\mathcal{G}_a$, 
and $\mathbb{E}_{\xi_a}[\cdot]$ shows the expectation over the agent-specific input distribution $\mathcal{D}_a$.

\noindent\textbf{Specialized Rewards.}
Each agent $a$ is optimized with a specialized reward tailored to its functional role (\S\ref{appendix:anagent:rl_training}).
Let $R_a$ denote the total reward for agent $a$, which decomposes into a weighted sum of multiple components:
\vspace{-3pt}
\begin{equation}
R_a = \sum_{m} \lambda_{a,m} \, r_{a,m}
\label{eq:reward_decomposition}
\end{equation}
\vspace{-6pt}

\noindent where $m$ is the index over reward components for agent $a$, $r_{a,m}$ is individual reward term, and $\lambda_{a,m}$ is its corresponding weight with $\sum_m \lambda_{a,m} = 1$.

%% file: tables/exp_main2_finetuned.tex

\begin{table*}[t]

\vspace{-12pt}

\small
\centering
\renewcommand{\arraystretch}{1.1}  
\setlength{\tabcolsep}{4pt}  

\begin{tabular*}{\textwidth}{@{\extracolsep{\fill}}l c ccc ccc ccc@{}}

\toprule

\multirow{2}{*}{\textbf{Model}} & \multirow{2}{*}{\textbf{Size}} & \multicolumn{3}{c}{\textbf{Semantic Accuracy (\%)}} & \multicolumn{3}{c}{\textbf{Lexical Accuracy (\%)}} & \multicolumn{3}{c}{\textbf{Overall Accuracy (\%)}} \\
\cmidrule(lr){3-5} \cmidrule(lr){6-8} \cmidrule(lr){9-11}
& & \textbf{\textsc{Cosine}} & \textbf{BERT} & \textbf{\textsc{Meteor}} & \textbf{\textsc{Rouge-L}} & \textbf{\textsc{Bleu}} & \textbf{\textsc{Word}} & $\bm{S}_{\textbf{\textsc{Sem}}}$ & $\bm{S}_{\textbf{\textsc{Lex}}}$ & $\bm{S}_{\textbf{\textsc{Avg}}}$ \\

\midrule

\multicolumn{11}{c}{\cellcolor{ourstrainedbg}\textbf{\ours | SFT (Zero-Shot)}} \\
\addlinespace[3pt]

\multirow{2}{*}{\textbf{InternVL-3.5}} & \textbf{4B} & 62.38 & 64.78 & 28.38 & 24.77 & 13.10 & 18.90 & \reldeltapct{43.78}{51.85} & \reldeltapct{9.37}{18.92} & \reldeltapct{26.58}{35.39} \\
& \textbf{8B} & \underline{64.16} & 65.82 & \underline{29.62} & 25.26 & \underline{14.27} & \underline{20.56} & \reldeltapctul{44.71}{53.20} & \reldeltapctul{9.98}{20.03} & \reldeltapctul{27.34}{36.61} \\

\multirow{2}{*}{\textbf{Qwen2.5-VL}} & \textbf{3B} & 60.84 & 65.77 & 25.09 & 24.27 & 11.19 & 17.58 & \reldeltapct{43.68}{50.57} & \reldeltapct{9.49}{17.68} & \reldeltapct{26.59}{34.12} \\
& \textbf{7B} & 63.21 & \underline{65.97} & 27.63 & 24.91 & 13.96 & 20.21 & \reldeltapct{44.74}{52.27} & \reldeltapct{9.98}{19.69} & \reldeltapct{27.31}{35.98} \\

\multirow{2}{*}{\textbf{Qwen3-VL}} & \textbf{4B} & 62.98 & 64.72 & 28.86 & \underline{25.75} & 14.11 & 19.68 & \reldeltapct{43.99}{52.19} & \reldeltapct{9.53}{19.85} & \reldeltapct{26.76}{36.02} \\
& \textbf{8B} & \textbf{64.70} & \textbf{66.98} & \textbf{31.33} & \textbf{27.93} & \textbf{16.47} & \textbf{22.09} & \reldeltapctbf{44.73}{54.34} & \reldeltapctbf{10.16}{22.16} & \reldeltapctbf{27.44}{38.25} \\


\multicolumn{11}{c}{\cellcolor{ourstrainedbg}\textbf{\ours | SFT (One-Shot)}} \\
\addlinespace[3pt]

\multirow{2}{*}{\textbf{InternVL-3.5}} & \textbf{4B} & 62.79 & 65.60 & 27.99 & 24.51 & 13.72 & 18.75 & \reldeltapct{43.78}{52.13} & \reldeltapct{9.37}{18.99} & \reldeltapct{26.58}{35.56} \\

& \textbf{8B} & \underline{64.97} & 66.63 & 29.39 & 25.01 & \underline{14.63} & \underline{20.79} & \reldeltapctul{44.71}{53.66} & \reldeltapct{9.98}{20.14} & \reldeltapctul{27.34}{36.90} \\

\multirow{2}{*}{\textbf{Qwen2.5-VL}} & \textbf{3B} & 61.25 & 66.54 & 25.27 & 24.27 & 11.26 & 17.53 & \reldeltapct{43.68}{51.02} & \reldeltapct{9.49}{17.69} & \reldeltapct{26.59}{34.35} \\

& \textbf{7B} & 63.93 & \underline{66.78} & \underline{29.31} & \underline{26.47} & 14.22 & 20.60 & \reldeltapct{44.74}{53.34} & \reldeltapctul{9.98}{20.43} & \reldeltapct{27.31}{36.89} \\

\multirow{2}{*}{\textbf{Qwen3-VL}} & \textbf{4B} & 63.51 & 65.22 & 28.30 & 26.34 & 14.05 & 19.77 & \reldeltapct{43.99}{52.34} & \reldeltapct{9.53}{20.05} & \reldeltapct{26.76}{36.20} \\

& \textbf{8B} & \textbf{65.07} & \textbf{67.13} & \textbf{31.45} & \textbf{28.08} & \textbf{16.01} & \textbf{22.59} & \reldeltapctbf{44.73}{54.55} & \reldeltapctbf{10.16}{22.22} & \reldeltapctbf{27.44}{38.39} \\


\multicolumn{11}{c}{\cellcolor{ourstrainedbg}\textbf{\ours | RL (Zero-Shot)}} \\
\addlinespace[3pt]

\textbf{Qwen2.5-VL} & \textbf{3B} & 56.54 & 60.49 & 21.37 & 18.87 & 5.67 & 12.72 & \reldeltapct{43.68}{46.13} & \reldeltapct{9.49}{12.42} & \reldeltapct{26.59}{29.28} \\

\textbf{Qwen3-VL} & \textbf{4B} & \textbf{58.97} & \textbf{60.99} & \textbf{22.33} & \textbf{19.09} & \textbf{5.91} & \textbf{12.76} & \reldeltapctbf{43.99}{47.43} & \reldeltapctbf{9.53}{12.58} & \reldeltapctbf{26.76}{30.03} \\


\multicolumn{11}{c}{\cellcolor{ourstrainedbg}\textbf{\ours | RL (One-Shot)}} \\
\addlinespace[3pt]

\textbf{Qwen2.5-VL} & \textbf{3B} & 58.19 & 61.10 & 22.01 & 19.40 & 6.20 & \textbf{13.63} & \reldeltapct{43.68}{47.10} & \reldeltapctbf{9.49}{13.07} & \reldeltapct{26.59}{30.09} \\
\textbf{Qwen3-VL} & \textbf{4B} & \textbf{60.42} & \textbf{62.24} & \textbf{22.70} & \textbf{19.51} & \textbf{6.21} & 12.95 & \reldeltapctbf{43.99}{48.45} & \reldeltapct{9.53}{12.89} & \reldeltapctbf{26.76}{30.67} \\


\multicolumn{11}{c}{\cellcolor{ourstrainedbg}\textbf{\ours | SFT+RL (Zero-Shot)}} \\
\addlinespace[3pt]

\textbf{Qwen2.5-VL} & \textbf{3B} & 62.61 & 65.90 & 27.22 & 25.22 & 12.59 & 18.78 & \reldeltapct{43.68}{51.91} & \reldeltapct{9.49}{18.86} & \reldeltapct{26.59}{35.39} \\
\textbf{Qwen3-VL} & \textbf{4B} & \textbf{63.13} & \textbf{66.66} & \textbf{29.46} & \textbf{26.87} & \textbf{14.84} & \textbf{20.73} & \reldeltapctbf{43.99}{53.08} & \reldeltapctbf{9.53}{20.81} & \reldeltapctbf{26.76}{36.95} \\


\multicolumn{11}{c}{\cellcolor{ourstrainedbg}\textbf{\ours | SFT+RL (One-Shot)}} \\
\addlinespace[3pt]

\textbf{Qwen2.5-VL} & \textbf{3B} & 62.92 & 66.63 & 27.64 & 26.82 & 14.41 & 19.63 & \reldeltapct{43.68}{52.40} & \reldeltapct{9.49}{20.29} & \reldeltapct{26.59}{36.34} \\
\textbf{Qwen3-VL} & \textbf{4B} & \textbf{63.75} & \textbf{67.89} & \textbf{30.79} & \textbf{27.91} & \textbf{15.80} & \textbf{22.03} & \reldeltapctbf{43.99}{54.14} & \reldeltapctbf{9.53}{21.92} & \reldeltapctbf{26.76}{38.03} \\

\bottomrule
\end{tabular*}

\vspace{-6pt}

\caption{\textbf{Evaluation of Finetuned Agents.} Comparing baselines (Tab.~\ref{tab:experiment:training_free_main_results}) with finetuned \ours ($M_e=5$) on \ourbench (\S\ref{appendix:anabench:train_eval_data}), \textbf{\textit{relative performance differences}} (Eq.~\ref{eq:relative_delta}) are shown as \textit{positive} \textcolor{deltagreen}{$\uparrow \Delta_{\textit{rel}}\%$} or \textit{negative} \textcolor{deltared}{$\downarrow \Delta_{\textit{rel}}\%$}.}
\label{tab:experiment:finetuned_main_results}

\vspace{12pt}

\end{table*}

%% file: sections/4_exps.tex
\section{Experiments}
\label{sec:experiments}


\textbf{Model.}
Across baselines and \ours (\S\ref{appendix:anagent:variants}), we evaluate two \textit{close-source} MLLMs, \texttt{GPT-4.1-mini} \cite{openai_gpt4_1} and \texttt{Gemini-2.5-Flash} \cite{google_gemini_2_5_flash}, and six \textit{open-source} MLLMs, \texttt{InternVL-3.5} \cite{wang2025internvl3_5}, \texttt{Qwen2.5-VL} \cite{bai2025qwen2_5_vl}, and \texttt{Qwen3-VL} \cite{bai2025qwen3_vl} of different sizes.

\noindent\textbf{Data.}
To maintain computationally efficiency while reflect real-world complexity, we randomly sample from \ourbench (\S\ref{sec:anabench}) across 170 different scientific domains (Tab.~\ref{tab:data_domain}), reducing the overall training size while maintaining the real-world data distribution.

\noindent\textbf{Implementation Details.}
During SFT, agents are trained for one \textit{epoch} on the SFT training set (Tab.~\ref{tab:sft_data_size}) with initial \textit{learning rate} $1\times10^{-4}$ and \textit{cosine} scheduler.
During RL training, each agent is optimized on tailored datasets (Tab.~\ref{tab:rl_data_size}) for one \textit{epoch} with initial \textit{learning rate} $1\times10^{-6}$ to develop specialized skills.
See detailed configurations and computation overhead in Tabs.~\ref{tab:exp:computation_overhead}-\ref{tab:exp:configuration}.
%

%% file: sections/5_results.tex
\section{Results}
\label{sec:results}

\subsection{Scientific Table \& Figure Analysis}
\label{subsec:results:main_results}

\noindent\textbf{Effectiveness of Training-Free \ours.}
Tab.~\ref{tab:experiment:training_free_main_results} summarizes the performance of baselines and training-free \ours. Across eight backbone MLLMs (\S\ref{sec:experiments}), \ours consistently achieves notable gains, demonstrating the effectiveness and robustness of multi-agent collaboration.
Under zero-shot setting, \ours yields marked improvements over baselines, with $\Delta_{rel} \geq$ $4.98\%$ on $S_{\textsc{Sem}}$, $\Delta_{rel} \geq$ $8.54\%$ on $S_{\textsc{Lex}}$, and $\Delta_{rel} \geq$ $6.19\%$ on overall $S_{\textsc{Avg}}$.
%
%
%
%
%
When extending to one-shot setting, the advantages of \ours are further amplified, with relative improvements $\Delta_{rel} \geq$ $7.78\%$ on $S_{\textsc{Sem}}$, $\Delta_{rel} \geq$ $17.52\%$ on $S_{\textsc{Lex}}$, and $\Delta_{rel} \geq$ $9.79\%$ on $S_{\textsc{Avg}}$.
Results of training-free \ours demonstrate its effectiveness in consistently enhancing scientific analysis across diverse backbone models, with few-shot prompting in synergizing with \ours to further unlock its potential.

\input{figures/exp_few_shot}

\noindent\textbf{Effectiveness of Finetuned \ours.}
%
\ours consistently benefits from finetuning, with the combination of SFT+RL exhibiting the strongest optimization effects (\textit{e.g.}, \texttt{Qwen3-VL-4B}: $S_{\textsc{Avg}}$=$38.03\%$) in comparison to their SFT (\textit{e.g.}, \texttt{Qwen3-VL-4B}: $S_{\textsc{Avg}}=36.02\%$) and RL (\textit{e.g.}, \texttt{Qwen3-VL-4B}: $S_{\textsc{Avg}}=30.03\%$) counterparts, highlighting the cumulative advantages of combining SFT with RL.
On the other hand, jointly considering performance (Tab.~\ref{tab:experiment:finetuned_main_results}) and computation overhead (Tab.~\ref{tab:exp:computation_overhead}) indicates SFT provides a more favorable trade-off than RL for long-context comprehension and long-output generation (\S\ref{appendix:subsec:rl_is_a_double_edged_sword}).
Also, similar to training-free variants (Tab.~\ref{tab:experiment:training_free_main_results}), finetuned \ours presents consistent improvements from zero-shot to one-shot scientific analysis ($\Delta \geq 3.04\%$). This reveals the effectiveness of one-shot learning in multi-agent systems, where agents with specialized objectives can better coordinate and reason with minimal task-specific demonstrations.

\subsection{Test-Time Optimization}
\label{subsec:results:inference_optimization}

%
%
%
%
%
%

\noindent\textbf{Enhancing Scientific Analysis via Few-Shot Learning.}
To better accommodate heterogeneous inputs, we incorporate $k$-shot learning to improve both individual adaptability and overall coordination (\S\ref{subsec:anagent:infer}).
Compared with zero-shot generation, one-shot learning yields marked gains (Tabs.~\ref{tab:experiment:training_free_main_results}-\ref{tab:experiment:finetuned_main_results}).
Increasing the number of shots further improves performance ($k>1$), with $k=3$ achieving the highest, though relative gains diminish as $k$ grows.
Considering computational efficiency, $k=1$ provides the most favorable trade-off between performance and cost.
As shown in Fig.~\ref{fig:exp:few_shot}, few-shot learning enables \ours to more effectively leverage prior knowledge and achieve improved coordination.

\input{figures/exp_ablation_agent_level_aug}

\noindent\textbf{Enhancing Scientific Analysis via Agent-Level Capability Augmentation.}
We conduct controlled experiments in which \texttt{GPT-4.1-mini} powers \Planner, \Expert, and \Critic, with four different MLLMs instantiating \Solver, respectively.
Results in Fig.~\ref{fig:exp:agent_level_aug} show that agent-level capability augmentation (\S\ref{subsec:anagent:infer}) consistently improves the overall performance of \ours across all four \Solver backbones ($\Delta_{\textit{rel}} \geq$ $10.68\%$). Notably, augmenting only selected agents with a more capable MLLM leads to marked gains over homogeneous \ours, despite leaving \Solver unchanged.
These findings highlight the significance of agent-level capability differentiation in multi-agent systems and demonstrate that selectively augmenting critical roles, especially those tasked with global guidance and complex reasoning, can effectively enhance coordination performance.


\subsection{Ablations On \ours Variants}
\label{subsec:results:ablations_on_anagent}

\noindent\textbf{Effectiveness of Multi-Agent Scientific Analysis.}
The performance of \ours variants (Tab.~\ref{tab:anagent:variants}) varies across training-free and finetuned settings (Fig.~\ref{fig:exp:ablation_agent}).
Comparing training-free variants, \textit{Omnion} consistently underperforms baselines ($\downarrow 3.89\%$ $\leq$ $\Delta_{\textit{abs}} \leq$ $\downarrow 6.90\%$), revealing that providing a standalone \Solver with diverse tools can overwhelm reasoning and fails to enable effective scientific analysis.
\textit{Symnion} improves upon \textit{Omnion} ($\Delta_{\textit{abs}} \geq$ $\uparrow 1.39\%$) by including \Expert to assist tool invocation and context comprehension, yielding performance that is generally above baselines but remains unstable and occasionally inferior. This unveils the key insight that the absence of global planning can lead to suboptimal coordination and misleading intermediate decisions.
In contrast, by integrating high-level planning, interactive executing, context-aware problem-solving, and reflective refinement (\S\ref{sec:anagent}), \ours consistently achieves the highest performance.
Among all variants, finetuning leads to marked gains over training-free counterparts ($\Delta_{\textit{abs}} \geq$ $\uparrow 3.12\%$), even finetuned \textit{Omnion} surpassing baselines, highlighting the importance of targeted finetuning in optimizing multi-agent coordination.

\noindent\textbf{Effectiveness of Critic-Guided Optimization.}
Comparing \ours with and without \Critic reveals contrasting effects (Fig.~\ref{fig:exp:ablation_agent}).
For training-free \ours, incorporating \Critic can degrade collaborative performance for smalls-size MLLM agents as a result of their ineffective reflection. 
This effect is different for more capable agents, unveiling the limited reasoning and reflection abilities of smaller MLLM agents.
In contrast, finetuned \Critic is able to more accurately assess intermediate solutions and identify key deficiencies, guiding effective refinements to improve overall performance.
These findings underscore both the challenges and the significance of equipping agentic systems with robust reflection and refinement abilities in tackling complex problems.

\subsection{In-Depth Analysis}
\label{subsec:results:ablations_on_params}

\noindent\textbf{Validation via MLLM-As-Judge \& Case Studies.}
Tab.~\ref{tab:experiment:mllm_as_judge_main_results} showcases consistent performance gains across six backbone MLLMs (\S\ref{appendix:extended_experiment_analysis:mllm_as_judge}), with overall $S_{\textsc{Mllm}}$ achieving up to $\Delta_{rel}=29.24\%$. This validates our design of multi-metric evaluation (\S\ref{subsec:anabench:evaluation_metrics}).
Through dedicated case studies (\S\ref{appendix:sec:failure_analysis}) on seven error patterns (Fig.~\ref{fig:preliminary-error-analysis}), Fig.~\ref{fig:exp:failure_analysis:error_distribution} reveals substantial reductions across all error types, indicating the effectiveness of \ours in advancing scientific reasoning \& understanding across seven complexity dimensions (\S\ref{fig:preliminary}).

\noindent\textbf{Modular Training Is Better Than End-to-End Training For Multi-Agent Optimization.}
Comparing modular training (\S\ref{subsec:anagent:train}) with end-to-end training, we evaluate their impact on multi-agent collaboration.
As shown in Tab.~\ref{tab:experiment:ablation_training_method_end2end_modular}, modular training consistently outperforms end-to-end training across all metrics with notable gains ($\Delta_{\textit{rel}} \geq$ $\uparrow33.10\%$). These results reveal that modular training more effectively supports coordinated behaviors and leads to stronger overall performance. In contrast, end-to-end training markedly constrains agents from developing and preserving specialized capabilities for designated roles. For example, in some cases, \Planner directly generates final solutions during the planning stage (\S\ref{appendix:subsec:failure_analysis:error8}), significantly undermining role specialization and leading to degraded performance ($\Delta_{\textit{abs}} \geq$ $\downarrow2.68\%$). This loss of specialization ultimately hampers collaborative effectiveness, highlighting the significance of modular optimization in multi-agent systems.


\input{tables/exp_ablation_train_method_end2end_vs_modular}

\noindent\textbf{Unpacking the Training Data Recipe For Multi-Agent Finetuning.}
To understand how training data affects multi-agent finetuning, we conduct ablation studies along four dimensions of the training set (Fig.~\ref{fig:exp:ablation_data_size_domain_format_type}).
As shown in Fig.~\ref{fig:exp:ablation_data_size_domain_format_type}(a), training 30K subset consistently underperforms training on the full set ($\Delta_{\textit{rel}} \geq$ $\downarrow10.27\%$), unveiling the benefits of larger-scale training data.
Fig.~\ref{fig:exp:ablation_data_size_domain_format_type}(b) compares domain-specific training with training on nine-domain full set. Domain-specific learning results in pronounced performance degradation ($\Delta_{\textit{rel}} \geq$ $\downarrow26.55\%$), revealing that restricting training domains significantly limits agents' generalizability to out-of-domain tasks.
Fig.~\ref{fig:exp:ablation_data_size_domain_format_type}(c) illustrates that single-format training impairs cross-format generalization, leading to consistent performance drops ($\Delta_{\textit{rel}} \geq \downarrow8.85\%$).
Fig.~\ref{fig:exp:ablation_data_size_domain_format_type}(d) further demonstrates that limiting training to a single data type markedly degrades performance ($\Delta_{\textit{rel}} \geq \downarrow9.53\%$).
We extend our discussion in 
\S\ref{appendix:subsec:extended_analysis_on_training_data}.

\noindent\textbf{Tools Are The Key To Open The Door of Good Scientific Analysis.}
Tools play a pivotal role in enabling high-quality scientific analysis by exposing \ours to extended knowledge and context.
Figs.~\ref{fig:exp:ablation_tool_call}-\ref{fig:exp:ablation_tool_strategy} demonstrate that performance gains arise not merely from the availability of tools, but from their strategic and objective-aligned utilization.
When tool functionalities are accurately matched to task demands, \ours is able to effectively retrieve relevant context and domain knowledge, ground reasoning in additional evidence, and adapt analysis to task-specific scientific scenarios.
We extend our discussion in \S~\ref{appendix:subsec:tool_use}.

%% file: figures/exp_few_shot.tex
\begin{figure}[H]

\vspace{0pt}

\centering
\includegraphics[width=1.0\linewidth]{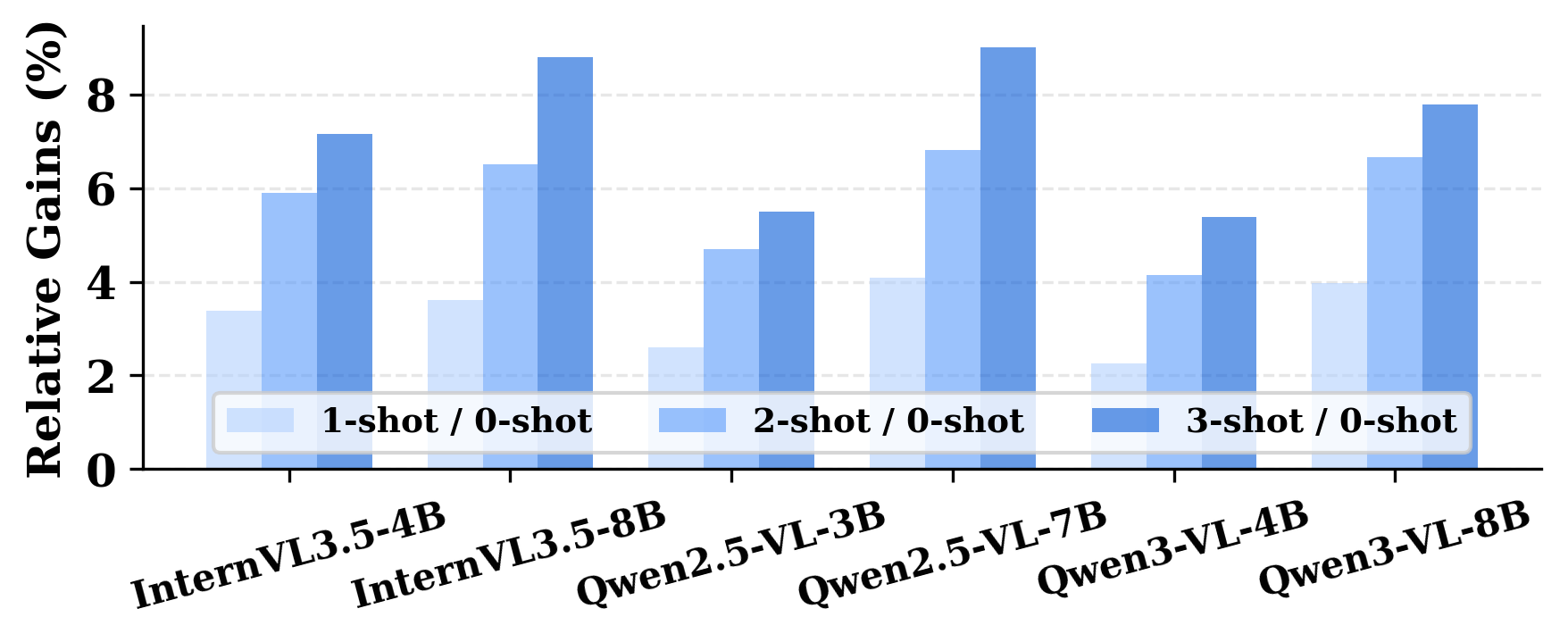}

\vspace{-3pt}

\caption{\textbf{Few-Shot Learning Optimization (\S\ref{subsec:anagent:infer})}}
\label{fig:exp:few_shot}

\vspace{0pt}

\end{figure}

%% file: figures/exp_ablation_agent_level_aug.tex
\begin{figure}[H]

\vspace{0pt}

\centering
\includegraphics[width=1.0\linewidth]{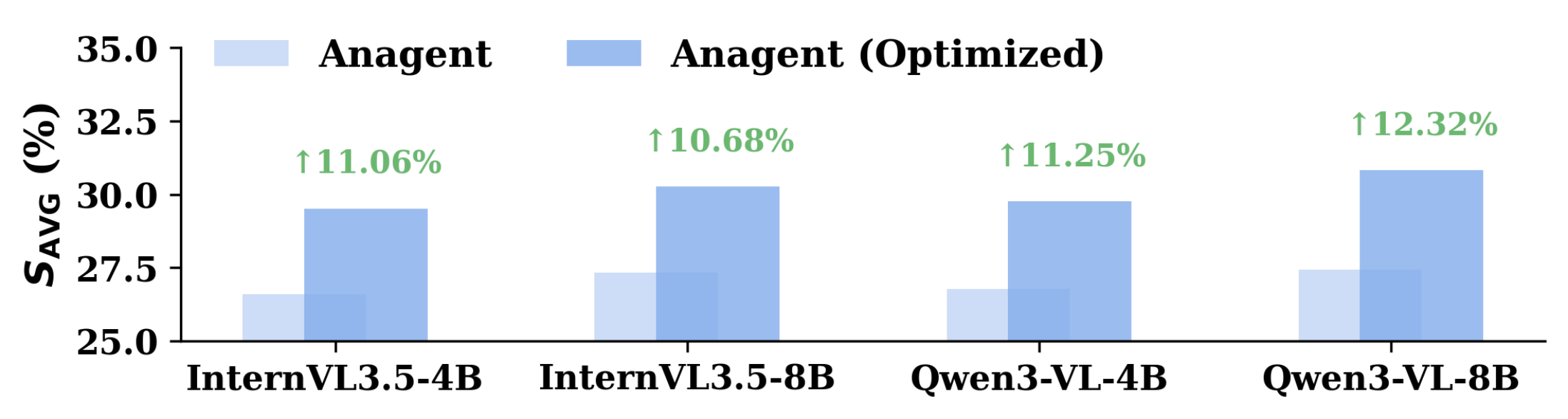}

\vspace{-3pt}

\caption{\textbf{Agent-Level Ability Augmentation (\S\ref{subsec:anagent:infer})}}
\label{fig:exp:agent_level_aug}

\vspace{0pt}

\end{figure}

%% file: tables/exp_ablation_train_method_end2end_vs_modular.tex
\begin{table}[t]

\vspace{0pt}

\small
\centering
\renewcommand{\arraystretch}{1.0}
\setlength{\tabcolsep}{6pt}

\begin{tabular*}{\columnwidth}{@{\extracolsep{\fill}}l c ccc}
\toprule

\textbf{Model} & \textbf{Size} & $\bm{S}_{\textbf{\textsc{Sem}}}$ & $\bm{S}_{\textbf{\textsc{Lex}}}$ & $\bm{S}_{\textbf{\textsc{Avg}}}$ \\

\midrule

\multicolumn{5}{c}{\cellcolor{oursbaselinebg}\textbf{\ours (Training-Free)}} \\
\addlinespace[3pt]

\textbf{Qwen2.5-VL} & \textbf{3B} & 46.18 & 10.91 & \reldeltapct{26.59}{28.55} \\
\textbf{Qwen3-VL} & \textbf{4B} & 46.95 & 10.50 & \reldeltapct{26.76}{28.73} \\

\multicolumn{5}{c}{\cellcolor{ourstrainedbg}\textbf{\ours (End-to-End)}} \\
\addlinespace[3pt]

\textbf{Qwen2.5-VL} & \textbf{3B} & 48.32 & 14.56 & \reldeltapct{26.59}{31.44} \\
\textbf{Qwen3-VL} & \textbf{4B} & 49.17 & 14.87 & \reldeltapct{26.76}{32.02} \\

\multicolumn{5}{c}{\cellcolor{ourstrainedbg}\textbf{\ours (Modular)}} \\
\addlinespace[3pt]

\textbf{Qwen2.5-VL} & \textbf{3B} & 51.91 & 18.86 & \reldeltapct{26.59}{35.39} \\
\textbf{Qwen3-VL} & \textbf{4B} & \textbf{53.08} & \textbf{20.81} & \reldeltapctbf{26.76}{36.95} \\

\bottomrule
\end{tabular*}

\vspace{-3pt}

\caption{\textbf{End-to-End Training vs. Modular Training.}}
\label{tab:experiment:ablation_training_method_end2end_modular}

\vspace{6pt}

\end{table}

%% file: sections/7_conclusion.tex
\section{Conclusion}
\label{sec:conclusion}

In advancing real-world scientific table \& figure analysis, we introduce \ourbench, a large-scale benchmark of $63,178$ instances built upon real-world scientific publications. By covering nine domains and seven complexity dimensions, \ourbench casts scientific table \& figure analysis as open-ended analysis writing. Building on \ourbench, we propose \ours, a multi-agent framework that improves analysis quality by up to $\uparrow 13.43\%$ with test-time optimization and $\uparrow 42.12\%$ with modular training. We hope \ourbench and \ours meaningfully support future research on agentic reasoning and scientific analysis.

%% file: sections/acknowledge.tex
\section*{Use of AI}
\label{sec:use_of_ai}

We used AI assistants only to polish writing and correct grammar. All research questions, methodologies, experimental design, analyses, and conclusions are the authors' own. This is separate from our methodological and experimental use of MLLMs within \ourbench and \ours, which is documented in the corresponding sections.

\section*{Limitations}
\label{sec:conclusion}

In this work, we address scientific table \& figure analysis by proposing (1) \ourbench (\S\ref{sec:anabench}), a benchmark with $63,178$ instances along seven complexity dimensions (Fig.~\ref{fig:preliminary}), and (2) \ours (\S\ref{sec:anagent}), a multi-agent system for enhanced scientific table \& figure analysis.
Through test-time optimization (\S\ref{subsec:anagent:infer}) and modular training (\S\ref{subsec:anagent:train}), \ours achieves substantial improvements on \ourbench (\S\ref{sec:results}), revealing the effectiveness of task-oriented decomposition, strategic knowledge retrieval, and context-aware problem-solving in tackling complex scientific problems.
Nevertheless, we acknowledge our limitations in four aspects, including computation overhead, domain coverage, tool dependency, and evaluation (\S\ref{sec:limitations}).
We hope \ourbench and \ours provide meaningful foundations to facilitate future research.

\section*{Acknowledgments}
The authors acknowledge William \& Mary Research Computing for providing computational resources and/or technical support that have contributed to the results reported within this paper. This work is supported by William \& Mary Applied Research \& Innovation Initiative. This work used DeltaAI at NCSA through allocation CIS260012 and CIS230280 from the Advanced Cyberinfrastructure Coordination Ecosystem: Services \& Support (ACCESS) program, which is supported by U.S. National Science Foundation grants \#2138259, \#2138286, \#2138307, \#2137603, and \#2138296. This material is based upon work supported by the Google Cloud Research Credits program with the award GCP19980904. This research was supported by the Intramural Research Program of the National Institutes of Health (NIH). The contributions of the NIH author(s) are considered Works of the United States Government. The findings and conclusions presented in this paper are those of the author(s) and do not necessarily reflect the views of the NIH or the U.S. Department of Health and Human Services.

%% file: sections/appendix.tex

\section{Preliminary Exploration}
\label{appendix:preliminary}

\subsection{What Challenges Exist In Scientific Table \& Figure Analysis?}
\label{appendix:preliminary:challenges}

The heterogeneity of \textit{table \& figure scientific analysis} poses critical challenges for MLLM agents in accurately understanding various modalities, structures, formats, contexts, domains, and writing demands.
To investigate how MLLM agents can perform in tackling these challenges, we employ Qwen3-VL-8B as the base model, generating analysis on 60 scientific tables and figures, respectively. These 120 data are randomly sampled from \ourbench with evenly distributed features across both \textbf{data complexity} (including \textit{data type}, \textit{data format}, \textit{data source}, and \textit{data domain}) and \textbf{analysis complexity} (including \textit{analysis objective}, \textit{analysis depth}, and \textit{analysis width}) (\S\ref{subsec:anabench:benchmark_construction}).

As shown in Fig.~\ref{fig:preliminary}, the agent presents varying performance on these dimensions, showcasing strengths on table structures over visual figures, XML formats over LaTeX, arXiv \cite{arxiv} papers over PubMed \cite{pubmed}, computer science related domains over biomedicine, experimental analysis over methodology, superficial summarization over in-depth analysis, and fully-grounded analysis over inferential writing.
Analyzing case-by-case, we observe seven distinct patterns (Fig.~\ref{fig:preliminary-error-analysis}) in agent's analysis failures, which are in close accordance with the main challenges we conclude in Fig.~\ref{fig:preliminary}.

Employing Qwen3-VL-8B as the analysis agent baseline, the agent performs scientific analysis on different data types, formats, sources, and domains (Figs.~\ref{fig:preliminary-error-analysis}-\ref{fig:preliminary-pattern-2-domain}).
As shown in Fig.~\ref{fig:preliminary-pattern-1-type}, the agent analyzes two distinct data types, \textit{table} and \textit{figure}, respectively. While the table is a text-only single modality input, it presents different structures than writing or drawing. On the other hand, the figure gives vision-language input, complicating the multimodal understanding that serves as the basis of accurate analysis. Compared with ground-truth analysis, the agent's outputs reveal their significant visual perception errors, together with their lack of scientific writing abilities with proper analysis depth and width.
Fig.~\ref{fig:preliminary-pattern-2-domain} shows two examples with different data types, sources, formats, and domains. As domain-specific knowledge further complicates the tasks, the agent analyzes both table and figure with notable hallucinated contents. Meanwhile, the agent's analysis contains significant misinterpretation on domain-specific terminologies and expressions, and lacks wider analysis coverage and deeper discussions.

\input{figures/preliminary-error-analysis}
\input{figures/preliminary-pattern-1-type}

\input{figures/preliminary-pattern-2-domain}

\subsection{How To Enhance Scientific Table \& Figure Analysis?}
\label{appendix:preliminary:method}

\paragraph{Benchmarking Scientific Analysis Under Realistic Challenges.}
Our preliminary exploration (\S\ref{appendix:preliminary:challenges}) reveals that failures in scientific table \& figure analysis are not stemming from isolated weaknesses, but rather from the compound difficulty introduced by heterogeneous data representations and diverse analytical demands (Fig.~\ref{fig:preliminary}).
To this end, we propose \ourbench (\S\ref{sec:anabench}), a benchmark that systematically instantiates these challenges based on real-world scientific publications. By covering diverse data modalities and structures, types and formats, sources and domains, \ourbench develops scientific analysis tasks with varying analysis widths, depths, and objectives, enabling controlled evaluation of MLLM agents' scientific analysis capabilities across various data and analysis complexities.
Aligning benchmark construction (\S\ref{subsec:anabench:benchmark_construction}) directly with the observed challenges (Fig.~\ref{fig:preliminary}), \ourbench establishes a principled foundation for evaluating and advancing scientific table \& figure analysis.

\input{figures/teaser}

\paragraph{Enhancing Scientific Analysis Through Multi-Agent Collaboration.}
The identified error patterns present across different stages of scientific writing, from input interpretation and contextual understanding to analysis writing and reflective correction. 
As such, a high-quality scientific analysis demands deliberate planning, accurate and domain-aware information acquisition, coherent synthesis, and careful verification, which require capabilities that are difficult to reliably achieve within a monolithic agent (\S\ref{subsec:results:ablations_on_anagent}).
\textit{How to enhance these specialized capabilities in one agentic framework?}
By decomposing the scientific analysis writing process into multiple stages, we introduce \ours, a multi-agent framework capable of collaborative scientific analysis with specialized completion and optimization.
Specifically, \ours consists of four specialized agents:
\Planner for high-level task planning, alleviating potential errors in analysis width, depth, and objectives;
\Expert for task-oriented exploration and retrieval, avoiding domain-specific and contextual understanding errors across varied modalities, structures, and formats;
\Solver for scientific analysis writing, reasoning through all the available knowledge and information as guided and supported by \Planner and \Expert;
and \Critic for self- reflection and correction to rectify inaccurate analyses and hallucinated contents.
Collectively, \ours targets the key failure patterns observed in Fig.~\ref{fig:preliminary-error-analysis} to enhance agent's scientific analysis capabilities for improved overall performance.

\paragraph{Evaluating Scientific Analysis With Multi-Dimensional Assessment.}
Given the diversity of challenges (Fig.~\ref{fig:preliminary}) and error patterns (Fig.~\ref{fig:preliminary-error-analysis}) involved in scientific analysis, single-scalar judgment is insufficient to reflect analysis quality. Accurate scientific analysis requires faithful interpretation of presented data, comprehensive coverage of key findings, adherence to task-specific requirements, clear and coherent scientific writing, and strict grounding in available evidence. Accordingly, our assessment considers these aspects jointly, capturing both analytical correctness and writing quality through the five-dimensional evaluation protocol (Fig.~\ref{fig:five_dimension_evaluation_prompt}): \textit{content accuracy}, \textit{analytical completeness}, \textit{format correctness}, \textit{clarity \& coherence}, and \textit{reliability \& faithfulness}.
To complement our rule-based assessment, we apply \textit{five-dimensional protocol} to both MLLM-As-Judge and human expert assessment (\S\ref{subsec:anabench:evaluation_metrics}), enabling fine-grained comparisons across models and settings.

\input{sections/6_related}

\section{\ourbench: Benchmark Analysis}
\label{appendix:anabench}

We construct \ourbench that covers seven key challenges (Fig.~\ref{fig:preliminary}), with our construction method scalable to different sizes for custom use (Fig.~\ref{fig:anabench}).
By developing an automated multi-stage benchmark construction method (\S\ref{fig:anabench}), \ourbench captures a wide range of data complexity (\S\ref{appendix:benchmark:curriculum:data_complexity}) and analysis complexity (\S\ref{appendix:benchmark:curriculum:analysis_complexity}), enabling more comprehensive evaluation of scientific analysis. Our multi-level filtering and quality-control procedures further ensure high data reliability.
The comparison between \ourbench and recent scientific benchmarks is summarized in Tab.~\ref{tab:benchmark_comparison}.

\input{figures/exp_ablation_train_data_size_domain_format_type}

\subsection{Benchmark Construction}
\label{appendix:anabench:construction}

As illustrated in Fig.~\ref{fig:anabench}, our dataset construction method comprises four progressive automated stages: (1) \textit{source collection}, (2) \textit{data extraction}, (3) \textit{instance construction}, and (4) \textit{task classification}. To ensure data quality, we implement multi-level filtering across stages, from \textit{source collection} to \textit{instance construction}.
Here, we elaborate on our benchmark construction in further detail to complement \ourbench:

\paragraph{Source Collection.}
During the initial stage of \textit{source collection}, we gather source papers from multiple dissemination platforms and apply a combination of paper-level filters, including domain-category filtering, publication-year filtering, keyword-based filtering, full-text access filtering, and maximum-source thresholding.
In particular, to mitigate the risk of data contamination during model pretraining, we restrict sources to papers published after 2023. Moreover, to ensure data quality and better coverage of recent work, we set the maximum source threshold for papers published in or after 2025 to be twice that of papers published before 2025.

\paragraph{Data Extraction.}
In the second stage of \textit{data extraction}, we perform both \textit{paper-level} and \textit{data-level} filtering based on automated data parsing. Specifically, we filter out papers and data instances that exhibit access failures or parsing errors. For each retained figure or table data, we extract the parsed data content along with the associated source files when available (\textit{e.g.}, PNG images).
In addition, we extract contextual information for each targeted data through $d$-depth hierarchical intra-document and inter-document reference retrieval.
Our $d$-depth hierarchical context retrieval method (Alg.~\ref{alg:k_depth_context}) is implemented in a recursive manner: the first-level context consists of elements that the target instance refers to or is referred to by; the second-level context includes elements that the first-level contexts refer to or are referred to by; and this process continues iteratively up to depth $d$. This hierarchical context retrieval enables the extraction of both internal and external relational information surrounding each data sample.

\paragraph{Gold Analysis.}
Every gold analysis in \ourbench is author-written text from the peer-reviewed source paper (Fig.~\ref{fig:anabench}): for each target table or figure, we extract verbatim the analytical prose in which the original authors discuss it. No reference analysis is model-generated or model-edited. Also, contexts are retrieved separately through $d$-depth hierarchical reference retrieval (Alg.~\ref{alg:k_depth_context}), so $d$ controls only the context scope supplied to a model. MLLM assistance in constructing \ourbench is confined merely to labeling two of the seven complexity dimensions, \emph{analysis objective} and \emph{analysis depth} (\S\ref{appendix:benchmark:curriculum:task_classification}); the other five and all filtering are rule-based (\S\ref{appendix:anabench:construction}, \S\ref{appendix:anabench:train_eval_data}). Domain experts rate the gold analyses at $93.78\%$ under the same five-dimensional protocol applied to model outputs (\S\ref{appendix:extended_experiment_analysis:human_domain_evaluation}).

\paragraph{Instance Construction.}
Supported by the prior two stages, the \textit{instance construction} stage integrates the targeted data, $d$-depth contexts, ground-truth analysis, and source metadata to create each scientific analysis instance.
This stage performs multi-level \textit{data cleaning}, including data filtering that excludes embedded elements, threshold-based filtering that removes instances with over-short or over-long inputs and outputs according to the predefined thresholds, and data validation that discards data with missing targeted samples or ground-truth analyses.
The resulting cleaned instances are then stored in \ourbench for subsequent task classification.

\input{figures/anabench_year_distribution}

\paragraph{Task Classification.}
We combine \textit{rule-based} task classification with \textit{MLLM-assisted} curriculum categorization to classify scientific analysis instances into fine-grained curriculum categories across seven complexity dimensions (\S\ref{appendix:benchmark:curriculum:task_classification}).
We summarize the complexity curriculum categories in Tab.~\ref{tab:anabench_complexity_curriculum}, with 23 task complexity categories across four data complexity dimensions (\S\ref{appendix:benchmark:curriculum:data_complexity}) and three analysis complexity dimensions (\S\ref{appendix:benchmark:curriculum:analysis_complexity}).

\paragraph{Quality Control.}
To ensure data quality, we implement multi-level filtering and data cleaning across different benchmark construction stages (\S\ref{appendix:anabench}).
Furthermore, to mitigate the risk of data contamination during model pretraining, we restrict paper sources to those published after 2023 at the initial source-collection stage of our benchmark construction (Fig.~\ref{fig:anabench}), with 2025 accounting for the majority of instances to mitigate data contamination (Fig.~\ref{fig:anabench_year_distribution}).
Accordingly, our evaluation set (\S\ref{appendix:anabench:train_eval_data}) is obtained by filtering \ourbench to instances derived from papers published in 2025 and then downsampling this subset.

\input{tables/algorithm1}

\subsection{Benchmark Curriculum}
\label{appendix:benchmark:curriculum}

According to the difficulty and diversity of the data, \ourbench is organized into a curriculum along two overarching dimensions (\S\ref{subsec:anabench:benchmark_construction}), \textbf{data complexity} (\S\ref{appendix:benchmark:curriculum:data_complexity}) and \textbf{analysis complexity} (\S\ref{appendix:benchmark:curriculum:analysis_complexity}), to capture and reflect real-world variations in both scientific inputs and analytical demands, enabling systematic evaluation across heterogeneous scenarios.
To determine benchmark curriculum, we perform fine-grained task classification across data and analysis complexities (\S\ref{appendix:benchmark:curriculum:task_classification}).

\subsubsection{Data Complexity}
\label{appendix:benchmark:curriculum:data_complexity}

\paragraph{Data Type.}
\ourbench covers different data modalities and structures commonly encountered in scientific literature. Specifically, the input data include structured \textit{tables} that present single-modality data with explicit tabular organization, and \textit{figures} that are inherently multimodal and consist of both visual and textual elements.
The \textit{table} category includes diverse tabular organizations with varying layouts, levels of sparsity, and semantic density, requiring structured parsing and relational reasoning.
On the other hand, the \textit{figure} category spans a wide range of visual structures, such as charts, plots, frameworks, diagrams, \textit{etc}., introducing additional challenges in visual interpretation and cross-modal alignment between textual and visual elements.

\paragraph{Data Format.}
To reflect the real-world diversity of scientific document representations, the input data are supplied in both \texttt{LaTeX} and \texttt{XML} formats. These formats differ substantially in syntactic structure and parsing complexity, requiring models to handle distinct markup conventions while preserving the underlying semantic content.

\paragraph{Data Source.}
The benchmark incorporates data collected from publications across not only dissimilar literature categories, including \textit{general} papers and \textit{review or survey} papers, but also different dissemination platforms, such as \textit{arXiv} \cite{arxiv} and \textit{PubMed} \cite{pubmed}.
These sources vary in writing structures, submission formats, and disciplinary emphasis, contributing to increased heterogeneity in data and domain distributions.

\paragraph{Data Domain.}
\ourbench spans 9 broad scientific domains, covering 170 fine-grained disciplines (Tab.~\ref{tab:data_domain} \& Figs.~\ref{fig:anabench_domain_distribution}-\ref{fig:test_domain_distribution}).
This domain diversity supports \ourbench to systematically evaluate the analytical capabilities of MLLM agents across varied domain-specific knowledge, terminologies, methodological conventions, and writing norms.

\input{tables/benchmark_comparison}
\input{tables/benchmark_complexity_curriculum}
\input{figures/anabench_distribution}
\input{figures/anabench_statistics}
\input{tables/data_domain}
\input{tables/data_size}

\subsubsection{Analysis Complexity}
\label{appendix:benchmark:curriculum:analysis_complexity}

\paragraph{Analysis Objective.}
Data in \ourbench are classified according to their analytical objectives. Specifically, each analysis is labeled as either (1) \textit{methodology-oriented analysis} that describes methodological designs, theoretical formulations, algorithmic principles, and implementation of methods, models, or experiments; or (2) \textit{experimental analysis} that interprets empirical results, identifies patterns or trends, and draws evidence-based conclusions.
This distinction reflects the diverse objectives of scientific reasoning involved in research analysis.

\paragraph{Analysis Width.}
Analysis width characterizes the scope of information referenced in the ground-truth analysis. We define four fine-grained classes: (1) analyses with \textit{no references}, which rely solely on the immediate inputs;
(2) \textit{internal references}, which draw upon other components within the same document;
(3) \textit{external references}, which incorporate information beyond the current document;
and (4) \textit{mixed references}, which combine both internal and external sources. This analysis width reflects the increasing breadth of contextual integration required for comprehensive analysis.

\paragraph{Analysis Depth.}
Analysis depth distinguishes between surface-level summarization and inference-driven analysis. \textit{Shallow} analyses involve direct restatement or aggregation of explicitly stated information, whereas in-depth analyses require implicit reasoning, interpretation, or synthesis that is not directly observable from the input. This analysis depth captures the degree of cognitive and analytical complexity demanded by each task.

\subsubsection{Task Curriculum}
\label{appendix:benchmark:curriculum:task_classification}

Of the seven complexity dimensions (\S\ref{subsec:anabench:benchmark_construction}), five are assigned by deterministic rule-based classification over the parsed document structure, and only two, \textit{analysis objective} and \textit{analysis depth}, require MLLM classification (\S\ref{appendix:anabench:construction}).
Employing Gemini-2.5-Flash \cite{google_gemini_2_5_flash} for \textit{MLLM-assisted classification} and conceptual criteria for \textit{rule-based classification}, we perform fine-grained task curriculum classification (Tab.~\ref{tab:anabench_complexity_curriculum}) on \ourbench according to the data complexity and analysis complexity of each instance:

\begin{itemize}[leftmargin=*, topsep=0pt, itemsep=1pt, parsep=1pt]

\item \textbf{Data Type:} Rule-based classification, based on the data type of the input data (\S\ref{appendix:benchmark:curriculum:data_complexity}).

\item \textbf{Data Format:} Rule-based classification, based on the data format of the input data(\S\ref{appendix:benchmark:curriculum:data_complexity}).

\item \textbf{Data Source:} Rule-based classification, based on the data source of the input data (\S\ref{appendix:benchmark:curriculum:data_complexity}).

\item \textbf{Data Domain:} Rule-based classification, based on the domain of the task (\S\ref{appendix:benchmark:curriculum:data_complexity}).

\item \textbf{Analysis Objective:} MLLM classification, according to the analysis objective of the task (\S\ref{appendix:benchmark:curriculum:analysis_complexity}).

\item \textbf{Analysis Width:} Rule-based classification, according to the references and citations included in the task (\S\ref{appendix:benchmark:curriculum:analysis_complexity}).

\item \textbf{Analysis Depth:} MLLM classification, according to the analysis level of the task (\S\ref{appendix:benchmark:curriculum:analysis_complexity}).

\end{itemize}

\input{figures/anabench_distribution_test}

\subsection{Benchmark Statistics}
\label{appendix:benchmark:statistics}

Following the major challenges (Fig.~\ref{fig:preliminary}) and failure errors (Fig.~\ref{fig:preliminary-error-analysis}) identified through preliminary exploration (\S\ref{appendix:preliminary}), we construct \ourbench across different data types, data formats, data sources, and data domains, thereby contributing to different data and analysis complexities (\S\ref{appendix:benchmark:curriculum}).

Starting from 9 broad scientific domains, we systematically delve into each domain to compile scientific analysis samples across 170 fine-grained subdomains (Tab.~\ref{tab:data_domain}). Additionally, we define the task curriculum of \ourbench according to the seven complexity levels of each task (\S\ref{appendix:benchmark:curriculum:task_classification}).

Through quantitative analysis, we summarize the statistics of our benchmark in Fig.\ref{fig:anabench_statistics}. Our dataset construction follows real-world distributions of data types, formats, and domains, preserving both natural data distribution imbalance and inherent complexity curriculum (Fig.~\ref{fig:anabench} \& \S\ref{sec:anabench}).

\subsection{Data Preprocessing}
\label{appendix:anabench:train_eval_data}

For training and evaluation, we implement additional filtering and downsampling to ensure both data quality and computation efficiency.

\input{figures/exp_ablation_agent}

\paragraph{Data Filtering.}
We add the first length filtering step to exclude samples with overly short ground-truth analysis to ensure the effectiveness of model learning, and then apply the second filtering step to filter out samples with overlong contexts or overlong analysis to preserve efficient training with affordable computational resources.

\paragraph{Data Downsampling.}
The large computation resources required by RL are intensified by the long-context inputs and long-analysis output. To this end, we further downsample a subset of the filtered dataset through random sampling with more narrowed thresholds.

\paragraph{Training Data.}
During the SFT stage, we leverage the filtered dataset with 31,350 samples. To explore the effects of data size, we extend our ablation studies to cover several variations of data sizes (Tab.~\ref{tab:sft_data_size}).
During RL optimization, we further downsample a smaller subset tailored for each agent's sub-goals (\S\ref{subsec:anagent:train} \& \ref{appendix:anagent:rl_training}).

\paragraph{Evaluation Data.}
Aiming for comprehensive evaluation, we randomly downsample the test set from \ourbench in a dimension-wise manner to cover all challenge dimensions (Fig.~\ref{fig:preliminary}).
As such, our test set (Fig.~\ref{fig:test_data_distribution}) consists of 7,319 test samples across different data types, formats, modalities, sources, and domains, with tasks in varying analysis depths, widths, and writing categories.
To mitigate data contamination (\S\ref{appendix:anabench}), our evaluation set consists of instances derived exclusively from source papers published in 2025 (Fig.~\ref{fig:anabench_year_distribution}).
We visualize the domain distribution of the evaluation dataset in Fig.~\ref{fig:test_domain_distribution}.

\section{\ours: Multi-Agent Collaborative Scientific Analysis}
\label{appendix:anagent}

\subsection{\ours Variants}
\label{appendix:anagent:variants}

We summarize five variants of \ours in Tab.~\ref{tab:anagent:variants}, which differ in their agent components and tool availability. To investigate the contribution of these design choices to scientific analysis generation, we conduct additional ablation studies using the same evaluation dataset and metrics (\S\ref{appendix:evaluation}).
As shown in Fig.~\ref{fig:exp:ablation_agent}, agents' performance exhibits a clear ordering: \ours consistently outperforms \textit{Symnion} variants, which in turn outperform \textit{Omnion} variants.
In the training-free setting, \textit{Omnion} underperforms all baselines, indicating that equipping a standalone \Solver with diverse tools is insufficient and can even overwhelm the agent, preventing effective interaction with task contexts and environments.
\textit{Symnion} alleviates this issue to some extent by introducing \Expert to assist with tool invocation and contextual interaction. While this generally improves performance over baselines, results remain unstable and occasionally fall below baseline levels, suggesting that tool orchestration alone is not enough for robust scientific reasoning.
In contrast, \ours integrates \Planner for high-level task decomposition, \Expert for contextual and domain-specific information retrieval, \Solver for context-aware problem solving, and \Critic for reflection and refinement. This structured multi-agent design leads to consistently superior performance across all MLLM backbones, demonstrating the significance of explicit planning, contextual grounding, and context-aware problem-solving in complex scientific analysis.
Comparing training-free and finetuned variants, all finetuned models achieve substantial gains over their training-free counterparts. Notably, even finetuned \textit{Omnion} surpasses baseline methods, highlighting the critical role of targeted finetuning in enabling agents to effectively leverage tools and interact with scientific environments.
Overall, these results underscore the complementary benefits of principled multi-agent architecture design and task-specific finetuning for reliable and high-performing scientific reasoning systems.


\input{tables/anagent_variants}

\subsection{Multi-Agent Collaboration}
\label{appendix:anagent:collaboration}

Implementing collaborative scientific analysis, \ours decomposes the end-to-end reasoning and writing process into four specialized agents with complementary roles (\S\ref{sec:anagent}).
As inspired by our preliminary exploration (\S\ref{appendix:preliminary:method}), rather than relying on a single monolithic agent to simultaneously plan, retrieve, reason, and reflect, our design explicitly separates these responsibilities to reduce error propagation and encourage iterative refinement (\S\ref{sec:intro}).
Accordingly, each agent is guided by a tailored prompt that defines its task, objective, and interaction protocol, enabling structured collaboration across different stages of scientific analysis (\S\ref{subsec:anagent:anagent}).
Fig.~\ref{fig:anagent} illustrates our proposed multi-agent scientific analysis workflow (with example data from \cite{guo2025syncmind}), as motivated by Fig.~\ref{fig:teaser} (with example data from \cite{wang2025papo}).

\paragraph{\Planner} is responsible for high-level task decomposition and strategic guidance. Given a scientific analysis task, \Planner identifies the core objectives, determines the required analytical depth and breadth, and outlines a step-by-step plan to guide downstream agents.
By explicitly structuring the reasoning process before content generation, \Planner mitigates common failures such as incomplete coverage, misaligned objectives, and shallow analysis.
The task prompt for \Planner is shown in Fig.~\ref{fig:planner_prompt}.

\paragraph{\Expert} focuses on task-oriented exploration, retrieval, and domain-specific clarification.
Under the guidance of \Planner, \Expert gathers relevant contextual knowledge, resolves ambiguities in terminology or context, and provides structured evidence or references across diverse modalities, formats, and domains.
This separation allows \ours to better handle domain-specific nuances and reduces errors arising from insufficient or incorrect contextual understanding.
The detailed task prompt for \Expert is presented in Fig.~\ref{fig:expert_prompt}.

\paragraph{\Solver} performs the core scientific analysis writing.
Guided by the problem-solving plan generated by \Planner and the supporting information provided by \Expert, \Solver synthesizes semantically coherent, logically structured, and scientifically grounded analysis.
\Solver's prompt (Fig.~\ref{fig:solver_prompt}) emphasizes systematic integration of provided information and rigorous scientific reasoning, maintaining close alignment between retrieved knowledge and analysis objectives.

\paragraph{\Critic} conducts self-reflection and post-hoc verification of the generated analysis.
By systematically reviewing the \Solver's solution, it identifies overlooked observations and findings, logical inconsistencies and reasoning flaws, groundless claims and hallucinated contents, as well as formatting errors and analysis inaccuracies, proposing targeted revisions for enhanced scientific analysis.
By explicitly modeling critique and reflection as an explicit step, \ours improves analytical reliability and scientific rigor.
By applying the five-dimensional evaluation protocol (\S\ref{appendix:preliminary:method} \& \ref{appendix:evaluation:five_dimension}),
\Critic's prompt defines both evaluation criteria and critique objectives for high-quality scientific analysis (Fig.~\ref{fig:critic_prompt}).

\subsection{Scientific Toolkits}
\label{appendix:anagent:tool}

In scientific research, human researchers rely on a diverse set of skills to observe, analyze, and reason over complex scientific materials. These skills include \textit{reading and comprehending scientific documents}, \textit{retrieving targeted knowledge}, \textit{searching for related literature and information}, \textit{analyzing multimodal data with dissimilar structures and formats}, and \textit{performing interactive or computational explorations}.
To develop AI agents into AI scientists, we take inspiration from human scientific research, equipping AI agents with five specialized toolkits (Tab.~\ref{tab:five_toolkits}) to support scientific reasoning and analysis.

\subsection{Modular Optimization}
\label{appendix:anagent:rl_training}

To enhance the scientific analysis performance of \ours, we implement modular optimization, training each agent on its specialized task using GRPO \cite{Shao2024DeepSeekMathPT}.


Concretely, by constructing four RL training datasets respectively designed for four agents' specialized tasks, each agent $a \in \{\Planner, \Expert, \Solver, \Critic\}$ is optimized through a specialized reward function tailored to its own task objectives.
As shown in Eq.~\ref{eq:reward_decomposition}, the reward for agent $a$ is decomposed into weighted components with $\sum_m \lambda_{a,m} = 1$ (\S\ref{subsec:anagent:train}).

\subsubsection{\Planner Optimization}
\label{appendix:anagent:rl:planner}

\Planner is optimized through selecting optimal problem-solving strategies from multiple candidate plans for each scientific analysis task
We formulate this as a multi-choice preference selection task, where the agent is asked to identify the most effective decomposition strategy for the given analysis task.

\paragraph{Data.}
We construct the \Planner's RL preference dataset by generating different versions of planning strategies for each scientific analysis instance using two models, where \texttt{Qwen3-VL-8B} acts as the baseline \Planner and \texttt{Gemini-2.5-Flash} as the reference \Planner.
For each input $(x, s, q)$, we collect multiple candidate plans from both models. We then filter the generated problem-solving plans by executing the complete \ours workflow with each reference plan and retaining only those \texttt{Gemini-2.5-Flash} plans that can improve end-to-end performance over the \texttt{Qwen3-VL-8B} baseline:

\begin{align}
\forall \Delta S \in \{\Delta S_{\textsc{Lex}}(y^*, y),\; \Delta S_{\textsc{Sem}}(y^*, y),\; \Delta S_{\textsc{Avg}}(y^*, y)\}\;(\Delta S &> 0)
\label{eq:planner_critic:filter}
\end{align}

\noindent
As shown in Eq.~\ref{eq:planner_critic:filter}, $S_{\textsc{Lex}}(y^*, y)$ (Eq.~\ref{eq:lexical_score}), $S_{\textsc{Sem}}(y^*, y)$ (Eq.~\ref{eq:semantic_score}), and $S_{\textsc{Avg}}(y^*, y)\}$ are the final accuracy scores in our rule-based evaluation (\S\ref{appendix:evaluation:rule_based}).
With \texttt{Qwen3-VL-8B} plans serving as baseline options, the performance-validated \texttt{Gemini-2.5-Flash} plans are designated as ground-truth preferred choices. This approach ensures \Planner to learn to select problem-solving strategies that lead to measurably better scientific analysis quality.

\paragraph{Reward.}
Optimized to make strategic decisions from predefined option sets, \Planner's reward function combines format compliance and answer accuracy:

\begin{equation}
R_{\text{\Planner}} = \lambda_{\text{\textcolor{PlannerColor}{Pf}}} \cdot r_{\text{\textcolor{PlannerColor}{Pf}}}(z_a) + \lambda_{\text{\textcolor{PlannerColor}{Pacc}}} \cdot r_{\text{\textcolor{PlannerColor}{Pacc}}}(z_a, z_a^*)
\label{eq:planner_reward}
\end{equation}
where $r_{\text{\textcolor{PlannerColor}{Pf}}}(z_a)$ validates structural correctness, and $r_{\text{\textcolor{PlannerColor}{Pacc}}}(z_a, z_a^*)$ measures answer accuracy. For multi-choice selections, accuracy is computed as:

\begin{equation}
r_{\text{\textcolor{PlannerColor}{Pacc}}}(z_a, z_a^*) = \begin{cases}
1 & \text{if } \mathcal{O}(z_a) = \mathcal{O}(z_a^*) \\
\frac{2 P_{\mathcal{O}} R_{\mathcal{O}}}{P_{\mathcal{O}} + R_{\mathcal{O}}} & \text{otherwise}
\end{cases}
\label{eq:planner_multichoice_accuracy}
\end{equation}
where $\mathcal{O}(\cdot)$ extracts the set of selected options, $P_{\mathcal{O}} = \frac{|\mathcal{O}(z_a) \cap \mathcal{O}(z_a^*)|}{|\mathcal{O}(z_a)|}$ is precision, and $R_{\mathcal{O}} = \frac{|\mathcal{O}(z_a) \cap \mathcal{O}(z_a^*)|}{|\mathcal{O}(z_a^*)|}$ is recall.

\subsubsection{\Expert Optimization}
\label{appendix:anagent:rl:expert}

As we observe notable inaccurate and hallucinated tool calls in domain-specific knowledge retrieval failures, \Expert is optimized for task-oriented tool calling and execution through GRPO.
Unlike general-purpose tool-use benchmarks, our optimization focuses on domain-specific information retrieval tools tailored to scientific analysis tasks (Tab.~\ref{tab:five_toolkits}), with the specialized RL dataset built on downsampled scientific analysis instances (Tab.~\ref{tab:rl_data_size}).

\paragraph{Data.}
We construct \Expert's RL dataset by pairing each candidate tool with tool-specific queries and formats. Each training instance consists of tool prefix, current knowledge state $\mathcal{K}_{i-1}$, and the ground-truth tool invocation $z_{\text{expert}}^*$.
The dataset emphasizes correct tool selection, proper parameter formatting, and contextually appropriate query aligned with the specific analysis objectives.

\paragraph{Reward.}
\Expert performs tool-based information retrieval. Its reward evaluates both format validity and tool execution correctness:

\begin{equation}
R_{\text{\Expert}} = \lambda_{\text{\textcolor{ExpertColor}{Ef}}} \cdot r_{\text{\textcolor{ExpertColor}{Ef}}}(z_a) + \lambda_{\text{\textcolor{ExpertColor}{Eacc}}} \cdot r_{\text{\textcolor{ExpertColor}{Eacc}}}(z_a, z_a^*)
\label{eq:expert_reward}
\end{equation}

\noindent
where $r_{\text{\textcolor{ExpertColor}{Ef}}}(z_a)$ validates that the action $z_a$ conforms to the expected tool specification query and format.
The accuracy component $r_{\text{\textcolor{ExpertColor}{Eacc}}}$ verifies tool selection and parameter correctness through:

\begin{equation}
r_{\text{\textcolor{ExpertColor}{Eacc}}}(z_a, z_a^*) = \mathbb{I}[\mathcal{T}(z_a) = \mathcal{T}(z_a^*)] \cdot \rho(z_a, z_a^*)
\label{eq:expert_accuracy}
\end{equation}
where $\mathcal{T}(\cdot)$ extracts the tool type, $\mathbb{I}(\cdot)$ is the indicator function, and $\rho(z_a, z_a^*)$ measures parameter correctness using tool-specific validation rules.

\subsubsection{Solver}
\label{appendix:anagent:rl:solver}

\Solver is optimized to generate high-quality scientific analysis. Its RL optimization objective, therefore, focuses on synthesizing retrieved knowledge with input data to produce coherent, accurate, and contextually appropriate analysis writing.

\paragraph{Data.}
\Solver's RL dataset consists of instances $(x, s, q, \mathcal{K}_n, y^*)$, where $y^*$ represents the ground-truth analysis.
We use SciBERT \cite{Beltagy2019SciBERT} as the reward model to evaluate semantic quality, guiding \Solver's scientific analysis generation with improved scientific accuracy, terminology usage, and writing style.
\Solver's reward function incorporates format compliance, length appropriateness (\S\ref{appendix:extended_experiment_analysis:requirement_prompting}), and semantic similarity to guide \Solver toward generating well-structured and high-quality analysis.

\paragraph{Reward.}
As discussed above, \Solver's reward combines format compliance, length appropriateness, and semantic quality (Eq.~\ref{eq:solver_reward}):

\begin{align}
R_{\text{\Solver}} &= \lambda_{\text{\textcolor{SolverColor}{Sf}}} \cdot r_{\text{\textcolor{SolverColor}{Sf}}}(z_a)
+ \lambda_{\text{\textcolor{SolverColor}{Slen}}} \cdot r_{\text{\textcolor{SolverColor}{Slen}}}(z_a, z_a^*) + \lambda_{\text{\textcolor{SolverColor}{Sacc}}} \cdot r_{\text{\textcolor{SolverColor}{Sacc}}}(z_a, z_a^*)
\label{eq:solver_reward}
\end{align}

\noindent
where $r_{\text{\textcolor{SolverColor}{Slen}}}(z_a, z_a^*)$ penalizes outputs with overlong or overshort length relative to the ground truth (Eq.~\ref{eq:length_reward}):

\begin{equation}
r_{\text{\textcolor{SolverColor}{Slen}}}(z_a, z_a^*) = \mathbb{I}[0.5 |z_a^*| \leq |z_a| \leq 1.5 |z_a^*|]
\label{eq:length_reward}
\end{equation}

\noindent
and $r_{\text{\textcolor{SolverColor}{Sacc}}}(z_a, z_a^*)$ computes semantic similarity using SciBERT token-level embeddings (Eq.~\ref{eq:semantic_reward}):

\begin{equation}
r_{\text{\textcolor{SolverColor}{Sacc}}}(z_a, z_a^*) = \frac{2 P_{\text{emb}} R_{\text{emb}}}{P_{\text{emb}} + R_{\text{emb}}}
\label{eq:semantic_reward}
\end{equation}
where $P_{\text{emb}}$ and $R_{\text{emb}}$ are computed from the maximum token-level cosine similarities between SciBERT embeddings of $z_a$ and $z_a^*$.

\subsubsection{\Critic Optimization}
\label{appendix:anagent:rl:critic}

Similar to \Planner, \Critic is optimized through multi-choice solution preference selection, but focuses on assessing analysis quality and providing constructive feedback for analysis refinement.

\paragraph{Data.}
We follow the same data construction methodology as \Planner (\S\ref{appendix:anagent:rl:planner}).
Concretely, for each scientific analysis $y_i$ generated by \Solver, we collect critique feedback from both Qwen2-VL-8B (serving as baseline \Critic) and Gemini-2.5-Flash (serving as reference \Critic).
Following \Planner data filtering, we filter the feedback by evaluating whether applying the suggested revisions leads to improved analysis quality (Eq.~\ref{eq:planner_critic:filter}).
As such, only those Gemini-2.5-Flash critiques that can enhance \Solver's analysis writing are retained as ground-truth preferred feedback options, while Qwen2-VL-8B critiques serve as baseline options.
This helps \Critic to learn to identify key quality deficiencies and provide actionable improvement suggestions across multiple evaluation dimensions (Fig.~\ref{fig:critic_prompt}).

\paragraph{Reward.}
Similar to \Planner, \Critic's reward also consists of format compliance and
answer accuracy (Eq.~\ref{eq:critic_reward}):

\begin{equation}
R_{\text{\Critic}} = \lambda_{\text{\textcolor{CriticColor}{Cf}}} \cdot r_{\text{\textcolor{CriticColor}{Cf}}}(z_a) + \lambda_{\text{\textcolor{CriticColor}{Cacc}}} \cdot r_{\text{\textcolor{CriticColor}{Cacc}}}(z_a, z_a^*)
\label{eq:critic_reward}
\end{equation}

\noindent
where $r_{\text{\textcolor{CriticColor}{Cf}}}(z_a)$ validates evaluation formatting correctness, and $r_{\text{\textcolor{CriticColor}{Cacc}}}(z_a, z_a^*)$ measures answer accuracy.
Same to \Planner, answer accuracy is computed as (Eq.~\ref{eq:critic_multichoice_accuracy}):

\begin{equation}
r_{\text{\textcolor{CriticColor}{Cacc}}}(z_a, z_a^*) = \begin{cases}
1 & \text{if } \mathcal{O}(z_a) = \mathcal{O}(z_a^*) \\
\frac{2 P_{\mathcal{O}} R_{\mathcal{O}}}{P_{\mathcal{O}} + R_{\mathcal{O}}} & \text{otherwise}
\end{cases}
\label{eq:critic_multichoice_accuracy}
\end{equation}

\section{Scientific Analysis Evaluation}
\label{appendix:evaluation}

\subsection{Rule-Based Evaluation}
\label{appendix:evaluation:rule_based}

We design rule-based evaluation in both lexical and semantic dimensions.
Lexical evaluation consists of \textsc{Rouge-L} (Eq.~\ref{eq:rouge_l}) \cite{lin2004rouge}, \textsc{Bleu} (Eq.~\ref{eq:bleu}) \cite{papineni2002bleu}, and word overlap (Eq.~\ref{eq:word_overlap}) metrics. For semantic evaluation, we employ cosine similarity (Eq.~\ref{eq:cosine_sim}), SciBERT (Eq.~\ref{eq:scibert_score}) \cite{Beltagy2019SciBERT}, and \textsc{Meteor} (Eq.~\ref{eq:meteor}) \cite{banerjee2005meteor} metrics to calculate the semantic assessment scores.
In addition to the lexical score $S_\textsc{Lex}$ (Eq.~\ref{eq:lexical_score}) and semantic score $S_\textsc{Sem}$ (Eq.~\ref{eq:semantic_score}) that are calculated as the mean of their three metrics, respectively. The overall score $S_\textsc{Avg}$ (Eq.~\ref{eq:overall_score}) is averaged across all six metrics.

Concretely, given the model-generated analysis $y$ and ground-truth analysis $y^*$, the lexical and semantic evaluation scores are calculated as follows:

\paragraph{\textsc{Rouge-L}} measures the longest common subsequence (\textsc{LCS}) between $y$ and $y^*$:
\begin{equation}
\textsc{Rouge-L}(y^*, y) = \frac{\textsc{LCS}(y^*, y)}{\max(|y^*|, |y|)}
\label{eq:rouge_l}
\end{equation}
where $\textsc{LCS}(y^*, y)$ computes the length of the longest common subsequence, and $|\cdot|$ denotes sequence length.

\paragraph{\textsc{Bleu}} calculates n-gram precision with a brevity penalty (\textsc{BP}):
\begin{equation}
\textsc{Bleu}(y^*, y) = \textsc{BP} \cdot \exp\left(\sum_{n=1}^{N} w_n \log p_n\right)
\label{eq:bleu}
\end{equation}
where $p_n$ is the n-gram precision, $w_n$ is the weight for each n-gram (typically $w_n = 1/N$), and $\textsc{BP}$ is the brevity penalty to penalize short predictions.

\paragraph{Word Overlap} measures the Jaccard similarity between word sets:
\begin{equation}
\textsc{Word}(y^*, y) = \frac{|W(y^*) \cap W(y)|}{|W(y^*) \cup W(y)|}
\label{eq:word_overlap}
\end{equation}
where $W(\cdot)$ extracts the set of unique words from a text.

\paragraph{Cosine Similarity} measures the angle between sentence embeddings:
\begin{equation}
\textsc{Cosine}(y^*, y) = \frac{\mathbf{e}(y^*) \cdot \mathbf{e}(y)}{\|\mathbf{e}(y^*)\| \|\mathbf{e}(y)\|}
\label{eq:cosine_sim}
\end{equation}
where $\mathbf{e}(\cdot)$ is a sentence embedding function that maps text to a dense vector representation.

\paragraph{SciBERT Score} computes token-level semantic similarity using SciBERT embeddings:
\begin{equation}
\text{SciBERT}(y^*, y) = \frac{2 \cdot P_{\text{SciBERT}} \cdot R_{\text{SciBERT}}}{P_{\text{SciBERT}} + R_{\text{SciBERT}}}
\label{eq:scibert_score}
\end{equation}
where $P_{\text{SciBERT}}$ and $R_{\text{SciBERT}}$ are precision and recall computed from token-level cosine similarities between SciBERT embeddings of $y^*$ and $y$.

\paragraph{\textsc{Meteor}} evaluates based on unigram matching with synonym and paraphrase support:
\begin{equation}
\textsc{Meteor}(y^*, y) = F_{\text{mean}} \cdot (1 - \textsc{Pen})
\label{eq:meteor}
\end{equation}
where $F_{\text{mean}}$ is the harmonic mean of unigram precision and recall, and $\textsc{Pen}$ is a fragmentation penalty based on chunk count.

\paragraph{Final Scores}
include the lexical score $S_{\textsc{Lex}}$ (Eq.~\ref{eq:lexical_score}) and semantic score $S_{\textsc{Sem}}$ (Eq.~\ref{eq:semantic_score}) that are computed as:

\begin{equation}
S_{\textsc{Lex}}(y^*, y) = \frac{\textsc{Rouge-L} + \textsc{Bleu} + \textsc{Word}}{3}
\label{eq:lexical_score}
\end{equation}

\begin{equation}
S_{\textsc{Sem}}(y^*, y) = \frac{\textsc{Cosine} + \text{SciBERT} + \textsc{Meteor}}{3}
\label{eq:semantic_score}
\end{equation}

The overall evaluation score $S_{\textsc{Avg}}$ (Eq.~\ref{eq:overall_score}) is the average across all six metrics:

\begin{equation}
S_{\textsc{Avg}}(y^*, y) = \frac{1}{6}\sum_{i=1}^{6} m_i(y^*, y)
\label{eq:overall_score}
\end{equation}
where $m_i$ denotes each of the six individual metrics.

\subsection{Five-Dimensional Evaluation Protocol}
\label{appendix:evaluation:five_dimension}

We apply our five-dimensional evaluation protocol to both MLLM-As-Judge and human expert assessment (\S\ref{subsec:anabench:evaluation_metrics} \& \ref{appendix:preliminary:method}).
Fig.~\ref{fig:five_dimension_evaluation_prompt} shows the evaluation prompt for assessing scientific analysis quality.
Accordingly, we prompt and finetune \Critic for self- reflection and correction with the same five criteria (Fig.~\ref{fig:critic_prompt}).

\subsection{Performance Difference}
\label{appendix:evaluation:performance_delta}

To quantify the performance differences between two methods, we calculate \textit{absolute} performance difference ($\Delta_{\textit{abs}}\%$) and \textit{relative} performance difference ($\Delta_{\textit{rel}}\%$) across metrics.

\paragraph{Absolute Performance Difference.}
The absolute performance difference $\Delta_{\textit{abs}}$ (Eq.~\ref{eq:absolute_delta}) measures the direct performance gap between our method and the baseline:

\begin{equation}
\Delta_{\textit{abs}} = (S_{\textit{ours}} - S_{\textit{baseline}}) \%
\label{eq:absolute_delta}
\end{equation}

where $S$ denotes a metric score (\textit{e.g.}, SciBERT, \textsc{BLEU}, etc.), and the $\Delta_{\textit{abs}}$ result is expressed in percentage points.

\paragraph{Relative Performance Difference.}
The relative performance difference $\Delta_{\textit{rel}}$ (Eq.~\ref{eq:relative_delta}) measures the proportional improvement or degradation with respect to the baseline:

\begin{equation}
\Delta_{\textit{rel}} = \frac{S_{\textit{ours}} - S_{\textit{baseline}}}{S_{\textit{baseline}}} \times 100\%
\label{eq:relative_delta}
\end{equation}

where $S$ denotes a metric score, and the result represents the percentage change relative to the baseline performance.

\section{Implementation Details}
\label{appendix:configuration}

To complement the implementation details presented in \S\ref{sec:experiments}, we further summarize our experiment configurations and computation overhead in Tabs.~\ref{tab:exp:configuration}-\ref{tab:exp:computation_overhead}.
As can be seen in Tab.~\ref{tab:exp:computation_overhead}, RL demands significantly more computation resources for smaller MLLMs with model sizes of 3B–4B parameters. Therefore, we finetuned only Qwen2.5-VL-3B and Qwen3-VL-4B models for computation efficiency. Their performance (Tab.~\ref{tab:experiment:finetuned_main_results}) further substantiates the effectiveness of combining supervised finetuning and reinforcement learning, delivering remarkable cumulative optimization benefits even for small-size MLLMs.

\input{tables/exp_config}
\input{tables/exp_main3_mllm_as_judge}
\input{figures/length}

\input{tables/method_toolkit}

\section{A Deeper Dive Beneath the Results}
\label{appendix:extended_experiment_analysis}

\subsection{MLLM-As-Judge Evaluation On Scientific Analysis}
\label{appendix:extended_experiment_analysis:mllm_as_judge}

Leveraging \texttt{GPT-4.1-mini} \cite{openai_gpt4_1} as the MLLM judger, we conduct MLLM-as-judge evaluation following the five-dimensional evaluation protocol (\S\ref{appendix:evaluation:five_dimension}).
Due to cost constraints (approximately \~\$45/100 instances), for each agent, we evaluate performance on a randomly sampled subset of 300 instances drawn from the complete evaluation set (\S\ref{appendix:anabench:train_eval_data}).
Results are summarized in Tab.~\ref{tab:experiment:mllm_as_judge_main_results}. Across five evaluated dimensions and the averaged $S_{\textsc{Mllm}}$, results demonstrate a monotonic improvement from baselines to training-free \ours, with the finetuned version achieving the highest performance. Comparing between \ours finetuned through SFT and the combination of SFT+RL, the latter \ours consistently outperforms SFT-only counterparts, revealing the significance of specialized ability development in multi-agent systems.
Moreover, while the absolute evaluation scores differ, our MLLM-as-judge assessment yields performance trends that are consistent with those observed under rule-based evaluation (Tabs.~\ref{tab:experiment:training_free_main_results}-\ref{tab:experiment:finetuned_main_results}). \ours that perform well on rule-based metrics also achieve higher rankings under MLLM-based judgment, supporting the use of rule-based evaluation and MLLM-as-judge for scientific analysis assessment.

\subsection{Specialized RL as a Double-Edged Sword for Enhancing Scientific Table \& Figure Analysis}
\label{appendix:subsec:rl_is_a_double_edged_sword}

Trained on tailored datasets (\S\ref{subsec:anagent:train}), agents benefit substantially from specialized RL optimization, achieving notable performance improvements (Tab.~\ref{tab:experiment:finetuned_main_results}).
In particular, RL finetuning yields consistent gains ($\Delta_{\textsc{Avg}} \geq$ $10.12\%$), while combining SFT with RL further amplifies performance, achieving even larger gains ($\Delta_{\textsc{Avg}} \geq$ $33.10$).
Nevertheless, these gains come at a high computational cost.
As shown in Tab.~\ref{tab:exp:computation_overhead}, specialized RL introduces substantially higher computation overhead as compared to SFT alone.
When jointly considering performance optimization and computation efficiency, SFT demonstrates a more favorable trade-off for scientific table \& figure analysis tasks that demand long-context understanding and long-output generation (\S\ref{subsec:results:main_results}).

Overall, while RL is highly effective at optimizing specialized agents for specialized tasks, it functions as a double-edged sword: delivering stronger task-specific optimization at the expense of markedly increased computational cost, which may limit its practicality when finetuning larger models in resource-intensive scientific settings.


\subsection{How Training Data Recipe Shapes Finetuning Outcomes?}
\label{appendix:subsec:extended_analysis_on_training_data}

\textbf{Benefits of Increased Data Size.}
Fig.~\ref{fig:exp:ablation_data_size_domain_format_type} shows the performance of \ours finetuned on different subsets of the training set (\S\ref{appendix:anabench:train_eval_data}), revealing the benefits of finetuning with increased data sizes. 
While Fig.~\ref{fig:exp:ablation_data_size_domain_format_type}(a) isolates the effects of data size by comparing finetuning on a 30K-instance subset against the full dataset, Fig.~\ref{fig:exp:ablation_data_size_domain_format_type}(b)-(d) evaluate subsets constructed along different data types, domains, and formats, respectively.
As shown in Fig.~\ref{fig:exp:ablation_data_size_domain_format_type}(a), finetuning on 30K instances leads to consistent performance degradation across all four MLLM agents as compared to training on the full set ($\Delta_{\textit{rel}} \geq$ $\downarrow10.27\%$), demonstrating the clear benefits of increased data size. Results in Fig.~\ref{fig:exp:ablation_data_size_domain_format_type}(b)-(d) lend further evidence to this advantage on subsets with varying data sizes and compositions ($\Delta_{\textit{rel}} \geq$ $\downarrow9.29\%$).

\paragraph{Domain-Specific Learning Leads To Decreased Generalizability.}
Fig.~\ref{fig:exp:ablation_data_size_domain_format_type}(b) examines the impact of domain-specific finetuning on agent performance by comparing models trained exclusively on computer science (CS), biology, and a mixture of all nine domains. Across all four MLLM agents, training on the full set of nine domains consistently yields the strongest performance, while domain-specific training leads to notable degradation ($\Delta_{\textit{rel}} \geq$ $\downarrow26.55\%$). In particular, agents finetuned solely on biology exhibit the lowest performance ($S_{\textsc{Avg}} \leq$ $29.79\%$), followed by those trained only on CS ($S_{\textsc{Avg}} \leq$ $31.92\%$). This trend reveals that restricting training data to a single domain substantially limits agents’ ability to generalize beyond domain-specific patterns. Moreover, the larger performance gap observed for biology-only (\textit{25 subdomains}) training compared to CS (\textit{40 subdomains}) indicates that narrower or more specialized domains impose stronger constraints on cross-domain transfer, further emphasizing the significance of domain diversity for robust general-purpose scientific reasoning.

\paragraph{Single-Format Learning Limits Cross-Format Adaptability.}
Fig.~\ref{fig:exp:ablation_data_size_domain_format_type}(c) assesses the effects of data format on agent performance by comparing models finetuned exclusively on LaTeX, exclusively on XML, and on both.
Across all four MLLM agents, training on LaTeX ($S_{\textsc{Avg}} \geq$ $31.98\%$) consistently outperforms training on XML ($S_{\textsc{Avg}} \geq$ $26.13\%$), indicating that LaTeX-based supervision provides more effective signals for scientific reasoning. However, agents trained on either single format exhibit degraded performance compared to those trained on both ($\Delta_{\textit{rel}} \geq$ $\downarrow8.85\%$). These results suggest that restricting training to a single data format limits agents’ adaptability to heterogeneous input representations, while exposure to multiple formats enhances robustness and cross-format generalization.

\paragraph{Single-Type Learning Limits Multi-Type Adaptability.}
Fig.~\ref{fig:exp:ablation_data_size_domain_format_type}(d) further analyzes the impact of data type by comparing agents finetuned exclusively on tables, exclusively on figures, and on both.
Across all four MLLM agents, models trained only on tables ($S_{\textsc{Avg}} \geq$ $28.29\%$) consistently underperform those trained only on figures \& tables ($S_{\textsc{Avg}} \geq$ $30.57\%$), revealing limited generalization from structured tabular inputs to multimodal reasoning. More importantly, agents trained on either single type exhibit substantial performance degradation relative to those trained on both tables and figures ($\Delta_{\textit{rel}} \geq$ $\downarrow9.53\%$), indicating a pronounced loss of adaptability when exposure is restricted to a single data type. These trends highlight that learning from a single data type constrains agents' ability to handle heterogeneous layouts and modalities, while maintaining data heterogeneity in training is critical for enhanced scientific analysis performance.

\subsection{Significance of Contextual Information \& Domain Knowledge In Scientific Analysis}
\label{appendix:extended_experiment_analysis:context_significance}

Contextual information and domain-specific knowledge play a critical role in scientific analysis, supporting both accurate observations and well-grounded inference.
To examine the role of contextual information and domain knowledge in scientific analysis, we conduct a comparative evaluation across baselines and \ours.
Specifically, we consider baselines and training-free \ours under two settings:
(1) with gold contextual information and domain-specific knowledge included in the input, and (2) with them excluded from the input.
All models are evaluated on the same evaluation split of \ourbench (\S\ref{appendix:anabench:train_eval_data}), ensuring evaluation consistency across experimental conditions.

The provided contexts in setting (1) are gold contexts extracted from highly-relevant components of the source scientific papers of each instance, such as \textit{table and figure captions}, \textit{sections}, \textit{citations}, \textit{etc}., that both refer to and are referred by the task input ($d=1$). These contexts represent the most relevant and effectively optimal information for the task, as simulating what human researchers naturally attend to during scientific analysis. They provide a highly informative signal that agents would otherwise search for and extract themselves, often at greater computational cost and with lower accuracy.

Fig.~\ref{fig:exp:ablation_context} visualizes the quantitative results of this comparison and highlights consistent performance improvements when additional contextual information is provided.
Across both baseline methods and \ours, enriching the input with supplementary contexts and domain-specific knowledge yields marked gains, with relative improvements $\Delta_{rel} \geq 4.42\%$ on scientific analysis tasks. These improvements are observed consistently across evaluation settings, indicating that the benefit is not model-specific but instead reflects a general reliance on contextual grounding for complex scientific reasoning. In particular, domain-specific context enables more accurate interpretation of technical concepts, experimental setups, and implicit assumptions, which are often under-specified in isolated inputs. The results empirically validate the significance of contextual and domain knowledge in complex scientific analysis.

\input{figures/exp_ablation_context}

\subsection{Can AI Agents Deliver What Is Asked?}
\label{appendix:extended_experiment_analysis:requirement_prompting}

In \textit{human–AI co-discovery}, AI agents are expected to follow human-specified requirements, especially in terms of the \textit{depth}, \textit{breadth}, and \textit{length} of writing in our scientific analysis tasks.
This raises a central question: \textbf{\textit{Can AI agents reliably meet stated expectations?}}

Motivated by this question, we conduct an ablation study examining the effects of requirement prompting using three types of explicit statements: \textit{analysis length}, \textit{analysis width}, and \textit{analysis depth}.
Particularly, each prompt specifies the expected \textit{writing length}, \textit{contextual breadth}, or \textit{analytical depth} of the scientific analysis writing.
Relative to the baseline setting in which no requirements are specified, we evaluate agent performance under five conditions: (1) no requirement, (2) length requirement, (3) width requirement, (4) depth requirement, and (5) all three requirements combined.
To maintain valid comparisons, we leverage samples with \textit{medium} analysis width and \textit{in-depth} analysis depth as the test set (Fig.~\ref{fig:anabench_statistics} \& \S\ref{appendix:benchmark:statistics}), while remaining the heterogeneity of the dataset with varying data and analysis complexties (\S\ref{appendix:benchmark:curriculum:data_complexity} \& \S\ref{appendix:benchmark:curriculum:analysis_complexity}).

Comparing between baselines and \ours (Fig.~\ref{tab:anagent:variants}), results in Fig.~\ref{fig:abation:requirement_prompting} demonstrate consistent effects of explicit analysis requirements on MLLM agents’ performance.
As indicated by the red bars, imposing writing constraints on analysis length leads to a noticeable performance degradation, suggesting current AI agents' lack of sufficient awareness of how to effectively organize and prioritize their reasoning under stated space limitations.
In contrast, explicitly specifying requirements on analysis width or depth significantly improves performance, as such guidance helps agents better understand the expected analytical scope and level of detail for each task.
Notably, combining analysis length, width, and depth together yields the largest performance gains. This indicates a strong complementary effect among these dimensions, highlighting the importance of holistic analysis constraints in enhancing AI agents’ reasoning capabilities and informing the design of future AI research assistants and human-AI co-discovery systems.

\input{figures/exp_requirement_prompting}

To further investigate AI agents' awareness of generation length and the implications of explicitly stated length expectations, we extend our study to compare the lengths of agent-generated analyses with those of ground-truth analyses.
Fig.~\ref{fig:length_distribution_gpt} presents the distribution of analysis lengths for three settings using agents powered by GPT-4.1-mini:
(1) Baseline agent \textit{without} explicit analysis length requirement,
(2) Baseline agent \textit{with} explicit analysis length requirement,
and (3) \ours \textit{with} explicit analysis length requirement.
As shown in Fig.~\ref{fig:length_distribution_gpt}, the baseline agent without analysis length expectations consistently generates short analyses, in stark contrast to the substantial variability observed in ground-truth analyses colored in gray.
On the other hand, explicitly instructing the agent to generate analyses around the specified length not only increases the overall length of the generated outputs but also substantially enriches the variance in analysis length. \ours with length expectations exhibit variation more consistent with the ground-truth analyses.
Building upon the observations from GPT-4.1-mini (Fig.~\ref{fig:length_distribution_gpt}), we employ Qwen3-VL-8B to power agents under four settings:
(1) Baseline agent \textit{without} explicit analysis length requirement;
(2) Baseline agent \textit{with} explicit analysis length requirement;
(3) \ours \textit{with} explicit analysis length requirement;
and (4) \ours, powered by finetuned Qwen3-VL-8B, \textit{with} explicit analysis length requirement.
The upper three groups in Fig.~\ref{fig:length_distribution_qwen} exhibit patterns similar to those observed with GPT-4.1-mini: The baseline agent without explicit length constraints consistently produces short analyses, whereas introducing length requirements increases both the average length and the variance of the generated analyses. On the contrary, \ours finetuned through the combination of SFT+RL shows markedly higher variance in analysis length, accompanied by notably improved performance (Tab.~\ref{tab:experiment:finetuned_main_results}).
Collectively, Figs.~\ref{fig:length_distribution_gpt}-\ref{fig:length_distribution_qwen} unveil that current AI agents fail to reliably interpret explicitly stated analysis lengths, which function not only as generation constraints but also as implicit signals of the expected analytical coverage scope.



\subsection{Scientific Analysis Across Data Complexities}
\label{appendix:extended_experiment_analysis:data_complexity}

The complexity of each scientific analysis task is fundamentally shaped by the nature of the input data itself.
Across real-world scientific writing, analyses are conducted over heterogeneous data that vary in \textit{type}, \textit{source}, \textit{format}, and \textit{domain} (\S\ref{appendix:benchmark:curriculum:data_complexity} \& Fig.~\ref{fig:anabench_statistics}), each of which introduces distinct interpretive and integration challenges.
For example, tables and figures differ substantially in how information is structured and accessed, while variations in data domains shape what contextual information and domain-specific knowledge are needed for accurate reasoning and analysis.
These variations give rise to different levels of data complexity that directly affect MLLM agents' abilities to search, perceive, extract, and comprehend the related evidence.
To systematically examine these challenges across varying data complexities, we evaluate scientific analysis with different data \textit{types}, \textit{sources}, \textit{formats}, and \textit{domains}.
Beyond contrasting individual data types, we further extend our investigation to how models handle varying \textit{input scales}, where evidence is distributed across multiple tables, multiple figures, or multimodal combinations of different data types.
Results reveal how multiple facets of data complexity influence scientific table \& figure analysis, suggesting potential directions for future improvement.

\paragraph{Tabular Structures Are More Challenging Than Multimodal Figure Reasoning.}
Although \textit{figures} require multimodal reasoning, \textit{tables} pose distinct and often greater challenges due to the need to accurately parse and reason over complex tabular structures. As shown in Fig.~\ref{fig:exp:ablation_complexity_data:type}, both the baselines and \ours consistently achieve higher performance on scientific figures than on tables, with absolute improvements of $\Delta_{abs} \geq 2.40\%$ for baselines, $\Delta_{abs} \geq 1.47\%$ for \ours, $\Delta_{abs} \geq 4.83\%$ for finetuned \ours. Augmented with task-oriented \Expert, \ours narrows the performance gap to $1.47\%$. Nonetheless, the heterogeneous layouts and diverse semantic intents of scientific tables (\textit{e.g.}, reporting empirical results versus comparing methods) continue to present substantial challenges for reliable scientific analysis.

\input{figures/exp_ablation_complexity_data_type}

\paragraph{LaTeX Is Easier to Understand Than XML.}
We observe a clear and consistent performance advantage when scientific contents are represented in \texttt{LaTeX} rather than \texttt{XML}. As shown in Fig.~\ref{fig:exp:ablation_complexity_data:format}, all baselines as well as \ours achieve higher accuracy on LaTeX data ($S_{\textsc{Avg}} \geq 27.15$) as compared to XML ($S_{\textsc{Avg}} \geq 23.44$), indicating that \texttt{XML} introduces additional challenges for MLLM-powered agents.
Across both data formats, performance improves monotonically from baselines to \ours and further to the finetuned \ours, demonstrating the robustness of our approach regardless of representation. Nevertheless, the overall performance gap between \texttt{LaTeX} and \texttt{XML} persists ($\Delta_{abs} \geq 1.07$), suggesting that \texttt{XML}’s verbose and nested structure hinders effective reasoning.
A possible reason is that \texttt{LaTeX} is the dominant format for scientific writing and is therefore more prevalent in model pretraining corpora, leading to stronger prior familiarity. In contrast, XML often emphasizes structural markup over semantic clarity, requiring agents to interpret scientific meaning from less explicit cues and references, which further exacerbates the difficulty.

\input{figures/exp_ablation_complexity_data_format}

\paragraph{Review and Survey Papers Are More Challenging Than General Scientific Literature.}
We observe that both baselines and \ours consistently achieve higher performance on general scientific paper analysis tasks than on review and survey papers. As illustrated in Fig.~\ref{fig:exp:ablation_complexity_data:source}, the absolute performance gap between these two data sources remains substantial across all models ($\Delta_{abs} \geq 3.51\%$), reaching up to $8.07\%$ on the InternVL3.5-8B baseline. Despite this difficulty, performance improves monotonically from baselines to \ours and further to the finetuned \ours on both paper sources, indicating the effectiveness of our method across varying document sources and types. Notably, in most cases, \ours and finetuned \ours can narrow the gap between general papers and review/survey papers, indicating the effectiveness of specialized tools for information searching and retrieval. Through manual checking, we also observe that scientific analysis sections in review and survey papers are significantly less likely to be self-contained as compared with general papers. This supports our potential thinking that attributes the increased difficulty of review and survey papers to their heavy reliance on extensive internal and external references, cross-paper comparisons, and high-level synthesis, which makes their scientific analysis less self-contained than that of general research articles.

\input{figures/exp_ablation_complexity_data_source}

\paragraph{Enhancing Scientific Analysis Across Domains.}
We observe inconsistent variability in scientific analysis performance across domains. As shown in Fig.~\ref{fig:exp:ablation_complexity_data:domain}, baselines exhibit pronounced domain variability in scientific analysis performance.
In particular, the lowest baseline performance is observed on \textit{Statistics} domain ($S_{\textsc{Avg}}=17.08\%$), followed by \textit{Economics} ($S_{\textsc{Avg}}=20.28\%$) and \textit{Quantitative Finance} ($S_{\textsc{Avg}}=21.87\%$). In contrast, the highest baseline performance is achieved on \textit{Quantitative Biology} ($S_{\textsc{Avg}}=33.96\%$). This disparity reveals that baseline agents struggle to generalize scientific reasoning capabilities across domains with varying levels of mathematical abstraction, formalism, and domain-specific assumptions.
In contrast, both \ours and finetuned \ours consistently improve performance across all domains ($\Delta_{rel} \geq 4.06\%$), effectively elevating even the most challenging \textit{Statistics} domain from the lowest baseline performance to $S_{\textsc{Avg}}=36.84\%$ under finetuned \ours. Moreover, when comparing the performance distribution across domains, baseline agents show pronounced inter-domain performance gaps, while \ours significantly reduces such variation, and finetuned \ours further presents an even more uniform performance distribution across domains. This trend reveals the effectiveness of \ours in mitigating domain-specific reasoning bottlenecks, leading to more robust and consistent scientific analysis across diverse research areas.

\input{figures/exp_ablation_complexity_data_domain}

\paragraph{Scaling Inputs Furnishes Additional Contexts While Simplifying Data Complexity.}
In addition to the width, depth, and objectives of an analysis task, the nature of the input data plays a fundamental role in determining the complexity of scientific analysis. Input data may take the form of a single table or figure, multiple tables or figures, or a combination of both. To this end, we conduct an additional study examining input coverage across inputs that vary in layout, modality, and quantity.
As shown in Fig.~\ref{fig:exp:ablation_complexity_data:coverage}, expanding input coverage generally leads to more reliable and higher-quality analyses, with our method consistently outperforming baseline counterparts across all settings. This trend suggests that broader data coverage provides complementary contextual cues and domain knowledge that help agents better interpret experimental evidence and articulate scientific insights. Notably, agents tend to benefit from multi-input settings over single-input ones (i.e., \textit{darker} bars higher than \textit{lighter} bars in most cases), indicating that aggregating information across multiple sources often facilitates reasoning by grounding analysis in richer contextual support.
The benefits of increased coverage are particularly pronounced for multimodal inputs. MLLM-powered agents exhibit stronger reasoning abilities when figures are present, either alone or in combination with tables, as compared to tabular-only layouts. In some cases (\textit{e.g.}, Qwen2.5-VL baselines), single-figure inputs can yield performance comparable to, or even exceeding, that of other input coverage settings, underscoring the advantages of visual representations for improved scientific reasoning. While scaling inputs typically enhances performance, these gains are not unbounded. Agents with limited capacity may struggle to effectively integrate excessive information, resulting in diminished returns (\textit{e.g.}, Qwen3-VL-4B gains higher performance on \texttt{S-T\&S-F} over \texttt{M-T\&M-F}).
Overall, evaluation results across six input coverage settings highlight a nuanced trade-off in input scaling, where richer coverage can reduce effective data complexity and improve analysis, as long as it aligns with the agent’s reasoning capacity.

\input{figures/exp_ablation_complexity_data_coverage}

\subsection{Scientific Analysis Across Analysis Complexities}
\label{appendix:extended_experiment_analysis:analysis_complexity}

\input{figures/exp_ablation_complexity_analysis_width}

Variations in analysis \textit{depths}, \textit{widths}, and \textit{objectives} contribute to tiered curriculum complexities (\S\ref{appendix:benchmark:curriculum:analysis_complexity} \& Fig.~\ref{fig:anabench_statistics}) that reflect the level of detail, breadth of coverage, and overall focus in real-world scientific analysis. 
With varying writing widths, scientific analysis may either \textit{focus narrowly on the individual table or figure}, such as summarizing and interpreting it in isolation to draw specific findings; or \textit{adopt a broader perspective}, comparing across multiple tables and figures to identify patterns and synthesize insights.
The combination of analysis width and depth further accounts for how much of an analysis section is devoted to the targeted input. Some analyses may dedicate only a small portion to a specific table or figure, while others may use it as the central basis for extensive discussion and conclusions.
Moreover, targeting methodology interpretation or experimental demonstration, analysis objectives fundamentally shape how an analysis is formulated.
By evaluating the challenges introduced by varying analysis widths, depths, and objectives, results reveal their impact on scientific table \& figure analysis across different agents.

\paragraph{The Core Challenge Of Analysis Width Comes From External References.}
As suggested by the challenges (Fig.~\ref{fig:preliminary}) and the curriculum of \textit{(d) Analysis Width} (Fig.~\ref{fig:anabench_statistics}), we implement the four-level complexity curriculum for analysis width, including: \textbf{\textit{easy}} tasks for self-contained analysis writing, \textbf{\textit{moderate}} tasks with internal references within the source paper, \textbf{\textit{hard}} tasks with external references aside from the source paper, and \textbf{\textit{challenging}} tasks that contain both internal and external references.
Results in Fig.~\ref{fig:exp:ablation_complexity_analysis:width} reveal how increasing analysis width affects scientific table and figure analysis writing.
Specifically, across all complexity levels, \ours variants consistently outperform baselines, with further improvements observed after finetuning, unveiling the effectiveness of our multi-agent collaboration design. The reduced variations among four analysis width curricula further demonstrate the robustness of collaborative analysis writing under varying context widths. Notably, task difficulty does not monotonically increase with the curriculum level. For baselines, the highest performance is consistently achieved on self-contained tasks and degrades substantially when external references are involved, reflecting agents' limited capabilities to adaptively comprehend and incorporate additional contextual knowledge. In contrast, for both \ours and its finetuned variant, \textbf{\textit{moderate}} tasks achieve the strongest performance, suggesting that internal references, such as paper-specific definitions, methodologies, experiments, domain concepts, etc., provide the most effective contextual grounding for scientific analysis writing.
Performance declines markedly on \textbf{\textit{hard}} and \textbf{\textit{challenging}} tasks that involve external references, underscoring external knowledge integration as the primary bottleneck in scientific analysis. Interestingly, in some cases, agents exhibit lower performance on \textbf{\textit{hard}} tasks than on \textbf{\textit{challenging}} ones, despite the latter containing the broader combination of both internal and external references. This observation reveals that current MLLM-powered agents struggle to selectively identify, interpret, and adapt externally referenced knowledge that is closely connected to the internal context, whereas the presence of internal references in challenging tasks may partially anchor the use of external information.
Overall, these results highlight the critical role of highly relevant internal context in supporting scientific analysis, as it supplies essential concepts, core methodologies, and domain-specific cues that facilitate coherent reasoning. Conversely, self-contained tasks place greater emphasis on precise interpretation of the input tables and figures themselves, while external-reference-heavy tasks pose the greatest challenge due to the need for adaptive comprehension and task-specific integration of externally sourced knowledge.

\input{figures/exp_ablation_complexity_analysis_depth}

\paragraph{Increased Analysis Depth Leads to More Challenging Tasks.}
Following the curriculum design of \textit{(c) Analysis Depth} in Fig.~\ref{fig:anabench_statistics}, we instantiate two levels of analysis depth as illustrated in Fig.~\ref{fig:preliminary}: (1) \textbf{\textit{shallow}} analysis tasks, which primarily involve surface-level description and direct interpretation of tables and figures, and (2) \textbf{\textit{in-depth}} analysis tasks that require reasoning beyond evidence, causal and logical interpretation, as well as synthesis of underlying conclusions or experimental implications.
Ablation results are presented in Fig.~\ref{fig:exp:ablation_complexity_analysis:depth}.
Across all evaluated methods, performance on \textbf{\textit{in-depth}} analysis tasks is consistently lower than that on \textbf{\textit{shallow}} tasks, confirming that increased analysis depth introduces substantially greater difficulty. This performance gap suggests that deep scientific reasoning, such as drawing non-trivial inferences, explaining observed trends, or connecting empirical results to broader methodological or theoretical considerations, remains challenging for current MLLM agents.
Despite this increased difficulty, our approach demonstrates consistent improvements over baselines at both depth levels, with additional gains obtained through finetuning. Importantly, the performance gap between shallow and in-depth analysis is noticeably reduced for \ours ($\Delta_{\textit{abs}} \leq 4.25\%$) and its finetuned variant ($\Delta_{\textit{abs}} \leq 0.86\%$) as compared to baselines ($\Delta_{\textit{abs}} \leq 5.41\%$). This reduction indicates that collaborative analysis writing is effective at supporting not only shallow but also deeper reasoning, enabling agents to better decompose complex analytical requirements and progressively refine interpretations.
Overall, these results suggest that, while increased analysis depth substantially raises task difficulty, structured multi-agent collaboration provides a meaningful means of handling deeper scientific reasoning and analysis.

\input{figures/exp_ablation_complexity_analysis_objective}


\paragraph{Methodology-Oriented Analysis Is More Challenging Than Experiment-Oriented Analysis.}
Scientific analysis tasks are typically driven by different analytical objectives. Accordingly, we categorize scientific analysis into two categories according to the analysis objective of the task: (1) \textbf{\textit{methodology-oriented}} analysis, which aims to explain, interpret, and reason about methodological designs, theoretical formulations, algorithmic principles, and so on; and (2) \textbf{\textit{experiment-oriented}} analysis, which focuses on analyzing empirical results, experimental settings, and observed trends presented in tables or figures.
As shown in Fig.~\ref{fig:exp:ablation_complexity_analysis:objective}, the experimental results demonstrate a clear and consistent performance gap between these two analysis objectives across all evaluated agents.
The phenomenon where agents in general perform worse on \textbf{\textit{methodology-oriented}} analysis than on \textbf{\textit{experiment-oriented}} analysis ($\Delta_{\textsc{abs}} \geq \downarrow1.95\%$) indicates that accurately understanding and explaining methodological concepts and theoretical objectives is essentially more challenging than analyzing empirical evidence. This gap is observed consistently across different agent variants, suggesting a general limitation of current MLLM agents in methodology-level reasoning.
Despite this challenge, our approach improves performance on both methodology-oriented and experiment-oriented analysis tasks, with further gains achieved through finetuning. More importantly, comparing to baselines ($\Delta_{\textsc{abs}} \geq 4.91\%$), the performance gap between the two analysis objectives is notably reduced for \ours ($\Delta_{\textsc{abs}} \leq 3.60\%$) and finetuned \ours ($\Delta_{\textsc{abs}} \leq 3.29\%$). This observation suggests that the extended task-specific context and domain-specific knowledge collectively enable agents to better understand fundamental methods, clarify analytical objectives, and perform more coherent and objective-driven scientific analysis.
Overall, these results indicate that, while methodology-oriented analysis remains inherently more difficult than experiment-oriented analysis, providing richer contextual grounding and structured collaborative reasoning can enhance agents’ abilities in reasoning about both methodological contents and experimental evidence. The narrowed performance gap further underscores the significance of objective-aware contextual support for advancing scientific analysis capabilities beyond empirical result interpretation.

\input{figures/exp_human_domain_eval}

\subsection{Human Expert Evaluation of Domain Analysis}
\label{appendix:extended_experiment_analysis:human_domain_evaluation}


To more comprehensively assess the quality of scientific analysis writing, we incorporate an additional evaluation conducted by domain experts.
Specifically, human evaluators with expertise in computer science manually assess 100 agent-generated analyses in their respective domains.
Each analysis is generated based on explicit human requirements (\S\ref{fig:abation:requirement_prompting}) and evaluated through the same five-dimensional evaluation protocol (\S\ref{appendix:evaluation:five_dimension}).
Results in Fig.~\ref{fig:exp:human_domain_eval} show human assessment on two models in \textit{computer science} domain (Tab.~\ref{tab:data_domain}). The notable increase of "Good" analysis demonstrates significant performance gain achieved by \ours. Meanwhile, the decrease of five-dimensional errors further reveal the effective improvements on scientific analysis writing through systematic planning, reasoning, problem-solving, and reflective refinement (\S\ref{sec:anagent}).

\input{tables/human_eval}

\clearpage
\input{figures/exp_ablation_tool_call}
\input{figures/exp_ablation_tool_strategy}
\clearpage

To assess whether the improvements in Fig.~\ref{fig:exp:human_domain_eval} generalize beyond a single discipline, and to verify the quality of \ourbench's references directly, we extend the expert study to six domains: computer science, electronic engineering, biomedicine, mathematics, statistics, and economics (Tab.~\ref{tab:appendix:human_expert_multidomain}). In each domain, three experts independently rate $50$ agent-generated and $50$ gold analyses under the same five-dimensional protocol used by the MLLM judge (\S\ref{appendix:evaluation:five_dimension}), giving $600$ rated analyses in total. Analyses are produced by Qwen3-VL-4B, and the MLLM columns are scored by GPT-4.1-mini. As experts rate gold analyses under the identical protocol, model outputs and references are directly comparable on a single scale.

As shown in Tab.~\ref{tab:appendix:human_expert_multidomain}, \ours improves over the baseline on every dimension under expert rating, with the largest gains on grounding accuracy and analysis consistency, and every difference is significant at $p<0.001$ after Holm correction. In addition, inter-expert agreement is high throughout ($\alpha \geq 0.82$), so the ratings do not reflect individual annotator preference. Also, experts rate the gold analyses at $93.78\%$ overall, which empirically confirms the quality of \ourbench's author-written references and quantifies the remaining headroom between \ours and human-written scientific analysis.

\subsection{Tool Utilization}
\label{appendix:subsec:tool_use}

Equipped with five toolkits comprising sixteen tools in total (Tab.~\ref{tab:five_toolkits}), \ours extends its contextual awareness and domain expertise through autonomous tool invocation.
However, effective tool utilization requires accurate alignment between tool functionalities and task-specific execution objectives. Without such ability, agents exhibit substantially degraded performance due to non-strategic or failed tool calls (Fig.~\ref{fig:exp:ablation_tool_call}).

Fig.~\ref{fig:exp:ablation_tool_call} presents a comparative analysis between finetuned \ours (lighter colors) and training-free \ours (darker colors) across four backbone MLLMs.
Overall, finetuned \ours demonstrates a pronounced increase in the usage frequency of critical information-retrieval tools from the \textit{search toolkit}, such as \textit{abstract collector} ($\Delta_{\textit{rel}} \geq$ $\uparrow12.31\%$), \textit{information localizer} ($\Delta_{\textit{rel}} \geq$ $\uparrow61.27\%$), etc. This trend suggests that finetuning enables the model to more reliably identify when additional knowledge acquisition is necessary and to select appropriate tools accordingly.
Beyond increased tool engagement, finetuned \ours consistently achieves a markedly higher tool invocation success rate, accompanied by a substantial reduction in failed or invalid tool calls. These observations reveal that finetuning not only improves \ours’s awareness of \emph{which} tools to use, but also enhances \ours’s ability to correctly format and execute tool calls, thereby reducing execution-level errors and facilitating analysis generation.
More importantly, finetuned \ours exhibits clearly more \emph{strategic} and task-aware tool utilization. As illustrated by Fig.~\ref{fig:exp:ablation_tool_strategy}, in figure-centric scientific analysis tasks, finetuned \ours invokes tools from the \textit{vision toolkit} $\Delta_{\textit{rel}} \geq$ $\uparrow16.39\%$ more frequently than its training-free counterpart. This behavior reflects a stronger alignment between task objectives and tool selection, highlighting the effectiveness of finetuning in fostering objective-oriented reasoning and adaptive tool-use policies.
Taken together, these findings demonstrate that finetuning effectively enhances both the reliability and strategic capacity of tool utilization. The improved tool-call success rate and task-aware tool selection jointly contribute to the superior scientific analysis performance of \ours, meanwhile underscoring the critical role of objective-driven tool orchestration in multimodal agent systems.


\subsection{Controlled Experiments: Test-Time Scaling \& Compute Accounting}
\label{appendix:subsec:controlled_exps}

\input{tables/controlled_exps}

Two mechanisms could in principle explain \ours's gains without any contribution from multi-agent collaboration: the additional information retrieved by \Expert, and the larger inference-time compute consumed by multi-turn execution. \S\ref{appendix:extended_experiment_analysis:context_significance} addresses the first by supplying baselines with oracle context. Here we address the second directly, comparing \ours against single-model test-time scaling strategies whose call budgets match or exceed its own.

\paragraph{Controls.} We evaluate three standard single-model test-time scaling strategies on the same Qwen3-VL-4B backbone and the same evaluation set (\S\ref{appendix:extended_experiment_analysis:context_significance}). \textit{Self-Consistency} (SC@$N$) \citep{wang2023selfconsistency} samples $N$ candidate analyses and returns the one with the highest mean pairwise similarity to the remaining $N-1$. \textit{Best-of-$N$} (BoN@$N$) samples $N$ candidates and returns the one with the highest $S_{\textsc{Avg}}$. \textit{Self-Refine} (SR@$N$) \citep{madaan2023selfrefine} has a single \Critic iteratively critique and revise its own analysis for $N$ rounds. We set $N=10$ so that every control issues more MLLM calls than \ours's per-instance average of $6.64$. Because BoN selects using the evaluation metric itself, it is an oracle upper bound on best-of-$N$ selection rather than a deployable method. All methods are measured on the same hardware.

\paragraph{Additional inference compute alone does not produce gains.}
SC, BoN, and SR each issue $N=10$ MLLM calls, more than \ours's $6.64$. However, they improve over single-pass generation by only $\uparrow 0.11\%$ to $\uparrow 1.01\%$, against \ours's $\uparrow 7.36\%$ training-free and $\uparrow 34.60\%$ after modular training. This holds even though BoN is granted oracle selection, which upper-bounds what any best-of-$N$ rule could achieve at that budget. Spending more test-time compute on the base model, without structural change, does not reproduce the gains.

\paragraph{What tokens are spent on matters more than how many.}
SR is the largest token consumer among the controls, at $4.06\times$ tokens across $10$ calls, and still gains only $\uparrow 0.11\%$. \textit{Omnion}, \ours, and \ours (SFT) all operate in a comparable per-call regime of $4.38\times$ to $4.87\times$, yet span $\downarrow 14.54\%$ to $\uparrow 34.60\%$. SR is compute-neutral because resampling and self-critique redraw only from the model's own distribution and cannot introduce information the backbone lacks, while \textit{Omnion} shows that tool access without orchestration is actively harmful. Only orchestrated collaboration converts the budget into gains.

\paragraph{Training is orthogonal to inference cost.}
\ours (SFT) runs at essentially the same inference cost as its training-free counterpart, $6.79$ calls at $4.38\times$ tokens against $6.64$ calls at $4.52\times$, in fact slightly cheaper per call. This effectively reflects more reliable and more targeted tool calls after finetuning (\S\ref{appendix:subsec:tool_use}). It nonetheless raises the relative gain from $\uparrow 7.36\%$ to $\uparrow 34.60\%$. Together with the modular versus end-to-end comparison at identical inference cost (Tab.~\ref{tab:experiment:ablation_training_method_end2end_modular}), this indicates that the lever is the method rather than the inference budget.

\section{Example: \ours For Scientific Table \& Figure Scientific Analysis}

Here is an end-to-end scientific analysis writing example of \ours:

\input{figures/good_analysis_examples}

\section{Failure Analysis}
\label{appendix:sec:failure_analysis}

\input{figures/failure_analysis_error_decrease}

Following our preliminary exploration that reveals seven key error patterns in scientific table and figure analysis (Fig.~\ref{fig:preliminary-error-analysis}), we extend the evaluation to \ours, powered by the same backbone model \texttt{Qwen3-VL-8B} and tested on the identical 120-sample subset. Figure~\ref{fig:exp:failure_analysis:error_distribution} compares and visualizes the distributions of the seven error patterns between baseline and \ours.
As can be seen from Fig.~\ref{fig:exp:failure_analysis:error_distribution}, \ours consistently reduces error rates across all seven categories ($\Delta_{\textit{rel}} \geq$ $\downarrow3.28\%$), demonstrating the effectiveness of integrating high-level planning with low-level knowledge acquisition and problem-solving through multi-agent collaboration for enhancing scientific table and figure analysis.

Building on these seven error patterns (Figs.~\ref{fig:preliminary-error-analysis}-\ref{fig:exp:failure_analysis:error_distribution}), we conduct a systematic failure analysis and identify seven representative failure types corresponding to each error category, aiming to provide meaningful insights to inform future research.

\subsection{Multi- Modality \& Layout Perception Error}
\label{appendix:subsec:failure_analysis:error1}

Perception errors are a major obstacle to accurate multimodal reasoning in MLLMs \cite{wang2025papo}.
Although perception errors are notably decreased ($\Delta_{\textit{rel}}=\downarrow11.94\%$) by \ours, there are still cases where \ours's scientific analysis receives low evaluation scores as a result of perception errors.
Fig.~\ref{fig:failure_analysis:error1_perception_error} shows an example of a scientific analysis failure case ($S_{\textsc{AVG}}=25.64\%$) with \textit{multi- modality \& layout error}.

\subsection{Hallucination Error}
\label{appendix:subsec:failure_analysis:error2}

Hallucination remains a known challenge in MLLMs and poses a significant source of errors for scientific table \& figure analysis that requires faithful and reliable scientific generation.
Although hallucination errors are remarkably reduced ($\Delta_{\textit{rel}}=\downarrow32.65\%$) by \ours, there are still cases where \ours's scientific analysis writing receives low evaluation scores due to hallucination errors.
Fig.~\ref{fig:failure_analysis:error2_hallucination_error} shows an example of a scientific analysis failure case ($S_{\textsc{AVG}}=21.85\%$) with \textit{hallucination error}.

\subsection{Domain-Specific Error}
\label{appendix:subsec:failure_analysis:error3}

Considering the domain-specific nature of scientific literature, acquiring accurate understanding of domain-specific tables and figures remains a significant challenge for MLLMs.
While \ours markedly reduces domain-specific errors ($\Delta_{\textit{rel}}=\downarrow27.03\%$), there are still cases where \ours's scientific analysis writing receives low evaluation scores due to limited domain knowledge or misunderstandings of domain-specific contents.
Fig.~\ref{fig:failure_analysis:error3_domain_error} shows an example of a scientific analysis failure case ($S_{\textsc{AVG}}=11.28\%$) with \textit{domain-specific error}.

\subsection{Long-Context Understanding Error}
\label{appendix:subsec:failure_analysis:error4}

Scientific literature usually involves long-horizon contexts and cross-document references, posing significant challenges for MLLMs to effectively process, reason, and comprehend.
While \ours is able to reduce errors stemming from long-context understanding ($\Delta_{\textit{rel}}=\downarrow3.28\%$), it remains a high error rate (Fig.~\ref{fig:exp:failure_analysis:error_distribution}) for \ours.
For example, Fig.~\ref{fig:failure_analysis:error4_long_context_error} shows a scientific analysis failure case ($S_{\textsc{AVG}}=24.43\%$) as a result of \textit{long-context understanding error}.

\subsection{Analysis Width Error}
\label{appendix:subsec:failure_analysis:error5}

Scientific literature contains heterogeneous elements, such as tables, figures, sections, citations, etc., that introduce varying scopes of information coverage.
Although \ours reduces errors stemming from analysis width ($\Delta_{\textit{rel}}=\downarrow13.79\%$), it still exhibits a relatively high error rate (Fig.~\ref{fig:exp:failure_analysis:error_distribution}).
One representative failure case is shown in Fig.~\ref{fig:failure_analysis:error5_width_error}, where \ours achieves a low scientific analysis score ($S_{\textsc{AVG}}=25.96\%$) due to \textit{analysis width errors}.

\subsection{Analysis Depth Error}
\label{appendix:subsec:failure_analysis:error6}

The roles and implications of tables and figures in scientific literature impose varying demands on the \emph{depth} of analysis. Some only require shallow summarization directly derived from tabular or visual content, while others demand deeper reasoning, such as interpreting trends, drawing comparisons, or inferring conclusions supported by the presented evidence.
Although \ours reduces errors stemming from analysis depth ($\Delta_{\textit{rel}}=\downarrow9.78\%$), it still exhibits a relatively high error rate (Fig.~\ref{fig:exp:failure_analysis:error_distribution}).
One representative failure case is shown in Fig.~\ref{fig:failure_analysis:error6_depth_error}, where \ours attains a low scientific analysis score ($S_{\textsc{AVG}}=31.24\%$) due to \textit{analysis depth errors}.

\subsection{Analysis Objective Error}
\label{appendix:subsec:failure_analysis:error7}

Tables and figures in scientific literature serve diverse analytical objectives.
Some are designed to illustrate methodological designs, some to highlight benchmark innovations, and some to present empirical evidence in support of hypothetical claims, among a variety of scientific analysis objectives.
Although \ours reduces errors related to analysis objectives ($\Delta_{\textit{rel}}=\downarrow9.78\%$), correctly identifying and fulfilling these objectives remains challenging for MLLM agents (Fig.~\ref{fig:exp:failure_analysis:error_distribution}). 
A representative failure case is shown in Fig.~\ref{fig:failure_analysis:error7_objective_error}, where \ours achieves a low scientific analysis score ($S_{\textsc{AVG}}=31.45\%$) due to \textit{analysis objective errors}.

\subsection{Other Errors Due To MLLM Backbone's Ability Constraints}
\label{appendix:subsec:failure_analysis:error8}

The heterogeneity of multimodal content and the requirement for long-context understanding pose significant challenges for MLL agents. These challenges give rise to fundamental errors that profoundly impair scientific analysis, while stemming from the inherent ability constraints of the backbone MLLMs.
Fig.~\ref{appendix:subsec:failure_analysis:error8} illustrates two representative types of failures observed across different MLLMs: (a) \textit{repetitive content} in \ours's output, and (b) incorrectly generated \textit{intermediate solutions} by upstream agents prior to the scientific analysis writing performed by \Solver.  These phenomena reveal the significance of backbone MLLM’s capacity in multi-agent systems for effective global context modeling, logical consistency, and accurate information consolidation.

\input{sections/limitations}

\input{figures/failure_analysis_examples}

\input{figures/prompt_evaluation}
\input{figures/prompt_agent}

%% file: figures/preliminary-error-analysis.tex
\begin{wrapfigure}{r}{0.5\linewidth}
\centering
\includegraphics[width=\linewidth]{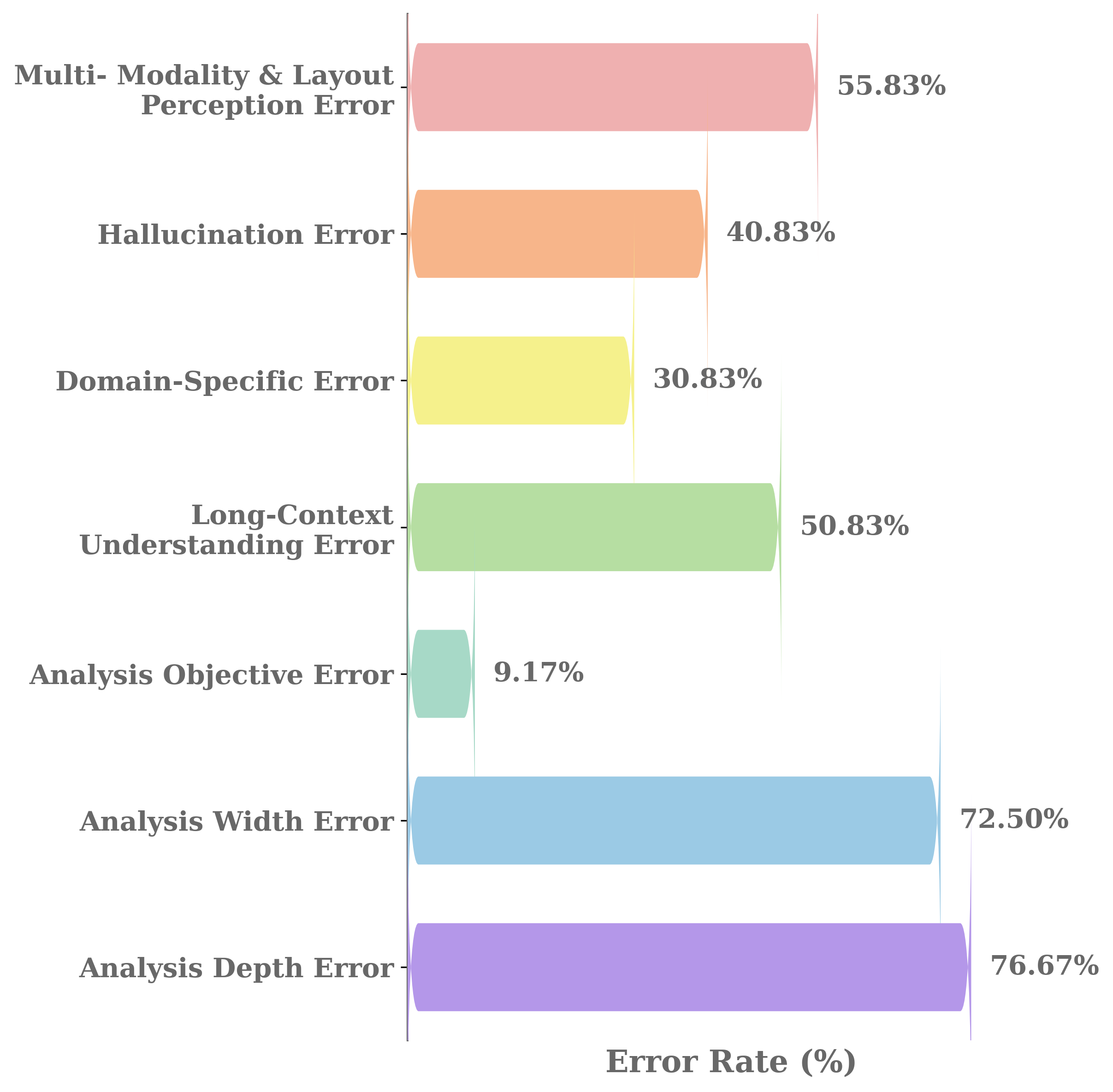}


\caption{\textbf{Key Error Patterns.} Through our case study on 120 samples, we identify seven key error patterns that correspond to the analysis challenges shown in Fig.~\ref{fig:preliminary}.}
\label{fig:preliminary-error-analysis}

\vspace{-6pt}
\end{wrapfigure}

%% file: figures/preliminary-pattern-1-type.tex
\begin{figure*}[!t]
\centering
\includegraphics[width=1.0\linewidth]{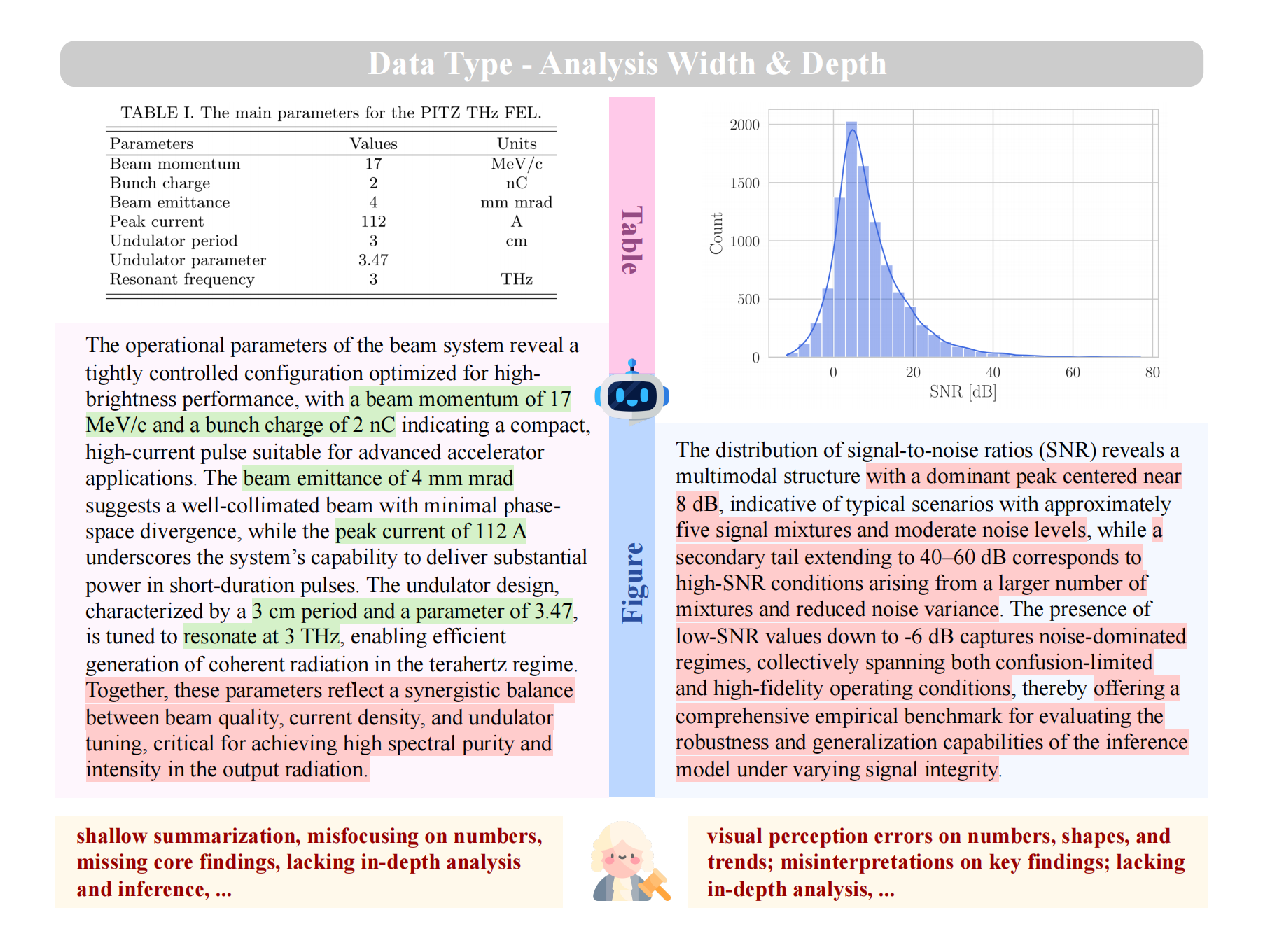}

\vspace{-6pt}
\caption{\textbf{Preliminary Analysis On Failure Patterns.} Analyzing on two different data types, \textit{table} \& \textit{figure}, the agent shows its lack of accurate visual perception and its incapability of writing with proper analysis depth and width.}
\label{fig:preliminary-pattern-1-type}
\end{figure*}

%% file: figures/preliminary-pattern-2-domain.tex
\begin{figure*}[!t]
\centering
\includegraphics[width=1.0\linewidth]{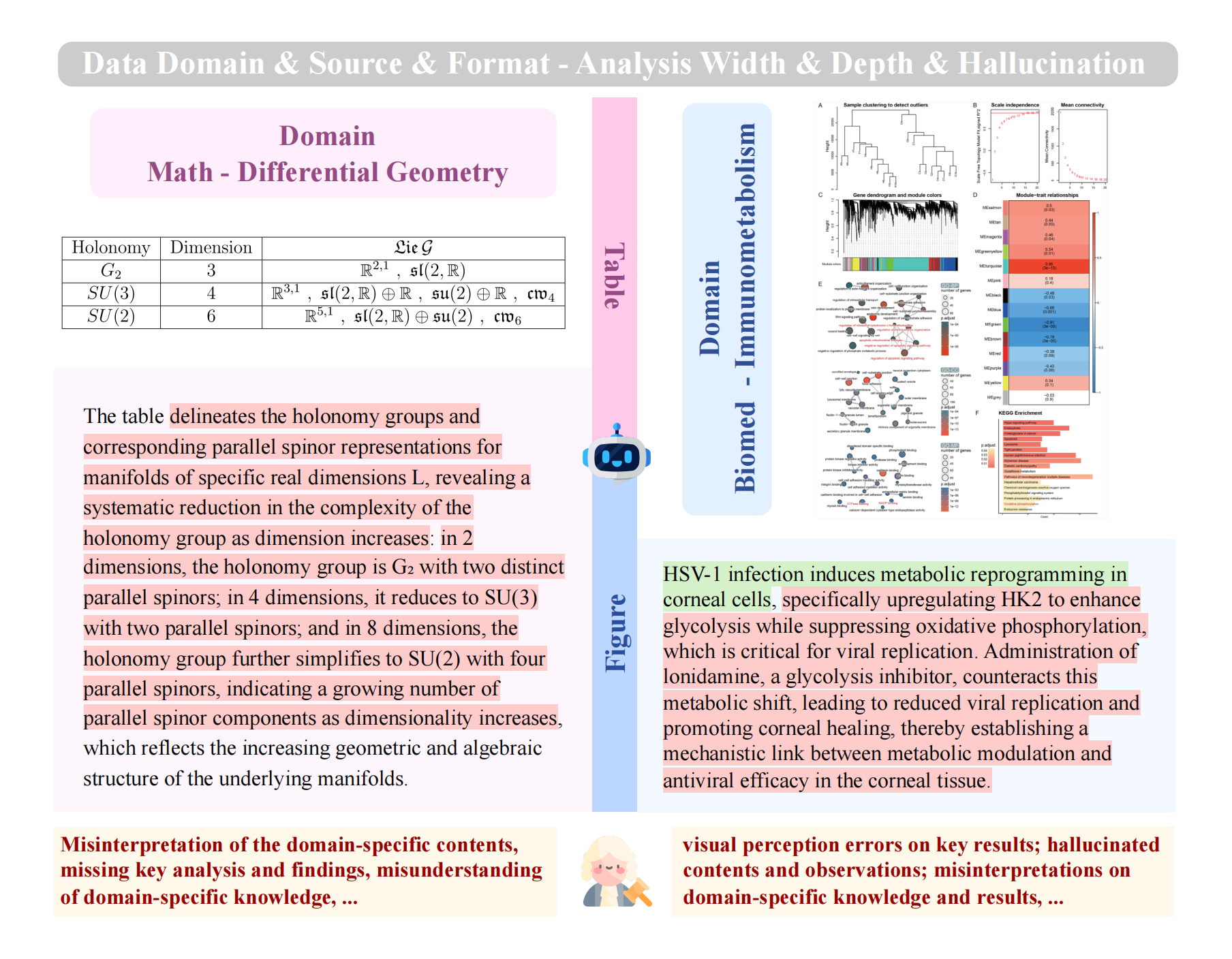}

\vspace{-6pt}
\caption{\textbf{Preliminary Analysis On Failure Patterns.} Analyzing on two data types with different data sources, formats, and domains, the agent generates analysis with significant hallucinated contents, meanwhile incapable of writing with proper analysis depth and width.}
\label{fig:preliminary-pattern-2-domain}
\end{figure*}

%% file: figures/teaser.tex
\begin{wrapfigure}{r}{0.5\linewidth}
\centering
\includegraphics[width=\linewidth]{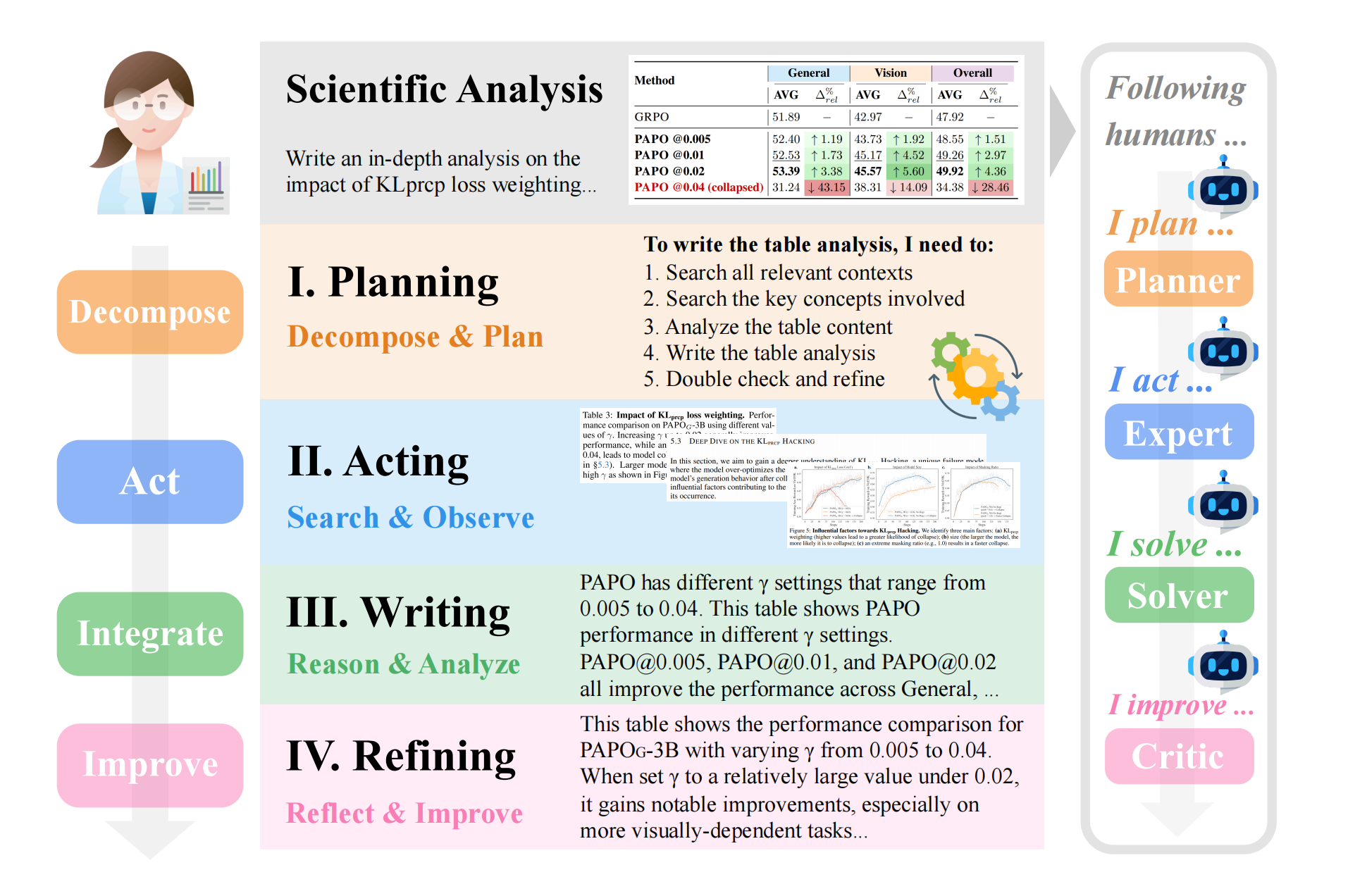}


\caption{\textbf{Scientific Analysis Workflow.} Motivated by how human researchers perform scientific analysis, we decompose the scientific analysis workflow into dedicated stages, which leads to \ours (Fig.~\ref{fig:anagent}).}
\label{fig:teaser}

\vspace{-6pt}
\end{wrapfigure}

%% file: sections/6_related.tex
\section{Related Work}
\label{sec:related_work}

\paragraph{AI For Scientific Table Understanding.}
\label{subsec:related_work:table}
Recent advances in AI greatly inspire research on table understanding \cite{erickson2025tabarena}, particularly scientific tables that exhibit diverse formats, layouts, domains, and analytical objectives.
As tables constitute a compact yet information-dense medium for conveying methodological details and empirical findings, benchmarks are proposed to evaluate distinct aspects of scientific table understanding: SCITAB \cite{lu-etal-2023-scitab} assesses table-based claim verification, S2abEL \cite{lou2023s2abel} targets entity linking, and other benchmarks address question answering (QA) \cite{pramanick2024spiqa,zhang2025scitat}, table-to-text generation \cite{bai_table_reasoning_2025}, literature-to-table \cite{newman-etal-2024-arxivdigestables}, etc.
However, existing benchmarks emphasize isolated tasks while lack principled curriculum to capture diverse data heterogeneity and reasoning complexity in long-horizon contexts, motivating our work to benchmark scientific table understanding across multiple complexity dimensions.
%

\paragraph{AI For Scientific Multimodal Understanding.}
\label{subsec:related_work:figure}
Scientific papers are inherently multimodal, combining text with figures, tables, algorithms, etc., to communicate complex scientific evidence \cite{zheng-etal-2025-automation, zhang_survey_2024}. Accordingly, multimodal reasoning and long-context comprehension are essential for scientific research.
However, existing benchmarks have significant limitations: SPIQA \cite{pramanick2024spiqa} for table \& figure QA shows limited coverage of cross-domain generalization and reasoning complexity. WildSci \cite{liu2026wildsci} targets QA across domains, yet fails to incorporate multimodal long-context reasoning that is fundamental to scientific inquiry. These limitations motivate \ourbench with structured reasoning curriculum to provide more comprehensive testbed for enhancing multimodal scientific understanding.

%% file: figures/exp_ablation_train_data_size_domain_format_type.tex
\begin{wrapfigure}{r}{0.65\linewidth}

\vspace{-6pt}

\centering
\includegraphics[width=\linewidth]{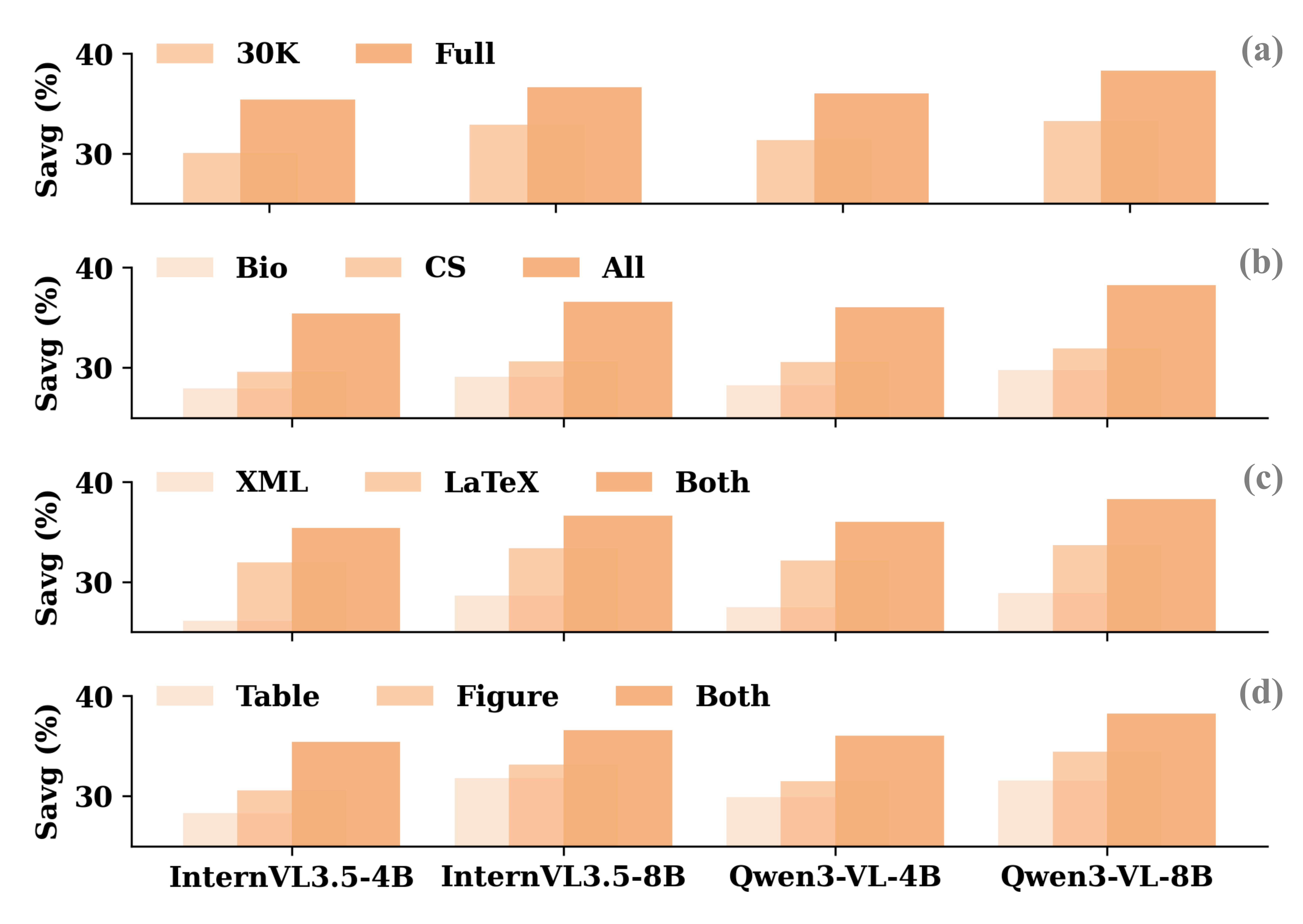}

\vspace{-3pt}

\caption{\textbf{Ablations On Training Data.} Performance visualization of ablation studies (\S\ref{subsec:results:ablations_on_params}), respectively on: (a) \textit{data size}, (b) \textit{data domain}, (c) \textit{data format}, (d) \textit{data type} (\S\ref{appendix:benchmark:curriculum:data_complexity}).}

\label{fig:exp:ablation_data_size_domain_format_type}

\vspace{6pt}

\end{wrapfigure}

%% file: figures/anabench_year_distribution.tex
\begin{wrapfigure}{r}{0.4\linewidth}

\vspace{-12pt}

\centering
\includegraphics[width=\linewidth]{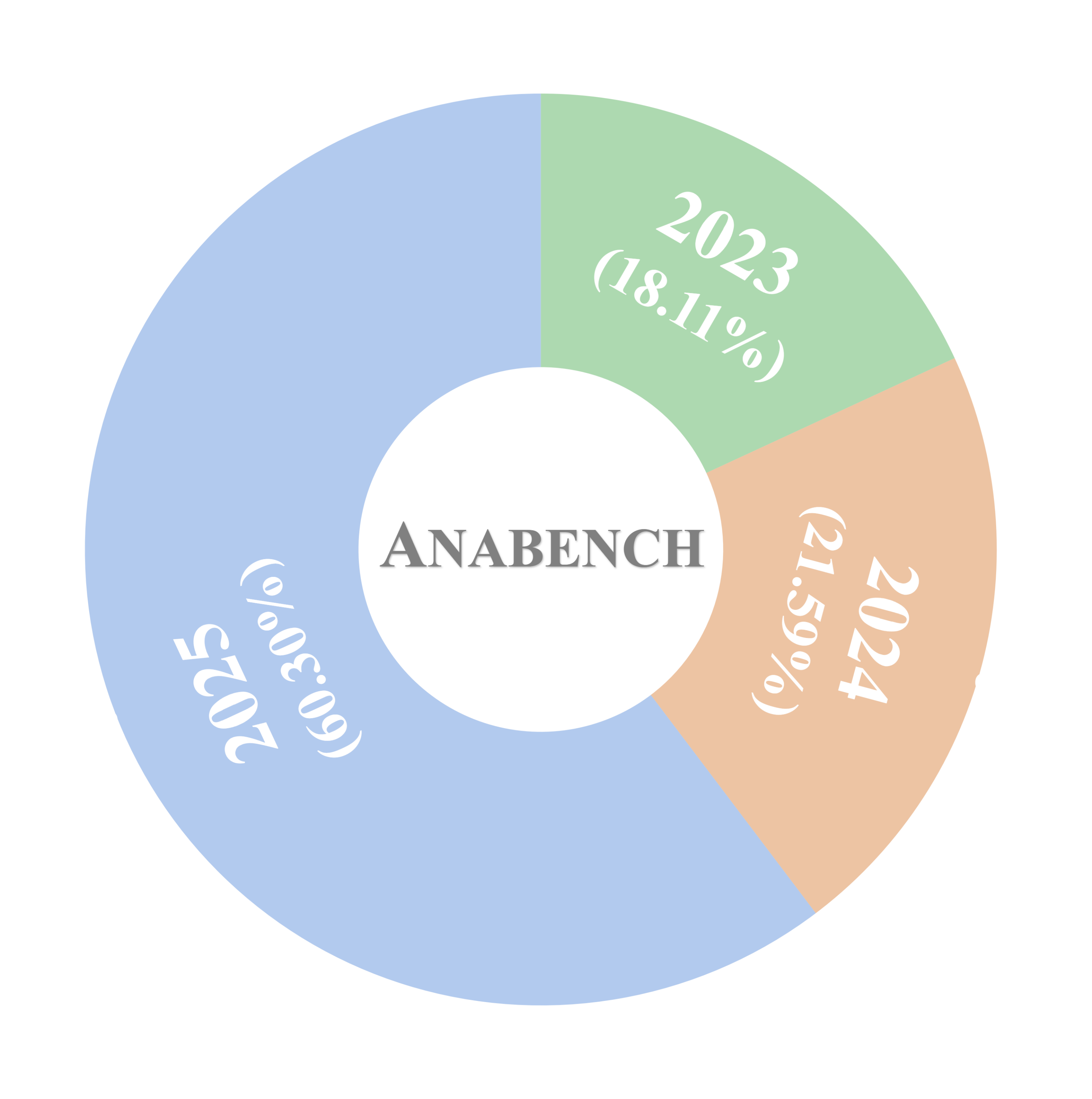}

\vspace{-12pt}

\caption{\textbf{Year Distribution of \ourbench.} Visualization of year distribution, with 2025 comprising the largest proportion to mitigate data contamination (\S\ref{appendix:anabench}).}
\label{fig:anabench_year_distribution}

\vspace{6pt}

\end{wrapfigure}

%% file: tables/algorithm1.tex
\begin{algorithm}[t]
\setstretch{1.1}
\caption{$k$-Depth Hierarchical Context Retrieval}
\label{alg:k_depth_context}
\begin{algorithmic}[1]
\Require Target data instance $d$, maximum context depth $k$, reference graph $\mathcal{G}$
\Ensure Hierarchical context set $\mathcal{C} = \{\mathcal{C}_1, \ldots, \mathcal{C}_k\}$
\State Initialize $\mathcal{C} \gets \emptyset$
\State Initialize visited set $\mathcal{V} \gets \{d\}$
\State Initialize frontier $\mathcal{F}_0 \gets \{d\}$
\For{$i = 1$ to $k$}
    \State Initialize $\mathcal{F}_i \gets \emptyset$
    \ForAll{$e \in \mathcal{F}_{i-1}$}
        \State Retrieve $\mathcal{R}_{\text{in}}(e)$ and $\mathcal{R}_{\text{out}}(e)$ from $\mathcal{G}$
        \ForAll{$e' \in \mathcal{R}_{\text{in}}(e) \cup \mathcal{R}_{\text{out}}(e)$}
            \If{$e' \notin \mathcal{V}$}
                \State Add $e'$ to $\mathcal{F}_i$
                \State Add $e'$ to $\mathcal{V}$
            \EndIf
        \EndFor
    \EndFor
    \State $\mathcal{C}_i \gets \mathcal{F}_i$
\EndFor
\State \Return $\mathcal{C}$
\end{algorithmic}
\end{algorithm}

%% file: tables/benchmark_comparison.tex
\begin{table}[H]

\vspace{36pt}

\centering
\setstretch{1.1}
\small 
\setlength{\tabcolsep}{4pt} 

\caption{\textbf{Benchmark Comparison.} We compare \ourbench with existing scientific benchmarks across multiple dimensions of data complexity (M-mod.: multimodal, M-Layout: multi-layout, M-Doc.: multi-document, M-Src.: multi-source, M-Fmt.: multi-format, M-Dom.: multi-domain) and reasoning complexity (Long-Ctx.: long-context reasoning, M-Width: multi-width, M-Depth: multi-depth, M-ObJ.: multi-objective), highlighting the comprehensiveness of our approach to evaluating autonomous scientific analysis capabilities of MLLM-powered scientific agents. Note: 9/26 denotes 9 domains and 26 subdomains; 9/170 denotes 9 domains and 170 subdomains. Compared benchmarks are listed in alphabetical order.}
\label{tab:benchmark_comparison}

\begin{tabular}{>{\centering\arraybackslash}m{1.6cm}
                >{\centering\arraybackslash}m{1.2cm}
                >{\centering\arraybackslash}m{1.2cm}
                >{\centering\arraybackslash}m{0.8cm}
                >{\centering\arraybackslash}m{1cm}
                >{\centering\arraybackslash}m{1cm}
                >{\centering\arraybackslash}m{0.8cm}
                >{\centering\arraybackslash}m{0.8cm}
                >{\centering\arraybackslash}m{0.8cm}
                >{\centering\arraybackslash}m{0.8cm}
                >{\centering\arraybackslash}m{0.8cm}
                >{\centering\arraybackslash}m{0.8cm}
                >{\centering\arraybackslash}m{0.8cm}}
                
\toprule

\multirow{2}{*}{\textbf{Benchmark}} & 
\multirow{2}{*}{\textbf{Task}} & 
\multirow{2}{*}{\textbf{Source}} & 
\multicolumn{6}{c}{\textbf{Data Complexity}} & 
\multicolumn{4}{c}{\textbf{Reasoning Complexity}} \\
\cmidrule(lr){4-9} \cmidrule(lr){10-13}
& & & 
\textbf{M-mod.} & 
\textbf{M-Layout} & 
\textbf{M-Doc.} & 
\textbf{M-Src.} & 
\textbf{M-Fmt.} & 
\textbf{M-Dom.} & 
\textbf{Long-Ctx.} & 
\textbf{M-Width} & 
\textbf{M-Depth} & 
\textbf{M-Obj.} \\

\midrule

\textbf{M3SciQA} \cite{li2024m3sciqa} & \textit{QA} & \textit{Partial} & \cmark & \cmark & \cmark & \xmark & \xmark & \xmark & \xmark & \xmark & \xmark & \xmark \\

\textbf{SCIDQA} \cite{singh-etal-2024-scidqa} & \textit{QA} & \textit{Full} & \xmark & \xmark & \cmark & \xmark & \xmark & \xmark & \cmark & \xmark & \xmark & \xmark \\

\textbf{SCITAB} \cite{lu-etal-2023-scitab} & \textit{Claim Verification} & \textit{Partial} & \xmark & \xmark & \xmark & \xmark & \xmark & \xmark & \xmark & \xmark & \xmark & \xmark \\

\textbf{SCITAT} \cite{zhang2025scitat} & \textit{QA} & \textit{Partial} & \xmark & \xmark & \xmark & \xmark & \xmark & \xmark & \xmark & \xmark & \xmark & \xmark \\

\textbf{SPIQA} \cite{pramanick2024spiqa} & \textit{QA} & \textit{Full} & \cmark & \cmark & \xmark & \cmark & \xmark & \xmark & \cmark & \xmark & \xmark & \xmark \\

\textbf{S2abEL} \cite{lou2023s2abel} & \textit{Entity Link} & \textit{Partial} & \xmark & \xmark & \xmark & \xmark & \xmark & \xmark & \xmark & \xmark & \xmark & \xmark \\

\textbf{PubMedQA} \cite{jin2019pubmedqa} & \textit{QA} & \textit{Partial} & \xmark & \xmark & \xmark & \xmark & \xmark & \xmark & \xmark & \xmark & \xmark & \xmark \\

\textbf{WildSci} \cite{liu2026wildsci} & \textit{QA} & \textit{Partial} & \xmark & \xmark & \xmark & \xmark & \xmark & \cmark (9/26) & \xmark & \xmark & \xmark & \xmark \\

\midrule

\textbf{\ourbench} & \textit{Scientific Analysis} & \textit{Full} & \cmark & \cmark & \cmark & \cmark & \cmark & \cmark (9/170) & \cmark & \cmark & \cmark & \cmark \\

\bottomrule
\end{tabular}

\vspace{36pt}

\end{table}

%% file: tables/benchmark_complexity_curriculum.tex
\definecolor{DATA_COMPLEXITY_BG}{HTML}{ffe9fd}
\definecolor{ANALYSIS_COMPLEXITY_BG}{HTML}{def0ff}


\begin{table}[H]
\small
\centering

\setstretch{1.1}

\caption{\textbf{\ourbench Complexity Curriculum.}  This table summarizes the \textit{task classification} stage of our benchmark construction method for the complexity curriculum of \ourbench, which defines 23 task complexity categories across seven data complexity and analysis complexity dimensions.}
\label{tab:anabench_complexity_curriculum}
\begin{tabularx}{\textwidth}{>{\centering\arraybackslash}m{1.9cm}
                >{\centering\arraybackslash}m{1.5cm}
                >{\centering\arraybackslash}m{2cm}
                X}
                
\toprule

\textbf{Curriculum} & \textbf{Broad} & \textbf{Fine-Grained} & \multicolumn{1}{c}{\textbf{Category Definition}} \\

\midrule

\cellcolor{DATA_COMPLEXITY_BG}
& \cellcolor{DATA_COMPLEXITY_BG}
& \cellcolor{DATA_COMPLEXITY_BG}\raisebox{-2.5em}{\textbf{Table}}
& \cellcolor{DATA_COMPLEXITY_BG}{\logo Tabular-structured data that organize information into rows and columns to systematically present structured contents, such as method characteristics, parameter configurations, benchmark comparisons, categorical classifications, experimental results, quantitative or qualitative summaries, and analytical breakdowns, enabling precise lookup, comparison, and reference.} \\

\cellcolor{DATA_COMPLEXITY_BG}
& \cellcolor{DATA_COMPLEXITY_BG}\multirow{-2}{1.5cm}{\centering\textbf{Type}}
& \cellcolor{DATA_COMPLEXITY_BG}\raisebox{-2.5em}{\textbf{Figure}}
& \cellcolor{DATA_COMPLEXITY_BG}{\logo Visual representations of scientific information that employ graphical, diagrammatic, or illustrative elements, such as plots, charts, schematics, diagrams, images, framework architectures, etc., to show relationships, patterns, processes, structures, conceptual approaches, and other critical information that may be complex or inefficient to express in textual form} \\

\cline{2-4}

\cellcolor{DATA_COMPLEXITY_BG}
& \cellcolor{DATA_COMPLEXITY_BG}
& \cellcolor{DATA_COMPLEXITY_BG}\raisebox{-1.2em}{\textbf{LaTeX}}
& \cellcolor{DATA_COMPLEXITY_BG}{\logo Scientific data obtained from documents in LaTeX, which is a structured format widely used in scholarly publishing to encode mathematical expressions, tables, figures, algorithms, and document structures} \\

\cellcolor{DATA_COMPLEXITY_BG}\multirow{-4}{2cm}[-2.0em]{\centering\textbf{Data Complexity}}
& \cellcolor{DATA_COMPLEXITY_BG}\multirow{-2}{1.5cm}[-0.8em]{\centering\textbf{Format}}
& \cellcolor{DATA_COMPLEXITY_BG}\raisebox{-1.8em}{\textbf{XML}}
& \cellcolor{DATA_COMPLEXITY_BG}{\logo Scientific data obtained from documents encoded in XML, which is a structured, machine-readable format that represents document content, metadata, and hierarchical relationships through tagged elements, commonly used for standardized archival and data exchange in scholarly publishing} \\

\cline{2-4}

\cellcolor{DATA_COMPLEXITY_BG}
& \cellcolor{DATA_COMPLEXITY_BG}
& \cellcolor{DATA_COMPLEXITY_BG}\raisebox{-1.8em}{\textbf{General}}
& \cellcolor{DATA_COMPLEXITY_BG}{\logo Scientific literature that primarily focuses on the development, analysis, or evaluation of a specific method, model, framework, or benchmark, typically proposing novel techniques or reporting empirical results within a defined research problem.} \\

\cellcolor{DATA_COMPLEXITY_BG}
& \cellcolor{DATA_COMPLEXITY_BG}\multirow{-2}{1.5cm}[0.0em]{\centering\textbf{Source}}
& \cellcolor{DATA_COMPLEXITY_BG}\raisebox{-1.8em}{\textbf{Review}}
& \cellcolor{DATA_COMPLEXITY_BG}{\logo Scientific literature that systematically examines, summarizes, and synthesizes existing research within a specific domain or research direction, aiming to provide an overview of prior work, identify trends, compare approaches, and highlight open challenges for future research.} \\

\cline{2-4}

\cellcolor{DATA_COMPLEXITY_BG}
& \cellcolor{DATA_COMPLEXITY_BG}\raisebox{-0.6em}{\textbf{Domain}}
& \cellcolor{DATA_COMPLEXITY_BG}\raisebox{-0.6em}{\textbf{9/170}}
& \cellcolor{DATA_COMPLEXITY_BG}{\logo Scientific data can span multiple research domains, among which \ourbench covers 9 broad domains encompassing 170 fine-grained disciplinary categories} \\

\midrule

\cellcolor{ANALYSIS_COMPLEXITY_BG}
& \cellcolor{ANALYSIS_COMPLEXITY_BG}
& \cellcolor{ANALYSIS_COMPLEXITY_BG}\raisebox{-0.6em}{\textbf{Methodology}}
& \cellcolor{ANALYSIS_COMPLEXITY_BG}{\logo Scientific analysis that explains and rationalize the proposed method, such as model designs, training algorithms, framework architectures, etc.} \\

\cellcolor{ANALYSIS_COMPLEXITY_BG}
& \cellcolor{ANALYSIS_COMPLEXITY_BG}\multirow{-2}{1.5cm}[-1.0em]{\centering\textbf{Objective}}
& \cellcolor{ANALYSIS_COMPLEXITY_BG}\raisebox{-1.8em}{\textbf{Experiment}}
& \cellcolor{ANALYSIS_COMPLEXITY_BG}{\logo Scientific analysis that focuses on empirically evaluating and analyzing diverse aspects like performance, robustness, trustworthiness, comparative effectiveness, etc., across different datasets, settings, or baselines using quantitative and qualitative results} \\

\cline{2-4}

\cellcolor{ANALYSIS_COMPLEXITY_BG}
& \cellcolor{ANALYSIS_COMPLEXITY_BG}
& \cellcolor{ANALYSIS_COMPLEXITY_BG}\raisebox{-0.6em}{\textbf{Self-Contained}}
& \cellcolor{ANALYSIS_COMPLEXITY_BG}{\logo Scientific analysis that relies solely on the given input, without referring to additional sources} \\

\cellcolor{ANALYSIS_COMPLEXITY_BG}
& \cellcolor{ANALYSIS_COMPLEXITY_BG}
& \cellcolor{ANALYSIS_COMPLEXITY_BG}\raisebox{-0.6em}{\textbf{Internal}}
& \cellcolor{ANALYSIS_COMPLEXITY_BG}{\logo Scientific analysis that integrates the given input with references drawn from within the same source paper} \\

\cellcolor{ANALYSIS_COMPLEXITY_BG}
& \cellcolor{ANALYSIS_COMPLEXITY_BG}
& \cellcolor{ANALYSIS_COMPLEXITY_BG}\raisebox{-0.6em}{\textbf{External}}
& \cellcolor{ANALYSIS_COMPLEXITY_BG}{\logo Scientific analysis that integrates the given input with references drawn from sources outside the source paper} \\

\cellcolor{ANALYSIS_COMPLEXITY_BG}
& \cellcolor{ANALYSIS_COMPLEXITY_BG}\multirow{-4}{1.5cm}[1.2em]{\centering\textbf{Width}}
& \cellcolor{ANALYSIS_COMPLEXITY_BG}\raisebox{-0.6em}{\textbf{Mixed}}
& \cellcolor{ANALYSIS_COMPLEXITY_BG}{\logo Scientific analysis that combines the given input with references drawn from both within and outside the source paper} \\

\cline{2-4}

\cellcolor{ANALYSIS_COMPLEXITY_BG}
& \cellcolor{ANALYSIS_COMPLEXITY_BG}
& \cellcolor{ANALYSIS_COMPLEXITY_BG}\raisebox{-1.2em}{\textbf{Shallow}}
& \cellcolor{ANALYSIS_COMPLEXITY_BG}{\logo Scientific analysis that focuses on surface-level observations, straightforward patterns, or direct summarization of the input, without extended reasoning, interpretation, or inference} \\

\cellcolor{ANALYSIS_COMPLEXITY_BG}\multirow{-8}{2cm}[5.0em]{\centering\textbf{Analysis\\Complexity}}
& \cellcolor{ANALYSIS_COMPLEXITY_BG}\multirow{-2}{1.5cm}[0.0em]{\centering\textbf{Depth}}
& \cellcolor{ANALYSIS_COMPLEXITY_BG}\raisebox{-1.2em}{\textbf{In-Depth}}
& \cellcolor{ANALYSIS_COMPLEXITY_BG}{\logo Scientific analysis that involves extended reasoning, deeper interpretation, and evidence-grounded inference beyond surface-level summarization to derive deeper insights, explanations, findings, or conclusions} \\

\bottomrule

\end{tabularx}
\end{table}

%% file: figures/anabench_distribution.tex

\begin{figure}[H]
\centering
\includegraphics[width=1.0\linewidth]{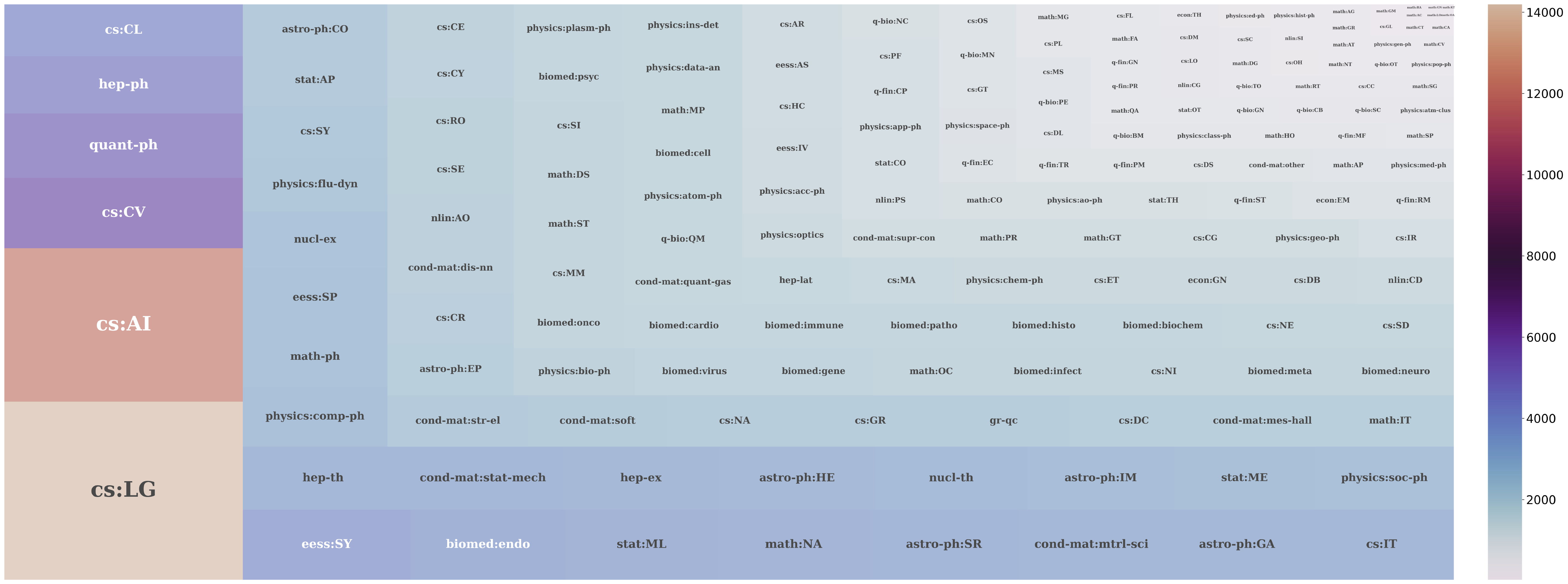}

\caption{\textbf{\ourbench Domain Distribution.} Domain distribution of \ourbench across 9 broad domain categories (Fig.~\ref{fig:broad_domain_distribution}.a) and 170 fine-grained subdomain categories (Fig.~\ref{fig:anabench_domain_distribution}).}
\label{fig:anabench_domain_distribution}

\vspace{-12pt}

\end{figure}


\begin{figure}[H]
\centering
\includegraphics[width=1.0\linewidth]{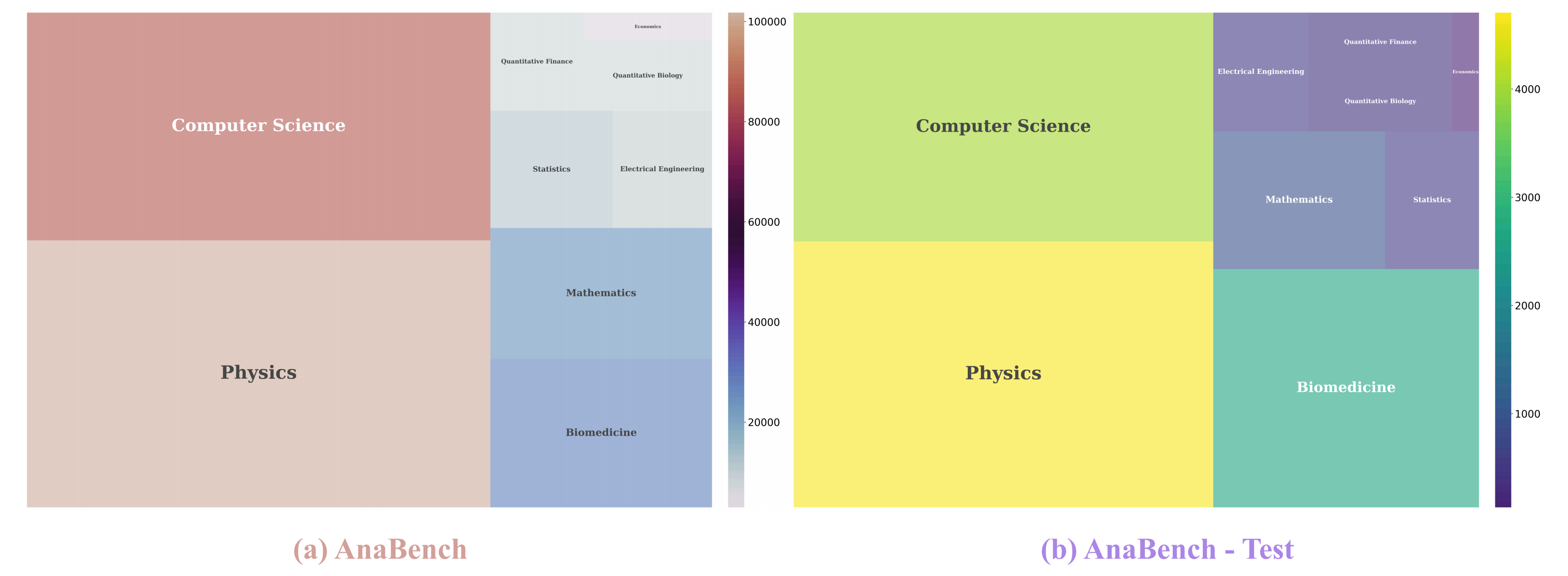}

\caption{\textbf{Broad Domain Distribution.} Broad domain distribution of \textbf{(a) \ourbench} and \textbf{(b) the evaluation set of \ourbench} across nine broad domain categories.}
\label{fig:broad_domain_distribution}

\vspace{-12pt}

\end{figure}


\begin{figure}[H]
\centering
\includegraphics[width=1.0\linewidth]{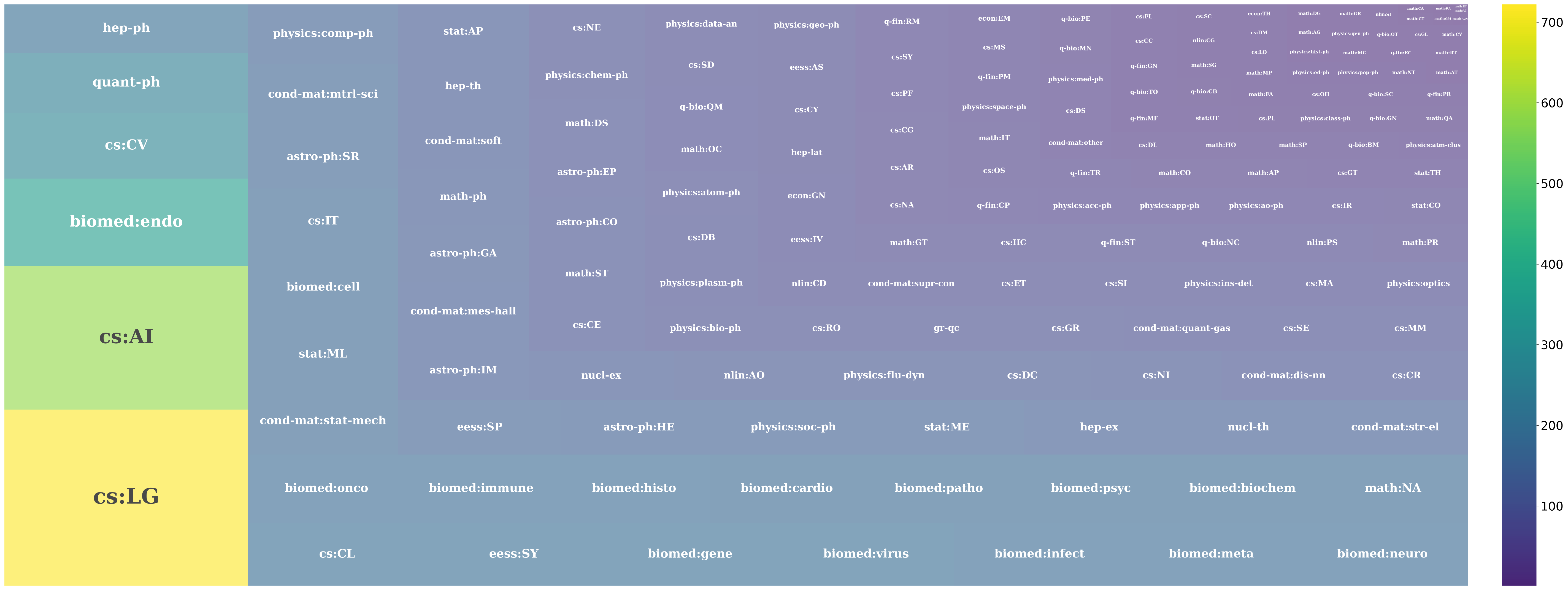}

\caption{\textbf{\ourbench Evaluation Domain Distribution.} Besides the domain distribution of \ourbench (Fig.~\ref{fig:anabench_domain_distribution} \& Fig.~\ref{fig:broad_domain_distribution}.a), we additionally visualize that of the downsampled evaluation set across 9 broad domains (Fig.~\ref{fig:broad_domain_distribution}.b) and 170 subdomains (Fig.~\ref{fig:test_domain_distribution}).}
\label{fig:test_domain_distribution}

\end{figure}

%% file: figures/anabench_statistics.tex
\begin{figure}[H]
\centering
\includegraphics[width=0.9\linewidth]{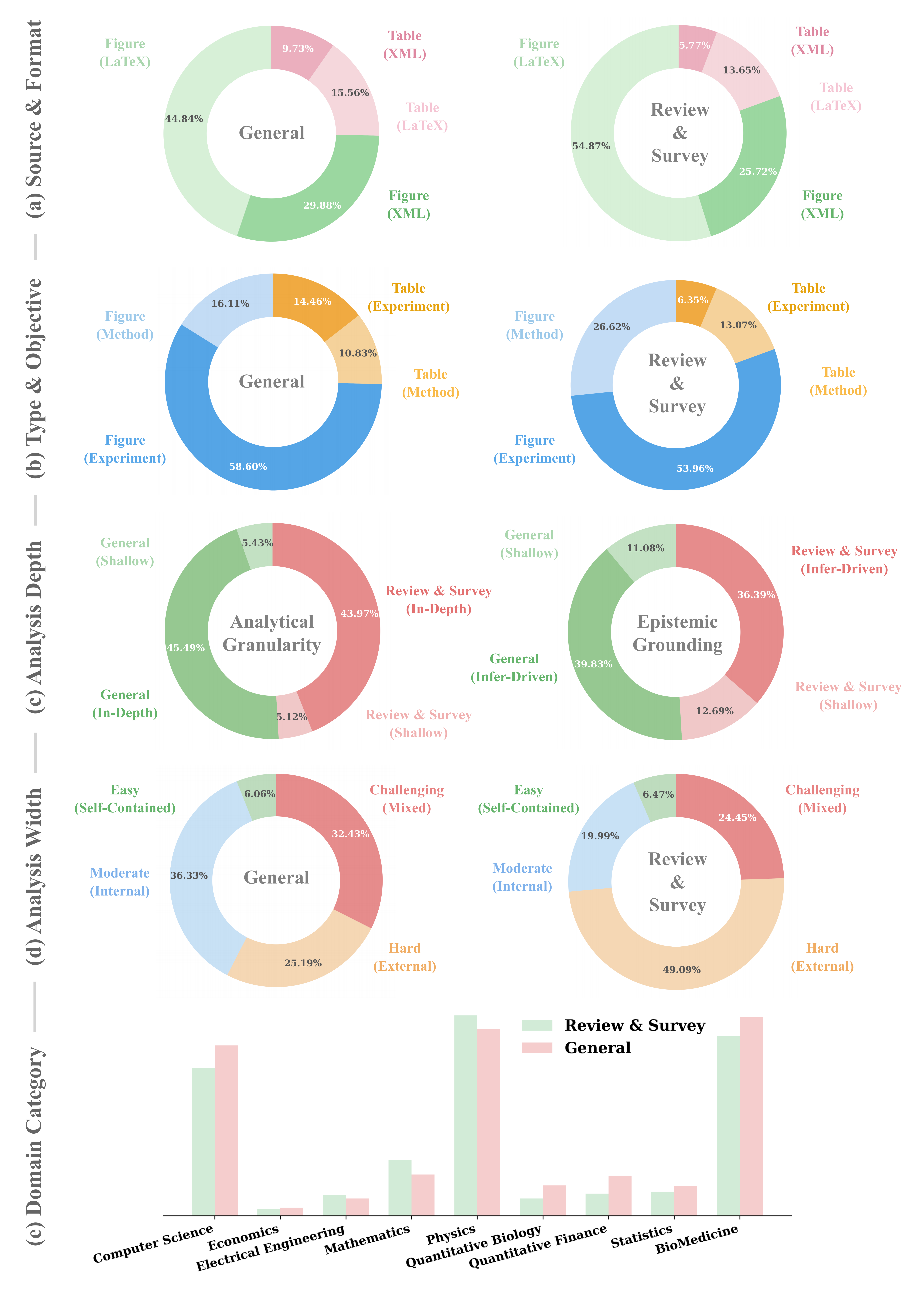}

\caption{\textbf{\ourbench: Benchmark Statistics.} We quantitatively analyze our benchmark in the dimensions of seven challenges (Fig.~\ref{fig:preliminary}) across varying data and analysis complexities.}
\label{fig:anabench_statistics}
\end{figure}

%% file: tables/data_domain.tex

\begin{table}[H]

\caption{\textbf{\ourbench Domains.} This table summarizes the 9 broad domains and 170 subdomains that \ourbench covers and collects data from. Directly summing subdomain categories is inaccurate due to subdomain overlapping.}
\label{tab:data_domain}

\centering
\setlength{\tabcolsep}{12pt} 
\renewcommand{\arraystretch}{1.5} 
\begin{tabularx}{\textwidth}{>{\centering\arraybackslash}m{0.2\textwidth} >{\raggedright\arraybackslash}m{0.7\textwidth}}
\toprule
\textbf{Broad Domain} & \multicolumn{1}{c}{\textbf{Fine-Grained Subdomain}} \\
\midrule

\makecell{\textbf{Computer Science} \\ (40 Subdomains)} & \textit{Artificial Intelligence; Hardware Architecture; Computational Complexity; Computational Engineering, Finance, and Science; Computational Geometry; Computation and Language; Cryptography and Security; Computer Vision and Pattern Recognition; Computers and Society; Databases; Distributed, Parallel, and Cluster Computing; Digital Libraries; Discrete Mathematics; Data Structures and Algorithms; Emerging Technologies; Formal Languages and Automata Theory; General Literature; Graphics; Computer Science and Game Theory; Human-Computer Interaction; Information Retrieval; Information Theory; Machine Learning; Logic in Computer Science; Multiagent Systems; Multimedia; Mathematical Software; Numerical Analysis; Neural and Evolutionary Computing; Networking and Internet Architecture; Other Computer Science; Operating Systems; Performance; Programming Languages; Robotics; Symbolic Computation; Sound; Software Engineering; Social and Information Networks; Systems and Control} \\

\makecell{\textbf{Economics} \\ (3 Subdomains)} & \textit{Econometrics; General Economics; Theoretical Economics} \\

\makecell{\textbf{Electrical Engineering} \\ (4 Subdomains)} & \textit{Audio and Speech Processing; Image and Video Processing; Signal Processing; Systems and Control} \\

\makecell{\textbf{Physics} \\ (51 Subdomains)} & \textit{Astrophysics (Cosmology and Nongalactic Astrophysics; Earth and Planetary Astrophysics; Astrophysics of Galaxies; High Energy Astrophysical Phenomena; Instrumentation and Methods for Astrophysics; Solar and Stellar Astrophysics); Condensed Matter (Disordered Systems and Neural Networks; Mesoscale and Nanoscale Physics; Materials Science; Other Condensed Matter; Quantum Gases; Soft Condensed Matter; Statistical Mechanics; Strongly Correlated Electrons; Superconductivity); General Relativity and Quantum Cosmology (General Relativity and Quantum Cosmology); High Energy Physics - Experiment; High Energy Physics - Lattice; High Energy Physics - Phenomenology; High Energy Physics - Theory; Mathematical Physics; Nonlinear Sciences; Nuclear Experiment; Nuclear Theory; Physics (Accelerator Physics; Atmospheric and Oceanic Physics; Applied Physics; Biological Physics; Chemical Physics; Classical Physics; Computational Physics; Data Analysis, Statistics and Probability; Physics Education; Fluid Dynamics; General Physics; Geophysics; History and Philosophy of Physics; Instrumentation and Detectors; Medical Physics; Optics; Plasma Physics; Popular Physics; Physics and Society; Space Physics); Quantum Physics} \\

\bottomrule
\end{tabularx}

\end{table}

\begin{table}[H]


\centering
\setlength{\tabcolsep}{12pt} 
\renewcommand{\arraystretch}{1.5} 
\begin{tabularx}{\textwidth}{>{\centering\arraybackslash}m{0.2\textwidth} >{\raggedright\arraybackslash}m{0.7\textwidth}}
\toprule
\textbf{Broad Domain} & \multicolumn{1}{c}{\textbf{Fine-Grained Subdomain}} \\
\midrule

\makecell{\textbf{Mathematics} \\ (32 Subdomains)} & \textit{Commutative Algebra; Algebraic Geometry; Analysis of PDEs; Algebraic Topology; Classical Analysis and ODEs; Combinatorics; Category Theory; Complex Variables; Differential Geometry; Dynamical Systems; Functional Analysis; General Mathematics; General Topology; Group Theory; Geometric Topology; History and Overview; Information Theory; K-Theory and Homology; Logic; Metric Geometry; Mathematical Physics; Numerical Analysis; Number Theory; Operator Algebras; Optimization and Control; Probability; Quantum Algebra; Rings and Algebras; Representation Theory; Symplectic Geometry; Spectral Theory; Statistics Theory} \\

\makecell{\textbf{Quantitative Biology} \\ (10 Subdomains)} & \textit{Biomolecules; Cell Behavior; Genomics; Molecular Networks; Neurons and Cognition; Other Quantitative Biology; Populations and Evolution; Quantitative Methods; Subcellular Processes; Tissues and Organs} \\

\makecell{\textbf{Quantitative Finance} \\ (9 Subdomains)} & \textit{Computational Finance; Economics; General Finance; Mathematical Finance; Portfolio Management; Pricing of Securities; Risk Management; Statistical Finance; Trading and Market Microstructure} \\

\makecell{\textbf{Statistics} \\ (6 Subdomains)} & \textit{Applications; Computation; Methodology; Machine Learning; Other Statistics; Statistics Theory} \\

\makecell{\textbf{Biomedicine} \\ (15 Subdomains)} & \textit{General Pathology; Infectious Disease; Neurological Disease; Endocrine \& Metabolic Disease; Psychiatry; Oncology; Cardiovascular System; Cell Biology; Genetics; Endocrinology; Immunology; Biochemistry; Metabolism; Histology; Virology} \\

\bottomrule
\end{tabularx}


\end{table}

%% file: tables/data_size.tex
\definecolor{DATA_BG}{HTML}{F6DAC0}

\begin{table}[H]

\caption{\textbf{SFT Data Distribution.} This table summarizes the data size, data type, data format, and domain categories of different downsampled datasets for SFT.}
\label{tab:sft_data_size}

\centering

\renewcommand{\arraystretch}{1.3}

\begin{tabular*}{\textwidth}{@{\extracolsep{\fill}} c cc cc cc c}
\toprule
\multirow{2}{*}{\textbf{Data Size}} &
\multicolumn{2}{c}{\textbf{Data Type}} &
\multicolumn{2}{c}{\textbf{Data Format}} &
\multicolumn{2}{c}{\textbf{Data Source}} &
\multirow{2}{*}{\textbf{Domain}} \\
\cmidrule(lr){2-3} \cmidrule(lr){4-5} \cmidrule(lr){6-7}
& Figure & Table & LaTeX & XML & General & Review \& Survey & \\

\midrule

\rowcolor{DATA_BG}
\multicolumn{8}{c}{\textbf{Single-Format}} \\

\addlinespace[3pt]

20,000 & 16,743 & 3,258 & 20,000 & 0 & 10,000 & 10,000 & 8 \\
24,210 & 11,778 & 12,432 & 0 & 24,210 & 22,860 & 1,350 & 1 \\
42,804 & 33,485 & 9,319 & 42,804 & 0 & 20,000 & 11,350 & 8 \\

\addlinespace[3pt]

\rowcolor{DATA_BG}
\multicolumn{8}{c}{\textbf{Multi-Format}} \\

\addlinespace[3pt]

31,350 & 26,245 & 5,105 & 20,000 & 11,350 & 21,901 & 20,903 & 9 \\
67,014 & 45,263 & 21,751 & 42,804 & 11,350 & 44,761 & 22,253 & 9 \\

\bottomrule

\end{tabular*}

\end{table}

\begin{table}[H]

\caption{\textbf{RL Data Distribution.} This table summarizes the data size, data type, data format, and domain categories of the downsampled datasests for agent-wise RL training.
}
\label{tab:rl_data_size}

\centering

\renewcommand{\arraystretch}{1.3}

\begin{tabular*}{\textwidth}{@{\extracolsep{\fill}} c cc cc cc c}

\toprule

\multirow{2}{*}{\textbf{Agent}} &
\multicolumn{2}{c}{\textbf{Data Type}} &
\multicolumn{2}{c}{\textbf{Data Format}} &
\multicolumn{2}{c}{\textbf{Data Source}} &
\multirow{2}{*}{\textbf{Domain}} \\
\cmidrule(lr){2-3} \cmidrule(lr){4-5} \cmidrule(lr){6-7}
& Figure & Table & LaTeX & XML & General & Review \& Survey & \\

\midrule

\rowcolor{DATA_BG}
\multicolumn{8}{c}{\textbf{Train}} \\

\addlinespace[3pt]

\textbf{Planner} & 7,870 & 2,584 & 7,632 & 2,822 & 5,829 & 4,625 & 9 \\
\textbf{Expert} & 10,628 & 2,894 & 8,058 & 5,464 & 7,883 & 5,639 & 9 \\
\textbf{Solver} & 18,737 & 6,444 & 16,331 & 8,850 & 16,868 & 8,313 & 9 \\
\textbf{Critic} & 8,761 & 4,065 & 7,655 & 5,171 & 7,650 & 5,176 & 9 \\

\addlinespace[3pt]

\rowcolor{DATA_BG}
\multicolumn{8}{c}{\textbf{Test}} \\

\addlinespace[3pt]

\textbf{Planner} & 1,168 & 416 & 1,000 & 584 & 1,000 & 584 & 9 \\
\textbf{Expert} & 1,594 & 499 & 1,342 & 751 & 1,312 & 781 & 9 \\
\textbf{Solver} & 1,992 & 749 & 1,773 & 968 & 1,838 & 903 & 9 \\
\textbf{Critic} & 1,242 & 431 & 1,089 & 584 & 1,034 & 639 & 9 \\
\bottomrule

\end{tabular*}

\vspace{12pt}

\end{table}

%% file: figures/anabench_distribution_test.tex

\begin{wrapfigure}{r}{0.48\linewidth}
\centering
\includegraphics[width=\linewidth]{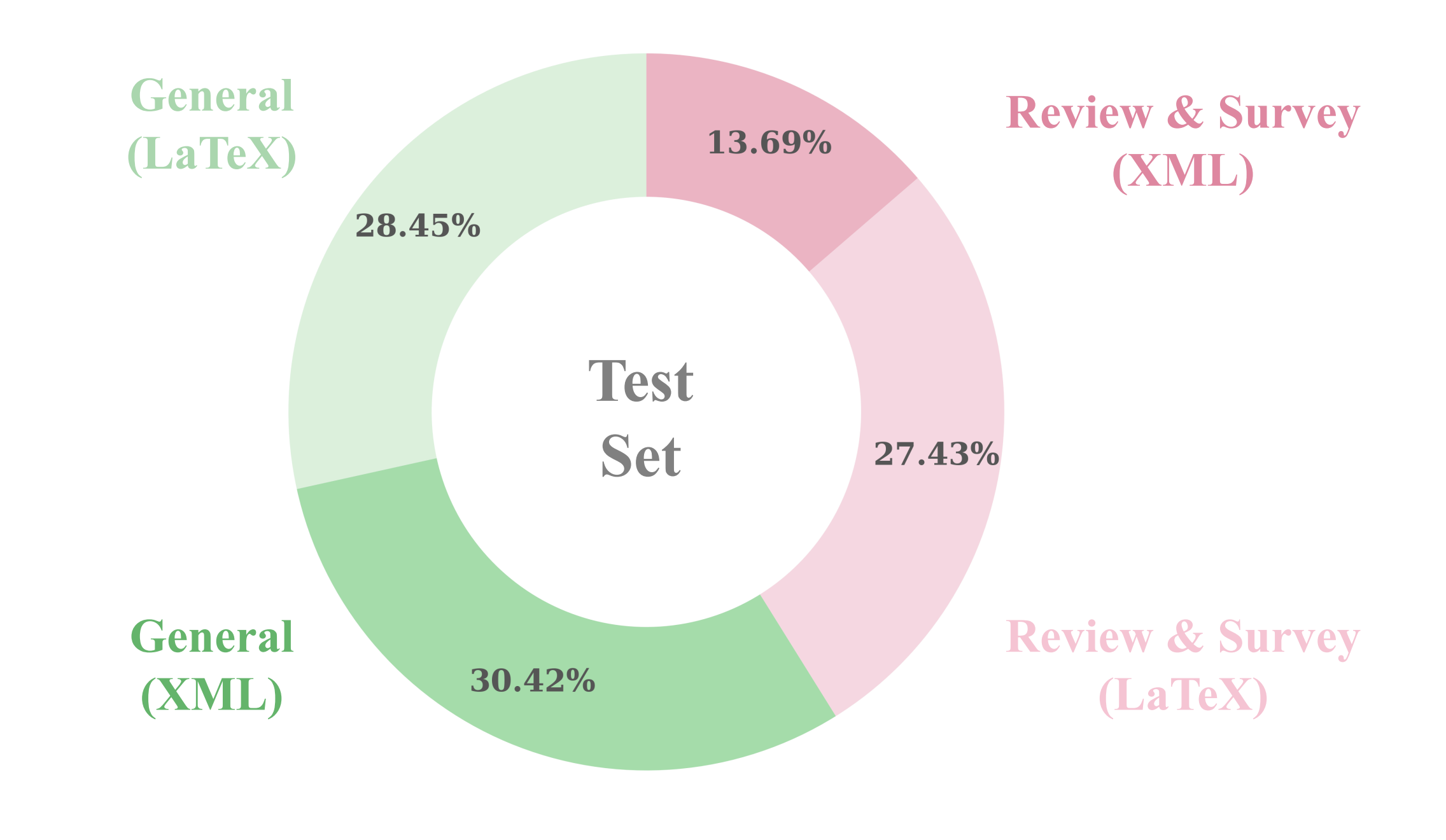}


\caption{\textbf{Evaluation Set Data Distribution.}}
\label{fig:test_data_distribution}
\end{wrapfigure}

%% file: figures/exp_ablation_agent.tex





\begin{wrapfigure}{r}{0.68\linewidth}

\vspace{-12pt}

\centering
\includegraphics[width=\linewidth]{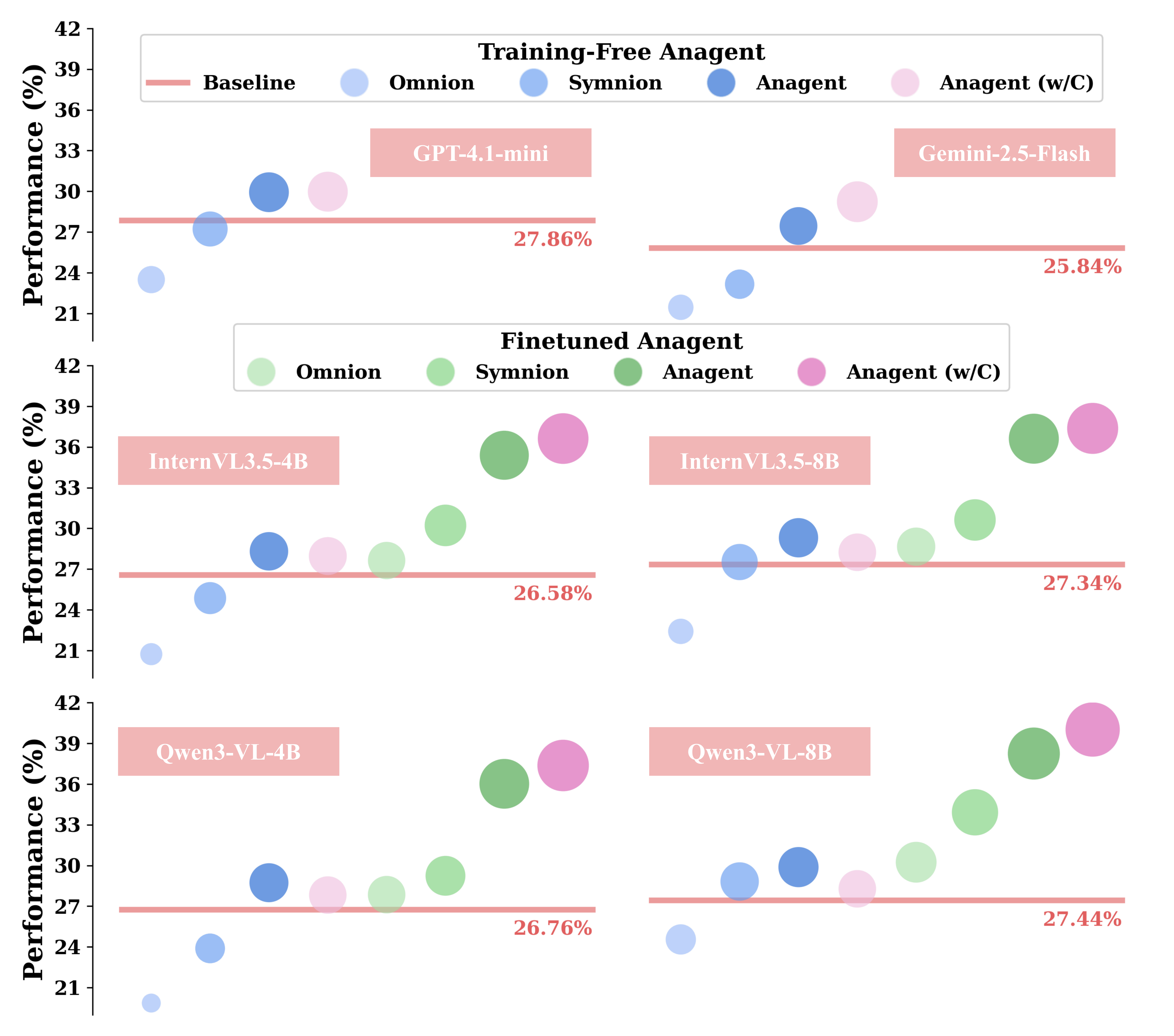}

\vspace{-3pt}

\caption{\textbf{Ablation Studies On \ours Variants.}}
\label{fig:exp:ablation_agent}

\vspace{-12pt}
\end{wrapfigure}

%% file: tables/anagent_variants.tex
\begin{table}[H]
\small
\centering
\caption{\textbf{\ours Variants.} Overview of \ours variants.}
\label{tab:anagent:variants}

\renewcommand{\arraystretch}{1.1}

\begin{tabular}{@{}>{\centering\arraybackslash}m{0.15\linewidth}>{\centering\arraybackslash}m{0.15\linewidth}>{\centering\arraybackslash}m{0.10\linewidth}>{\centering\arraybackslash}m{0.10\linewidth}>{\centering\arraybackslash}m{0.10\linewidth}>{\centering\arraybackslash}m{0.10\linewidth}>{\centering\arraybackslash}m{0.10\linewidth}@{}}

\toprule

\multicolumn{2}{c}{\multirow{2}{*}{\textbf{\ours Variants}}} & \multirow{2}{*}{\textbf{Tools}} & \multicolumn{4}{c}{\textbf{Agent Component}} \\

\cmidrule(lr){4-7}

\multicolumn{2}{c}{} & & \textbf{\Planner} & \textbf{\Expert} & \textbf{\Solver} & \textbf{\Critic} \\

\midrule

\textbf{Baselines} & \textbf{--} & \xmark & \xmark & \xmark & \xmark & \xmark \\

\textbf{Omnion} & \textbf{--} & \cmark & \xmark & \xmark & \cmark & \xmark \\

\textbf{Symnion} & \textbf{--} & \cmark & \xmark & \cmark & \cmark & \xmark \\

\midrule

\multirow{2}{*}{\textbf{\ours}} & \textbf{--} & \cmark & \cmark & \cmark & \cmark & \xmark \\

& \textbf{w/ \Critic} & \cmark & \cmark & \cmark & \cmark & \cmark \\

\bottomrule

\end{tabular}
\end{table}

%% file: tables/exp_config.tex
\begin{table}[H]
\renewcommand{\arraystretch}{1.3}
\small
\centering
\caption{\textbf{Computation Overhead.} This table summarizes the computation overhead during training and evaluation. $n_{MLLM}$ denotes the number of different backbone MLLMs.}
\label{tab:exp:computation_overhead}

\begin{tabular*}{\textwidth}{@{\extracolsep{\fill}}ccccc@{}}
\toprule
 & \multicolumn{4}{c}{\textbf{$\bm{n_{MLLM}}$}} \\
 
\cmidrule(lr){2-5}

 & $\bm{1}$ & $\bm{2}$ & $\bm{3}$ & $\bm{4}$ \\

\midrule

\textbf{Evaluation} & $A100$ $80GB \times 1$ & $A100$ $80GB \times 2$ & $A100$ $80GB \times 4$ & $A100$ $80GB \times 4$ \\
\textbf{SFT} & $A100$ $80GB \times 1$ & -- & -- & -- \\
\textbf{RL} & $A100$ $80GB \times 8$ & -- & -- & -- \\

\bottomrule

\end{tabular*}
\end{table}

\begin{table}[H]
\renewcommand{\arraystretch}{1.3}
\small
\centering
\caption{\textbf{Implementation Details.} This table summarizes the key configuration settings during training and evaluation. $M_p$, $M_e$, $M_s$, and $M_c$ denote the maximum number of turns allowed for Planner, Expert, Solver, and Critic, respectively. The maximum depth of context searching tools is set to $k=1$ across all settings. The maximum and minimum pixel values are denoted by $max_{\textit{pixel}}$ and $min_{\textit{pixel}}$. The learning rate is represented by $lr$. $n_{\textit{rollout}}$ denotes the number of rollouts during RL, and $n_{\textit{epoch}}$ denotes the number of training epochs.}
\label{tab:exp:configuration}

\resizebox{\textwidth}{!}{%
\begin{tabular}{ccccccccccc}

\toprule

 & \multicolumn{4}{c}{\textbf{Agent}} & \multicolumn{6}{c}{\textbf{Configuration}} \\
 
\cmidrule(lr){2-5} \cmidrule(lr){6-11}
 & \Planner & \Expert & \Solver & \Critic & \multicolumn{6}{c}{(All Agents)} \\
 
 & $\bm{M_p}$ & $\bm{M_e}$ & $\bm{M_s}$ & $\bm{M_c}$ & $\bm{k}$ & $\bm{max_{\textit{pixel}}}$ & $\bm{min_{\textit{pixel}}}$ & $\bm{lr}$ & $\bm{n_{\textit{rollout}}}$ & $\bm{n_{\textit{epoch}}}$ \\
 
\midrule

\textbf{Evaluation} & 1 & 5 & 2 & 1 & 1 & $1024 \times 1024$ & $128 \times 128$ & -- & -- & -- \\
\textbf{SFT} & 1 & 1 & 1 & 1 & 1 & $512 \times 512$ & $128 \times 128$ & $1\times10^{-4}$ & -- & $1$ \\
\textbf{RL} & 1 & 1 & 1 & 1 & 1 & $512 \times 512$ & $128 \times 128$ & $1\times10^{-6}$ & 4 & $1$ \\

\bottomrule

\end{tabular}%
}
\end{table}

%% file: tables/exp_main3_mllm_as_judge.tex
\begin{table}[H]

\caption{\textbf{MLLM-As-Judge Evaluation.} Evaluation of scientific table \& figure analysis across baselines, training-free \ours, and finetuned \ours, through five-dimensional evaluation protocol (\S\ref{appendix:evaluation:five_dimension}). Compared with baselines, \textbf{\textit{relative performance differences}} (Eq.~\ref{eq:relative_delta}) are shown as \textit{positive} \textcolor{deltagreen}{$\uparrow \Delta_{\textit{rel}}\%$} or \textit{negative} \textcolor{deltared}{$\downarrow \Delta_{\textit{rel}}\%$}.}
\label{tab:experiment:mllm_as_judge_main_results}

\small
\centering
\renewcommand{\arraystretch}{1.1}  
\setlength{\tabcolsep}{4pt}  

\begin{tabular*}{\textwidth}{@{\extracolsep{\fill}}l c ccccc c@{}}
\toprule
\multirow{2}{*}{\textbf{Model}} & \multirow{2}{*}{\textbf{Size}} & \multicolumn{5}{c}{\textbf{Five-Dimensional Evaluation (\%)}} & \multicolumn{1}{c}{\textbf{Overall Accuracy (\%)}} \\
\cmidrule(lr){3-7} \cmidrule(lr){8-8}
& & $\bm{S}_{\textbf{\textsc{Acc}}}$ & $\bm{S}_{\textbf{\textsc{Complete}}}$ & $\bm{S}_{\textbf{\textsc{Format}}}$ & $\bm{S}_{\textbf{\textsc{Clarity}}}$ & $\bm{S}_{\textbf{\textsc{Faith}}}$ & $\bm{S}_{\textbf{\textsc{Mllm}}}$ \\

\midrule

\multicolumn{8}{c}{\cellcolor{baselinebg}\textbf{Baselines}} \\
\addlinespace[3pt]

\multirow{2}{*}{\textbf{InternVL-3.5}} & \textbf{4B} & 42.00 & 27.17 & 69.67 & 64.33 & 55.33 & 51.70 \\
& \textbf{8B} & \underline{52.67} & 30.50 & 71.67 & \underline{68.50} & \underline{60.83} & 56.83 \\
\multirow{2}{*}{\textbf{Qwen2.5-VL}} & \textbf{3B} & 43.17 & 24.88 & 64.50 & 64.83 & 56.00 & 50.68 \\
& \textbf{7B} & 50.83 & \textbf{32.17} & \underline{73.50} & 67.50 & 60.33 & \underline{56.87} \\
\multirow{2}{*}{\textbf{Qwen3-VL}} & \textbf{4B} & 46.50 & 28.50 & 68.33 & 67.17 & 57.00 & 53.50 \\
& \textbf{8B} & \textbf{52.83} & \underline{31.17} & \textbf{74.17} & \textbf{69.50} & \textbf{61.17} & \textbf{57.77} \\


\multicolumn{8}{c}{\cellcolor{oursbaselinebg}\textbf{\ours (Training-Free)}} \\
\addlinespace[3pt]

\multirow{2}{*}{\textbf{InternVL-3.5}} & \textbf{4B} & 44.17 & 30.83 & 70.33 & 66.00 & 59.50 & \reldeltapct{51.70}{54.17} \\

& \textbf{8B} & \textbf{55.67} & 35.17 & 71.83 & 71.67 & \textbf{65.50} & \reldeltapctul{56.83}{59.97} \\

\multirow{2}{*}{\textbf{Qwen2.5-VL}} & \textbf{3B} & 47.67 & 25.13 & 67.83 & 66.17 & 56.50 & \reldeltapct{50.68}{52.66} \\

& \textbf{7B} & 52.17 & \underline{37.50} & \underline{74.17} & \underline{69.83} & 63.00 & \reldeltapct{56.87}{59.33} \\

\multirow{2}{*}{\textbf{Qwen3-VL}} & \textbf{4B} & 51.67 & 31.83 & 68.50 & 67.50 & 60.83 & \reldeltapct{53.50}{56.07} \\

& \textbf{8B} & \underline{54.17} & \textbf{39.50} & \textbf{77.50} & \textbf{72.83} & \underline{64.50} & \reldeltapctbf{57.77}{61.70} \\


\multicolumn{8}{c}{\cellcolor{ourstrainedbg}\textbf{\ours (SFT)}} \\
\addlinespace[3pt]

\multirow{2}{*}{\textbf{InternVL-3.5}} & \textbf{4B} & 50.83 & 47.50 & 77.83 & 69.33 & 64.67 & \reldeltapct{51.70}{62.03} \\

& \textbf{8B} & \underline{56.33} & 52.33 & 79.00 & \textbf{75.17} & 72.17 & \reldeltapct{56.83}{67.00} \\

\multirow{2}{*}{\textbf{Qwen2.5-VL}} & \textbf{3B} & 50.33 & 48.33 & 76.50 & 71.17 & 65.67 & \reldeltapct{50.68}{62.40} \\

& \textbf{7B} & 54.33 & \underline{53.67} & \underline{80.83} & 73.00 & \textbf{74.50} & \reldeltapctul{56.87}{67.27} \\

\multirow{2}{*}{\textbf{Qwen3-VL}} & \textbf{4B} & 53.50 & 50.17 & 76.17 & 72.17 & 65.33 & \reldeltapct{53.50}{63.47} \\

& \textbf{8B} & \textbf{57.17} & \textbf{54.83} & \textbf{83.83} & \underline{74.67} & \underline{73.83} & \reldeltapctbf{57.77}{68.87} \\


\multicolumn{8}{c}{\cellcolor{ourstrainedbg}\textbf{\ours (SFT + RL)}} \\
\addlinespace[3pt]

\textbf{Qwen2.5-VL} & \textbf{3B} & 54.00 & 55.17 & 78.83 & 72.17 & 67.33 & \reldeltapct{50.68}{65.50} \\

\textbf{Qwen3-VL} & \textbf{4B} & \textbf{56.50} & \textbf{57.17} & \textbf{79.33} & \textbf{73.33} & \textbf{68.50} & \reldeltapctbf{53.50}{66.97} \\

\bottomrule
\end{tabular*}

\end{table}

%% file: figures/length.tex

\begin{figure}[H]
\centering
\includegraphics[width=1.0\linewidth]{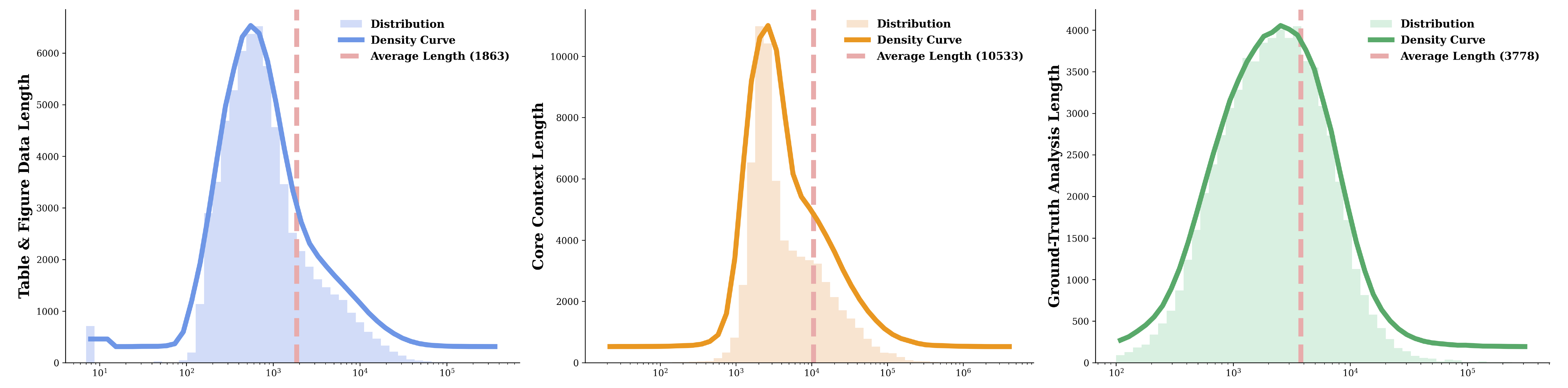}

\caption{\textbf{Length Distribution.} Visualization of Data Length in \ourbench. (1) Left: Input data length. (2) Center: Core context length. (3) Right: Ground-truth analysis length.}
\label{fig:length_distribution}

\end{figure}

\begin{figure}[H]
\centering
\includegraphics[width=1.0\linewidth]{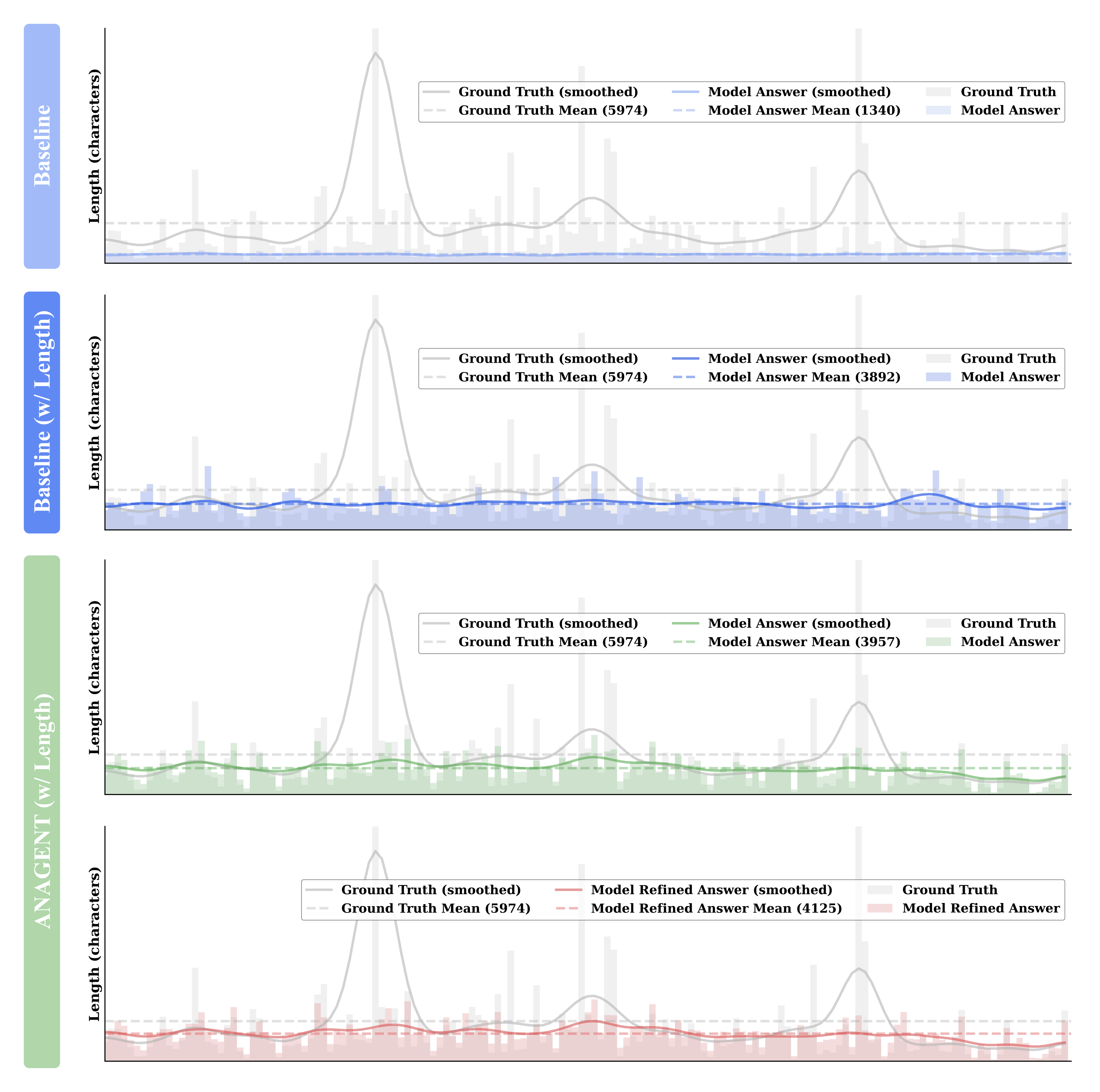}

\caption{\textbf{Length Distribution of GPT-4.1-mini.} Using GPT-4.1-mini as the base model, this figure visualizes the length distribution of 150 randomly sampled agent-generated analysis versus ground-truth analysis length.}
\label{fig:length_distribution_gpt}

\vspace{28pt}

\end{figure}

\begin{figure}[H]

\vspace{-9pt}

\centering
\includegraphics[width=1.0\linewidth]{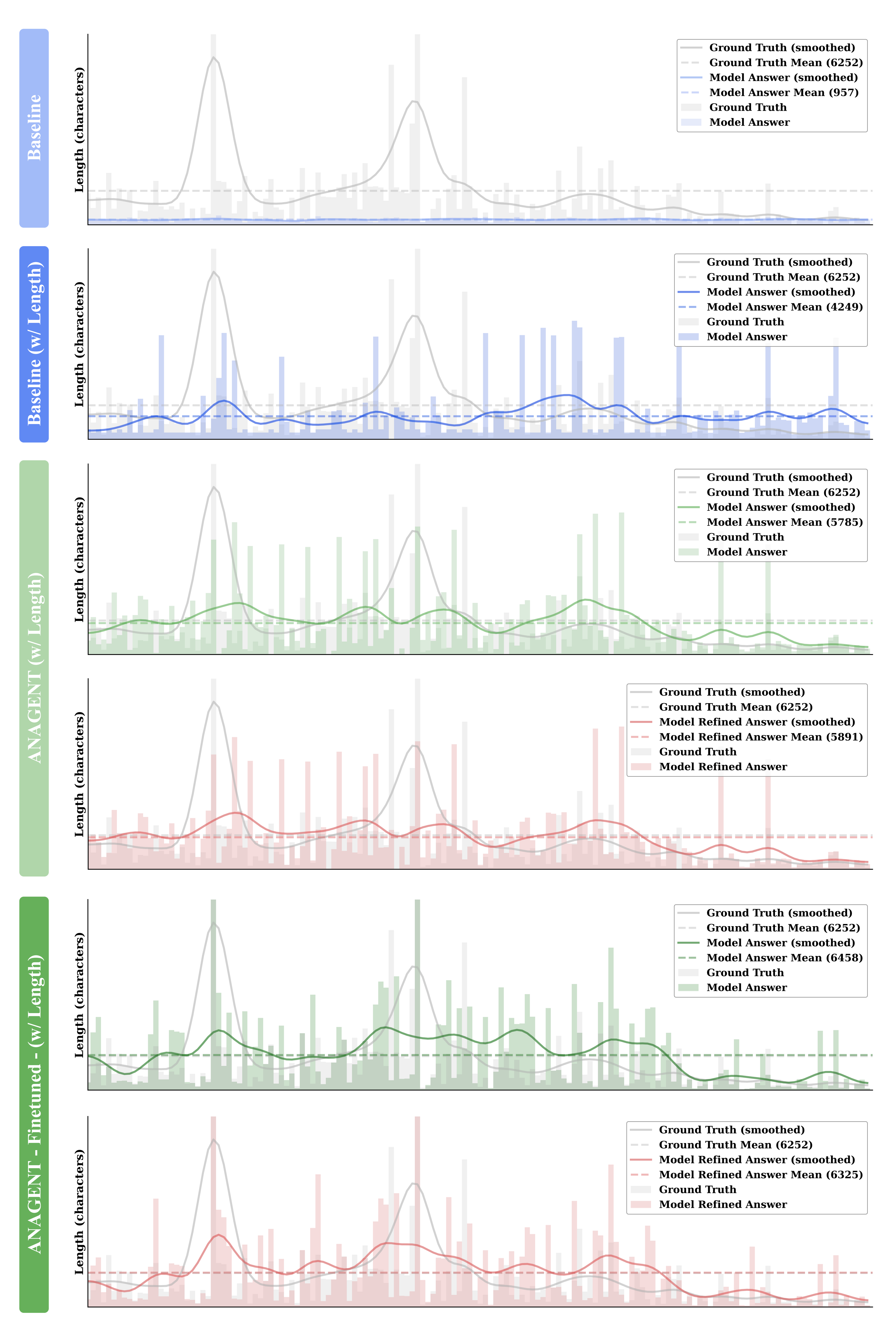}

\vspace{-10pt}

\caption{\textbf{Length Distribution of Qwen3-VL-8B.} Leveraging Qwen3-VL-8B, this figure visualizes the length distribution of 150 randomly sampled agent-generated analysis versus ground-truth analysis length.}
\label{fig:length_distribution_qwen}

\end{figure}

%% file: tables/method_toolkit.tex
\definecolor{DocumentToolkit}{HTML}{FFFDD9}
\definecolor{KnowledgeToolkit}{HTML}{FFEBD6}
\definecolor{SearchToolkit}{HTML}{E8FEE1}
\definecolor{VisionToolkit}{HTML}{DEF0FF}
\definecolor{SandboxToolkit}{HTML}{FFECFB}

\renewcommand{\tabularxcolumn}[1]{m{#1}}

\begin{table}[H]

\caption{\textbf{Five Specialized Toolkits of \ours.}
To enhance specialized scientific research capabilities, \ours is equipped with five complementary toolkits that respectively support document comprehension, knowledge retrieval, targeted searching, multimodal understanding, and autonomous coding exploration.}
\label{tab:five_toolkits}

\small
\centering
\renewcommand{\arraystretch}{1.45}

\begin{tabularx}{\linewidth}{
    >{\centering\arraybackslash}m{0.15\linewidth}
    >{\raggedright\arraybackslash}m{0.35\linewidth}
    >{\raggedright\arraybackslash}X
}
\toprule
\multicolumn{1}{c}{\textbf{Specialized Tools}} & \multicolumn{1}{c}{\textbf{Supported Inputs \& Formats}} & \multicolumn{1}{c}{\textbf{Functions}} \\

\midrule

\rowcolor{DocumentToolkit}
\multicolumn{3}{c}{\textbf{Document Toolkit}} \\

\cellcolor{DocumentToolkit}\textbf{Online Fetcher} & URL (arXiv, Semantic Scholar, PubMed, general web), Source Type \& Preferred Format (HTML, LaTeX, XML, PDF), DOI, Paper ID, Search Query & Fetch and parse documents from online sources, including arXiv papers (LaTeX/PDF), Semantic Scholar (search/metadata), PubMed Central (XML), and general URLs (HTML/PDF/XML) \\
\cellcolor{DocumentToolkit}\textbf{PDF Parser} & URL, Local Path, Bytes & Parse PDF documents to extract text, metadata, hierarchical contents with optional image saving \\
\cellcolor{DocumentToolkit}\textbf{XML Parser} & URL, Local Path, XML Strings, Bytes & Parse XML documents and extract metadata, hierarchical contents with configurable detail levels \\

\addlinespace[3pt]
\rowcolor{KnowledgeToolkit}
\multicolumn{3}{c}{\textbf{Knowledge Toolkit}} \\

\cellcolor{KnowledgeToolkit}\textbf{Abstract Collector} & Title, Author, Year, URL, DOI, arXiv ID, PMID, Local Path & Search and extract paper abstracts from multiple academic sources (e.g., arXiv, PubMed, Semantic Scholar, CrossRef) \\
\cellcolor{KnowledgeToolkit}\textbf{Information Localizer} & Search String (e.g., keyword, phrase, caption, title, equation), URL, Local Path & Localize and extract the complete section (broad meaning, e.g., section, subsection, figure, table, etc.) according to the input query \\
\cellcolor{KnowledgeToolkit}\textbf{Context Finder} & Search String (e.g., keyword, phrase, caption, title, equation), URL, Local Path & Localize a search string in a document and extract all bidirectional citation contexts with multi-level traversal support \\
\cellcolor{KnowledgeToolkit}\textbf{Section Extractor} & Section Identifier (e.g., keywords, numbers), URL & Extract the complete sections (broad meaning, e.g., section, subsection, figure, table, etc.) according to the input query \\

\addlinespace[3pt]
\rowcolor{SearchToolkit}
\multicolumn{3}{c}{\textbf{Search Toolkit}} \\

\cellcolor{SearchToolkit}\textbf{arXiv Searcher} & URL, Title, Keywords, Author, Category, Sort, Max Results, and Other arXiv Search Queries and Filters & Search and retrieve arXiv preprint scientific literature across diverse domains \\
\cellcolor{SearchToolkit}\textbf{PebMed Searcher} & URL, Title, Keywords, Author, Time Range, PMID, Sort, Time Range, Max Results, and Other PubMed Search Queries and Filters & Search and retrieve biomedical scientific literature in PubMed \\
\cellcolor{SearchToolkit}\textbf{Semantic Scholar Searcher} & URL, Title, Year, Field, Venue, Citations, Max Results, and Other Semantic Scholar Search Queries and Filters & Search and retrieve scientific literature via Semantic Scholar API \\
\cellcolor{SearchToolkit}\textbf{Web Searcher} & Search String, Keywords, Temporal Filters, Max Results, Date, Language, Search Level, and Other Web Search Queries and Filters & Search and retrieve web information via Google Search API \\
\cellcolor{SearchToolkit}\textbf{Wikipedia Searcher} & Search String, Title, Keywords, Search Mode, and Other Wikipedia Search Queries and Filters & Search and retrieve Wikipedia concepts, terminologies, and articles with intelligent routing across five MediaWiki APIs \\

\addlinespace[3pt]
\rowcolor{VisionToolkit}
\multicolumn{3}{c}{\textbf{Vision Toolkit}} \\

\cellcolor{VisionToolkit}\textbf{OCR Extractor} & Image Source, Language, Bounding Boxes, Threshold & Enhance visual perception through OCR \\
\cellcolor{VisionToolkit}\textbf{Figure Parser} & Image Source, Query, Contexts & Parse and extract visual information from scientific figures \\
\cellcolor{VisionToolkit}\textbf{Image Analyzer} & Image Source, Query, Focus, Detail Level, Contexts & Analyze images with interactive query-driven exploration \\

\addlinespace[3pt]
\rowcolor{SandboxToolkit}
\multicolumn{3}{c}{\textbf{Sandbox Toolkit}} \\

\cellcolor{SandboxToolkit}\textbf{Sandbox Explorer} & Python Code, Environment Dependencies & Execute Python code in isolated Docker/Apptainer sandbox with package installation and task data access for autonomous tool development and task exploration \\

\bottomrule
\end{tabularx}

\end{table}

%% file: figures/exp_ablation_context.tex
\begin{figure}[H]
\centering
\includegraphics[width=1.0\linewidth]{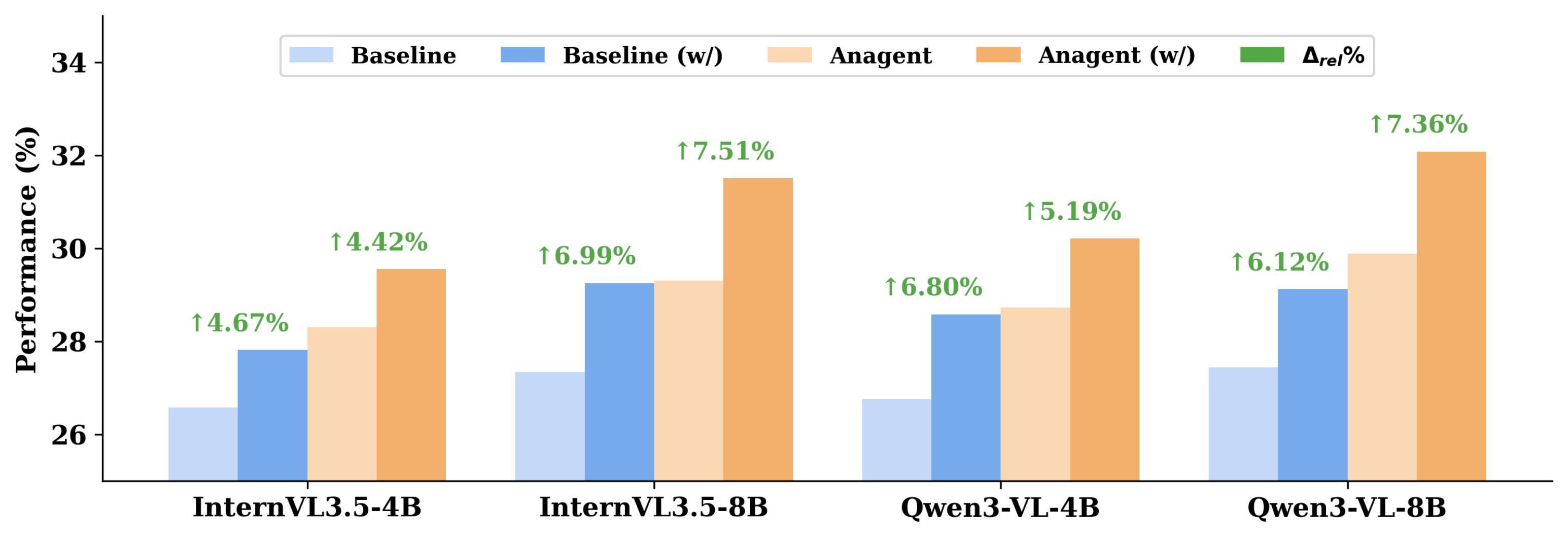}

\caption{\textbf{Ablation Studies On Additional Contexts \& Domain Knowledge.} Comparison between with ("w/" in the figure) and without \textit{gold contextual information and domain-specific knowledge}.}
\label{fig:exp:ablation_context}
\end{figure}

%% file: figures/exp_requirement_prompting.tex
\begin{figure*}[h]
\centering
\includegraphics[width=\linewidth]{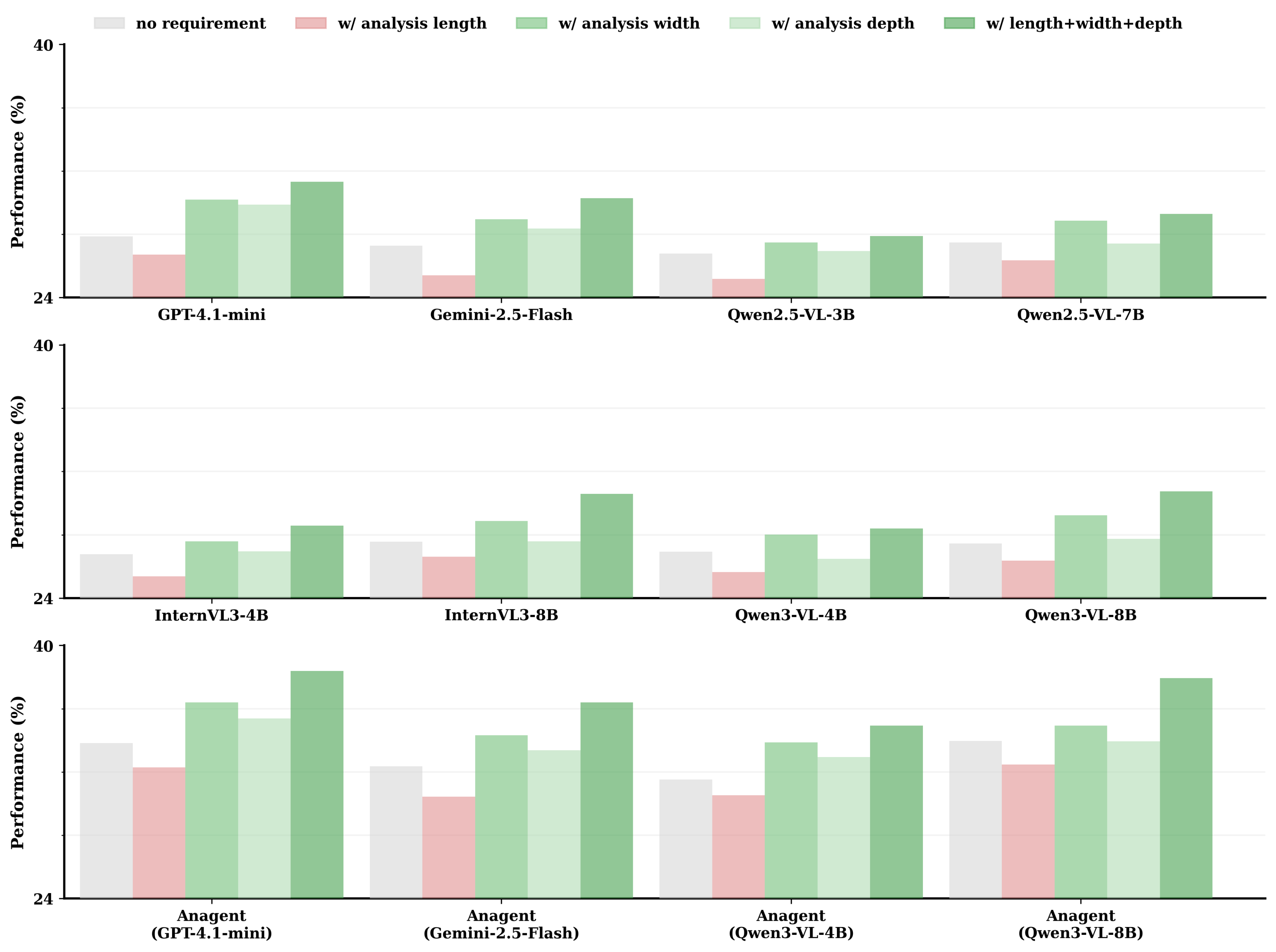}

\caption{\textbf{When AI Scientists Meet Stated Analysis Expectations.} Our experiments investigate AI agents' capabilities in properly comprehending and implementing scientific analysis with explicit expectations.}
\label{fig:abation:requirement_prompting}
\end{figure*}

%% file: figures/exp_ablation_complexity_data_type.tex
\begin{figure}[H]
\centering
\includegraphics[width=1.0\linewidth]{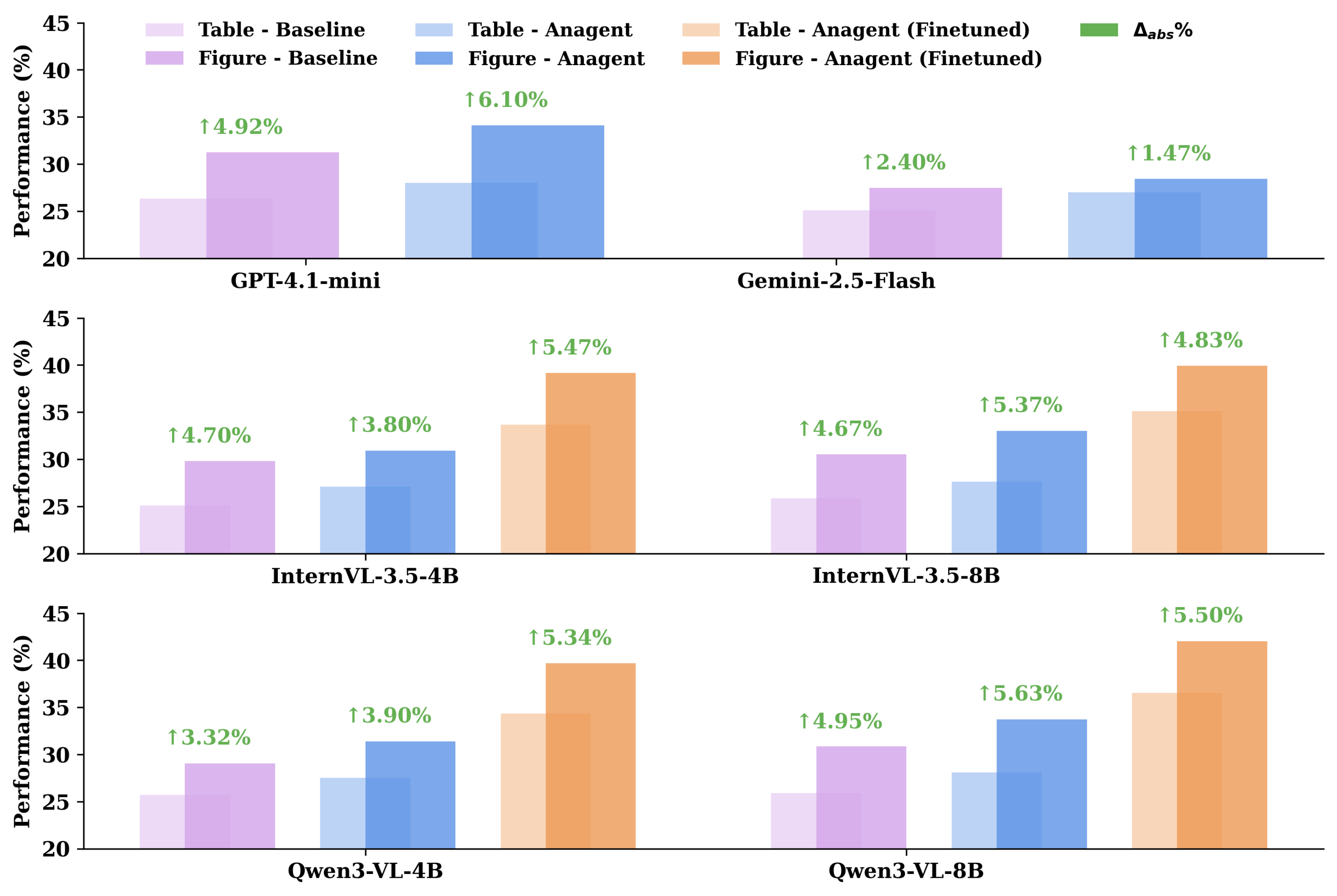}

\caption{\textbf{Ablation Study on Data Type.} Evaluation of agent performance across different data types (\S\ref{appendix:benchmark:curriculum:data_complexity}).}
\label{fig:exp:ablation_complexity_data:type}
\end{figure}

%% file: figures/exp_ablation_complexity_data_format.tex
\begin{figure}[H]
\centering
\includegraphics[width=1.0\linewidth]{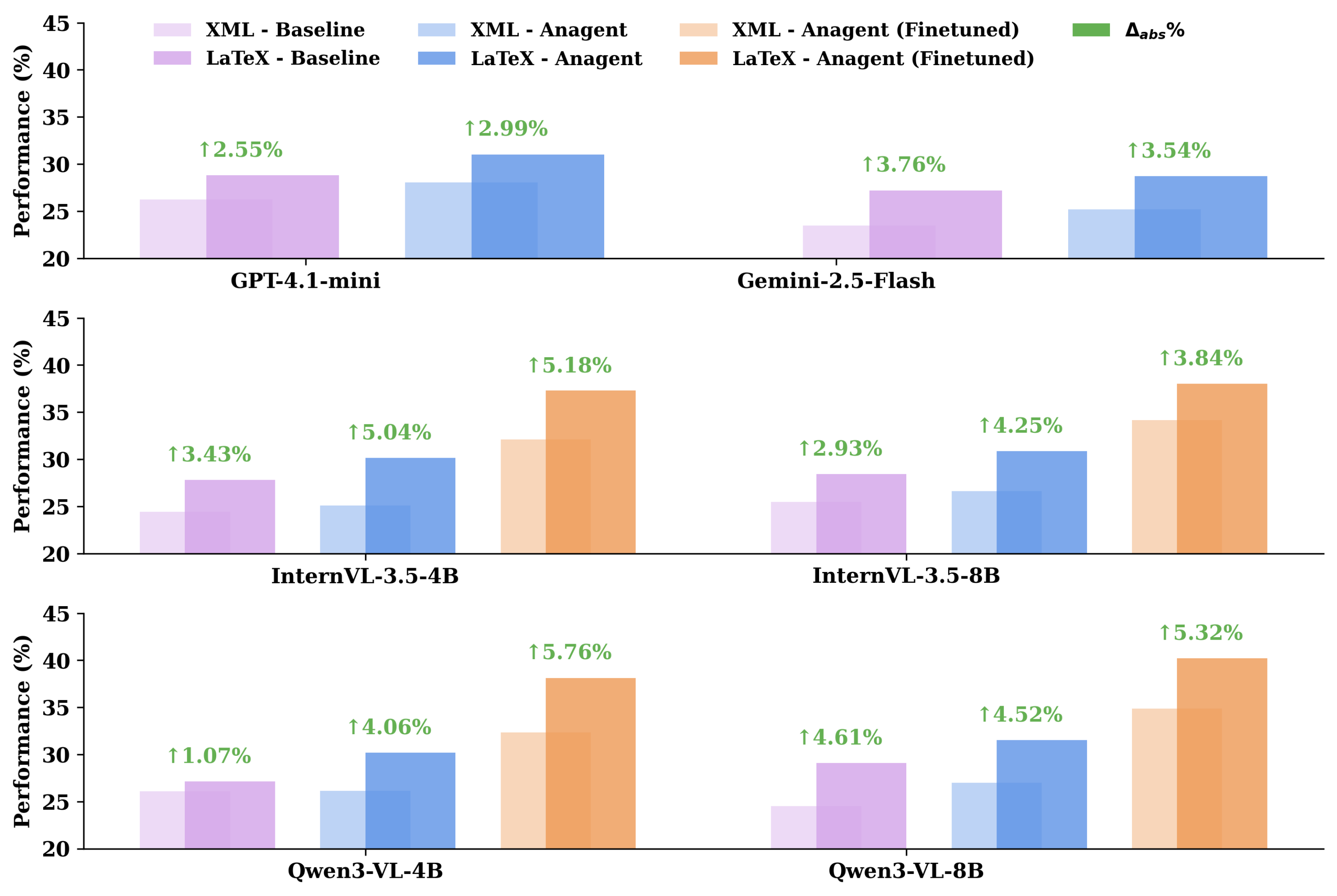}

\caption{\textbf{Ablation Study on Data Format.} Evaluation of agent performance across different data formats (\S\ref{appendix:benchmark:curriculum:data_complexity}).}
\label{fig:exp:ablation_complexity_data:format}
\end{figure}

%% file: figures/exp_ablation_complexity_data_source.tex
\begin{figure}[H]
\centering
\includegraphics[width=1.0\linewidth]{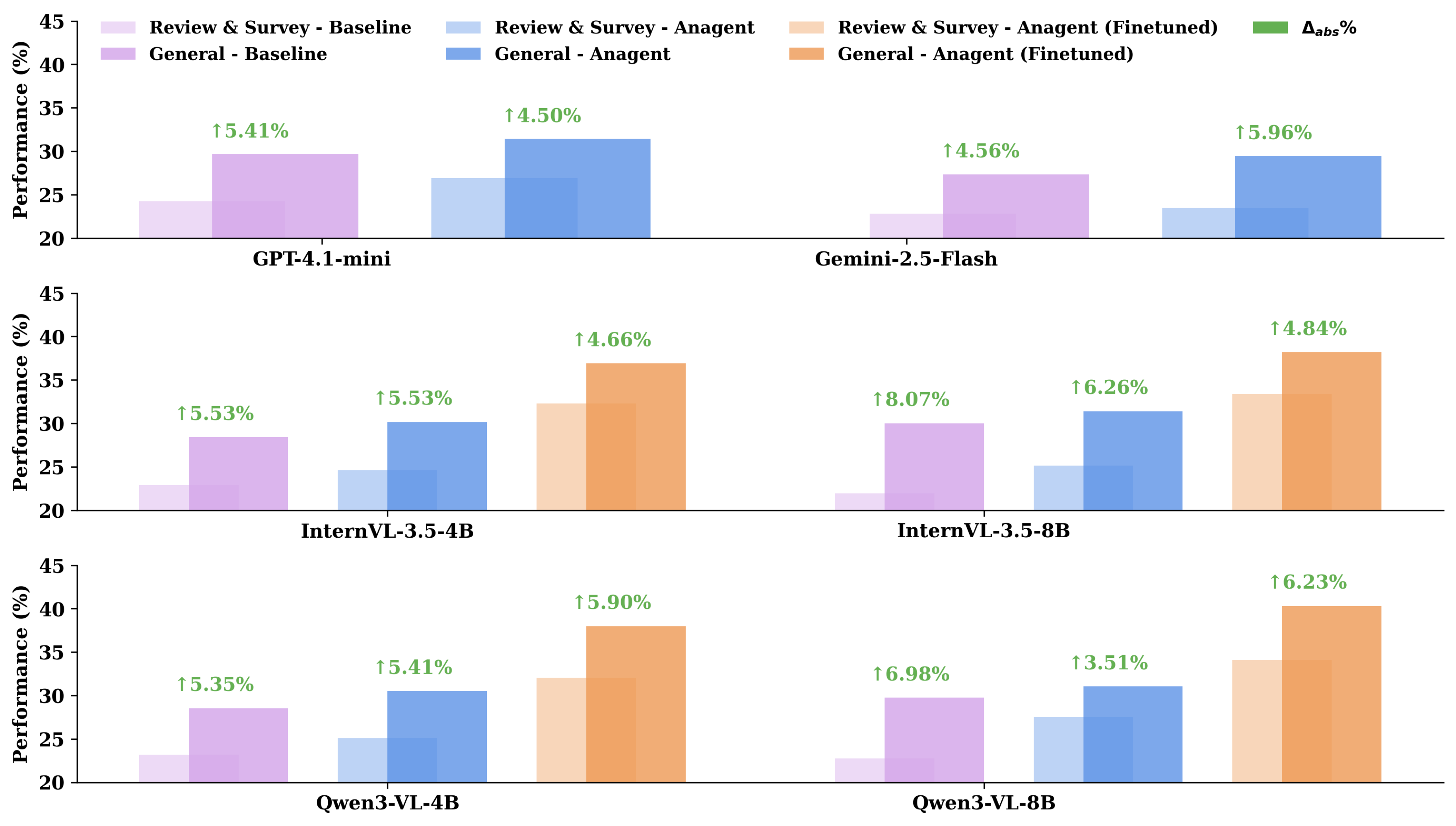}

\caption{\textbf{Ablation Study on Data Source.} Evaluation of agent performance across different data sources (\S\ref{appendix:benchmark:curriculum:data_complexity}).}
\label{fig:exp:ablation_complexity_data:source}
\end{figure}

%% file: figures/exp_ablation_complexity_data_domain.tex
\begin{figure}[H]
\centering
\includegraphics[width=1.0\linewidth]{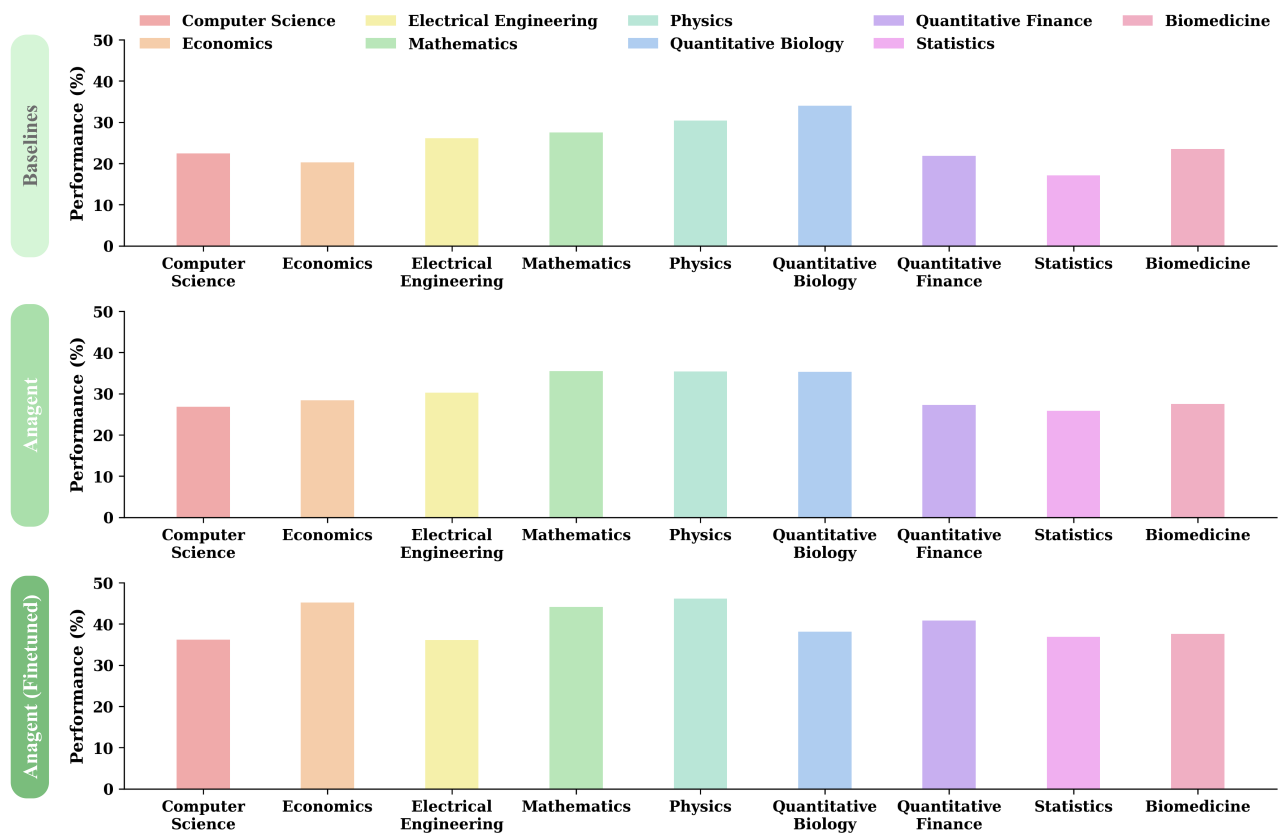}

\caption{\textbf{Ablation Study on Data Domain.} Evaluation of agent performance across nine broad domains (\S\ref{appendix:benchmark:curriculum:data_complexity}).}
\label{fig:exp:ablation_complexity_data:domain}
\end{figure}

%% file: figures/exp_ablation_complexity_data_coverage.tex
\begin{figure}[H]
\centering
\includegraphics[width=0.9\linewidth]{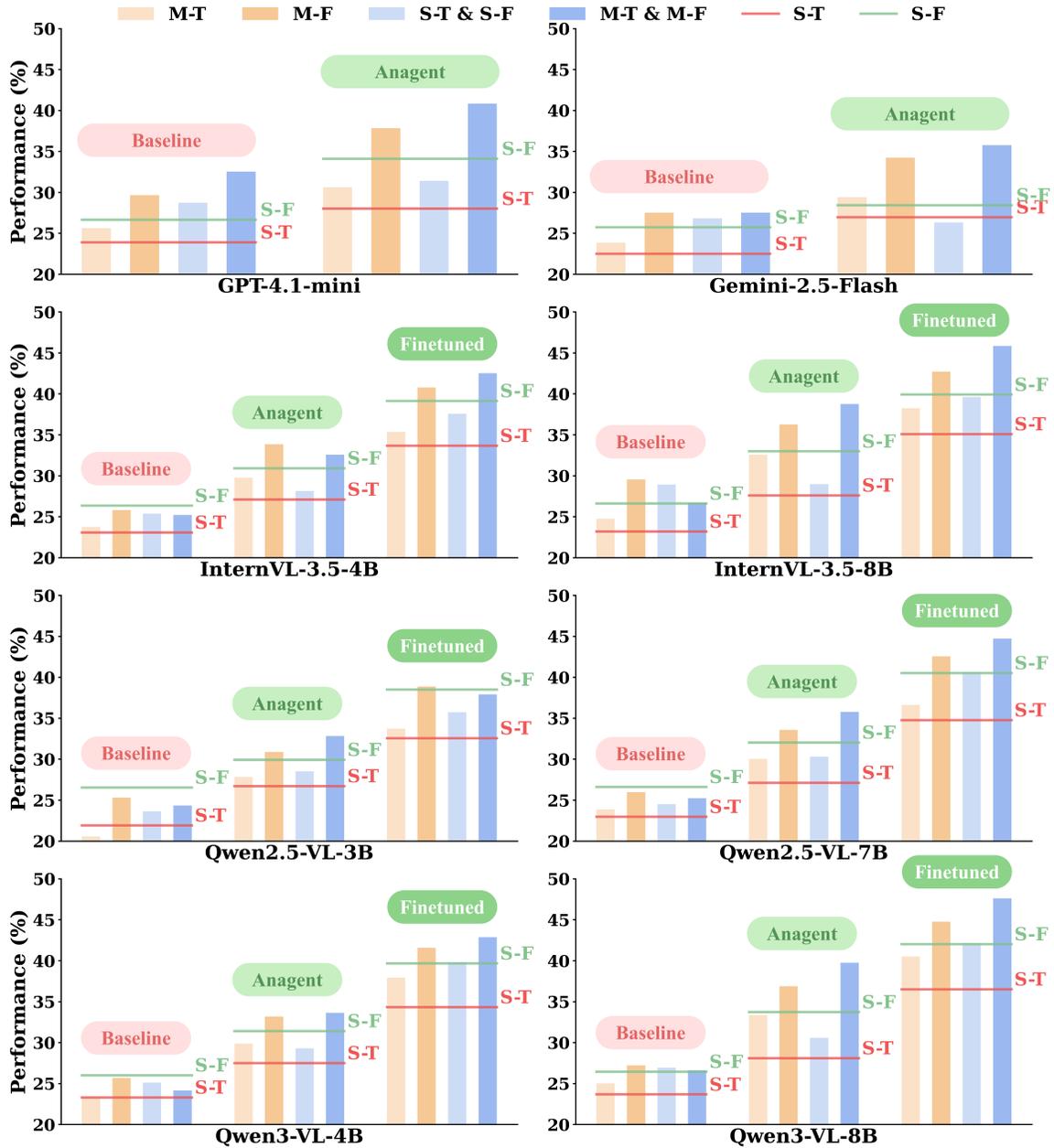}

\caption{\textbf{Ablation Study on Input Coverage.} Evaluation of agent performance across six different input coverage settings. \texttt{S-T} represents single-table input, \texttt{S-F} represents single-figure input, \texttt{M-T} represents multi-table input, and \texttt{M-F} represents multi-figure input. }
\label{fig:exp:ablation_complexity_data:coverage}
\end{figure}

%% file: figures/exp_ablation_complexity_analysis_width.tex
\begin{figure}[H]
\centering
\includegraphics[width=1.0\linewidth]{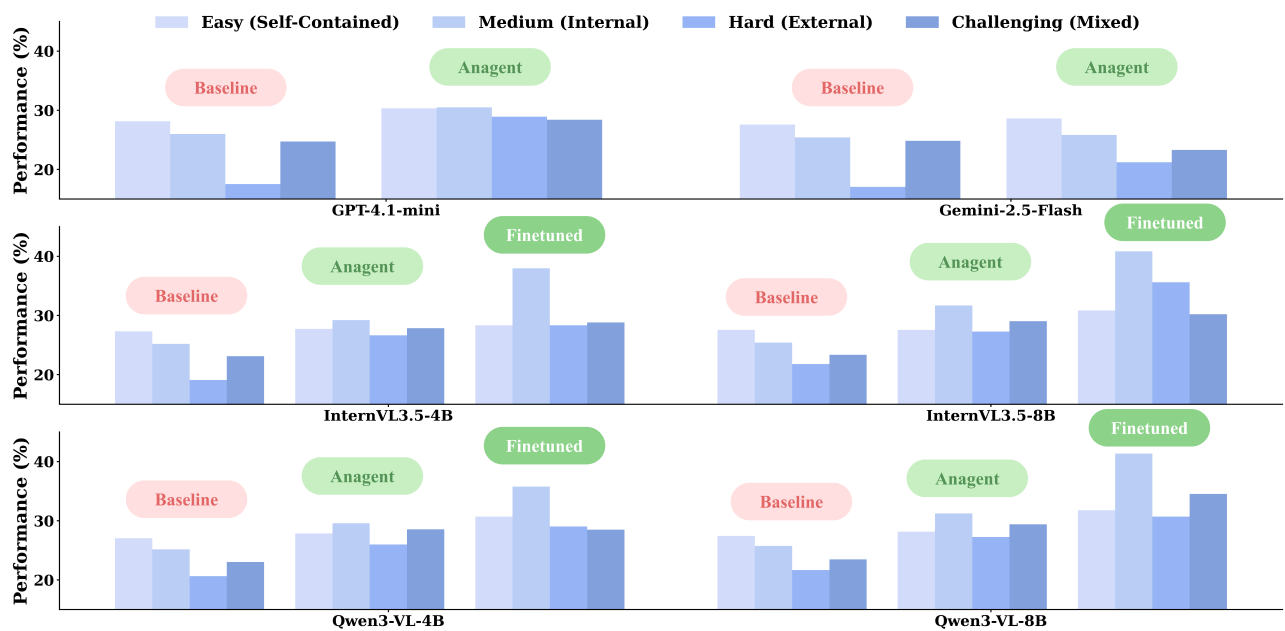}

\caption{\textbf{Ablation Study on Analysis Width.} Evaluation of agent performance across four analysis width curriculum (\S\ref{appendix:benchmark:curriculum:analysis_complexity}). }
\label{fig:exp:ablation_complexity_analysis:width}
\end{figure}

%% file: figures/exp_ablation_complexity_analysis_depth.tex
\begin{figure}[H]
\centering
\includegraphics[width=1.0\linewidth]{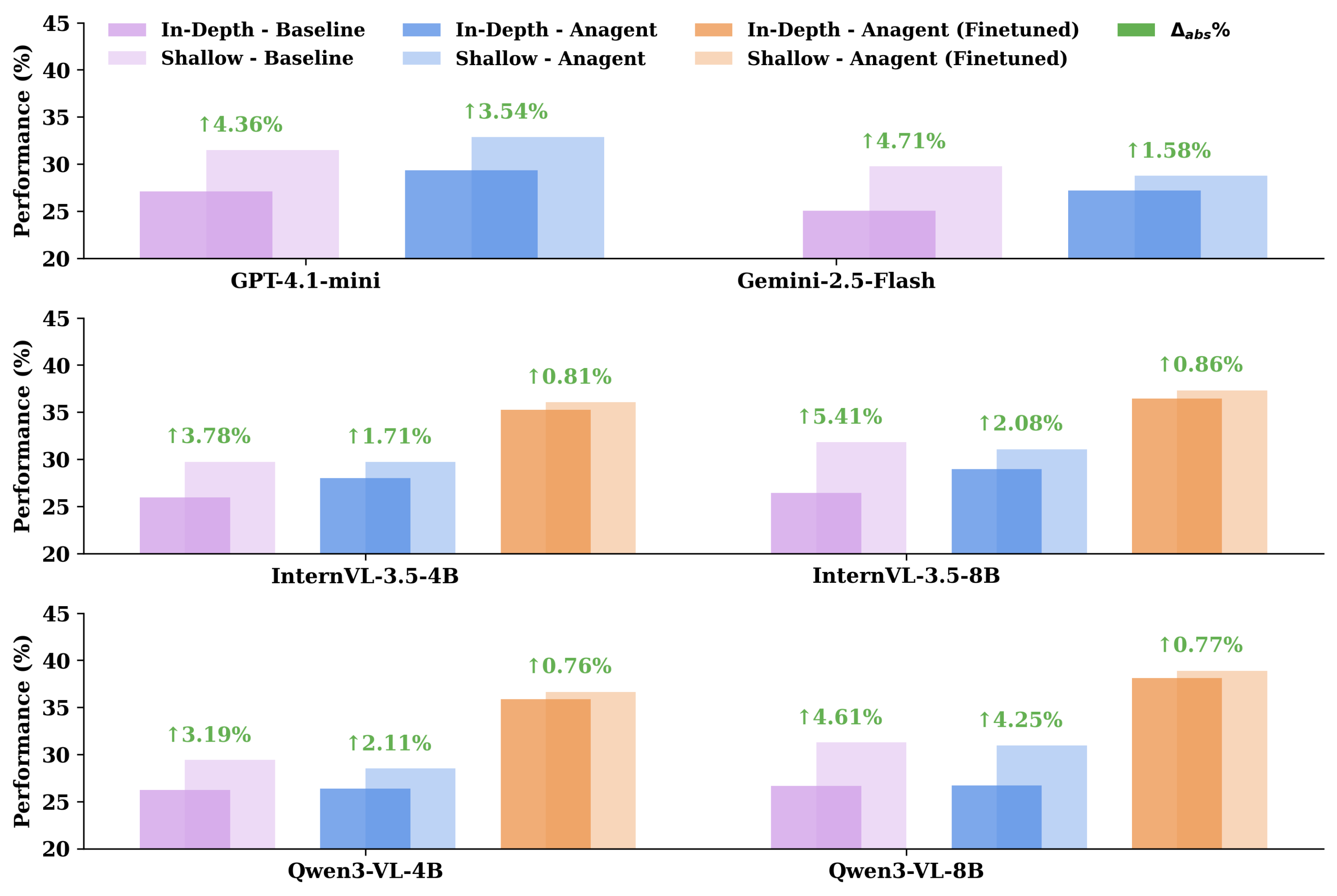}

\caption{\textbf{Ablation Study on Analysis Depth.} Evaluation of agent performance across two analysis depth: \textit{shallow analysis} and \textit{in-depth analysis} (\S\ref{appendix:benchmark:curriculum:analysis_complexity}). }
\label{fig:exp:ablation_complexity_analysis:depth}
\end{figure}

%% file: figures/exp_ablation_complexity_analysis_objective.tex
\begin{figure}[H]
\vspace{12pt}

\centering
\includegraphics[width=1.0\linewidth]{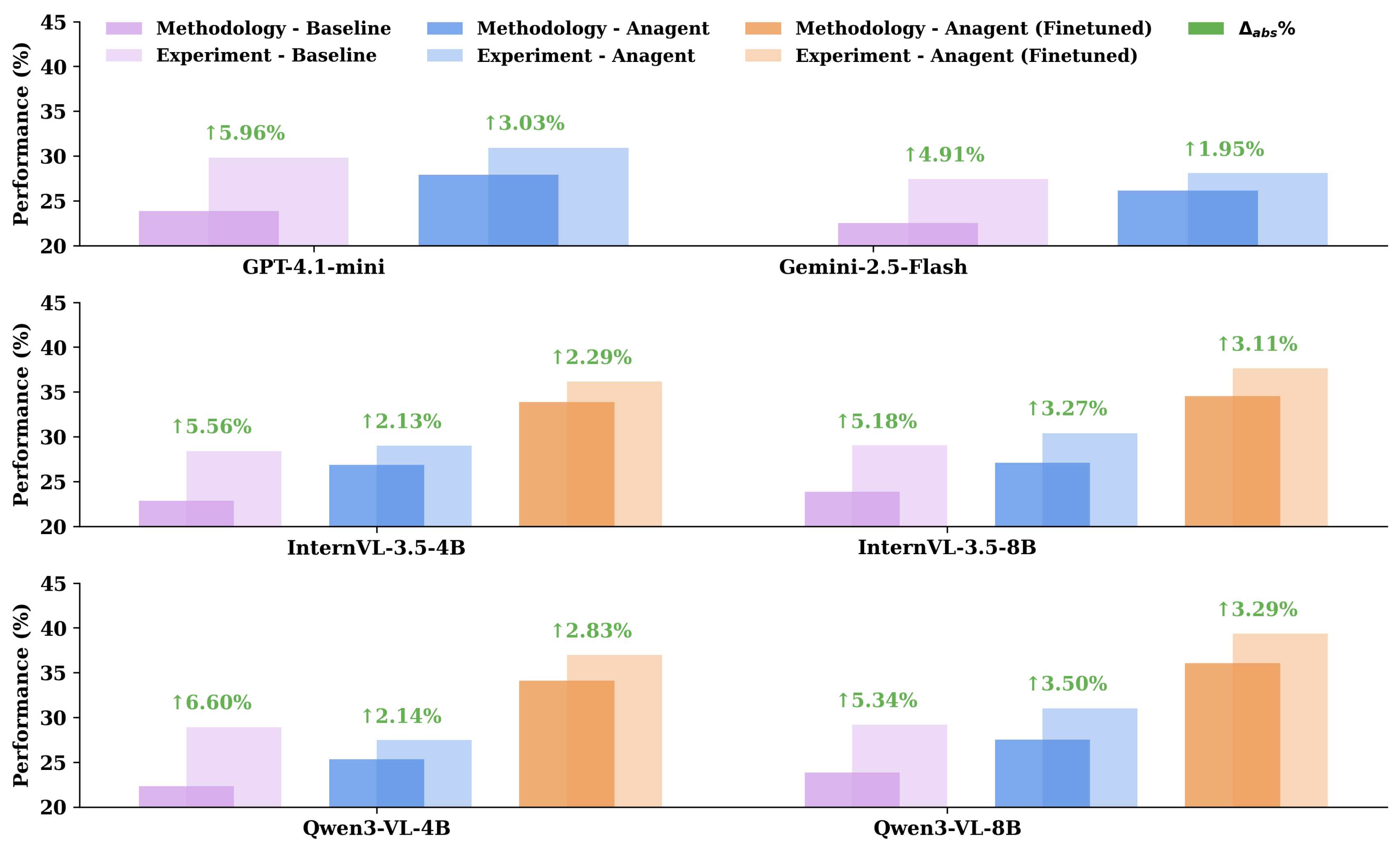}

\caption{\textbf{Ablation Study on Analysis Objective.} Evaluation of agent performance across two analysis objectives: \textit{methodology-oriented} analysis and \textit{experiment-oriented} analysis (\S\ref{appendix:benchmark:curriculum:analysis_complexity}).}

\vspace{12pt}

\label{fig:exp:ablation_complexity_analysis:objective}
\end{figure}

%% file: figures/exp_human_domain_eval.tex
\begin{figure*}[!t]
\centering
\includegraphics[width=1.0\linewidth]{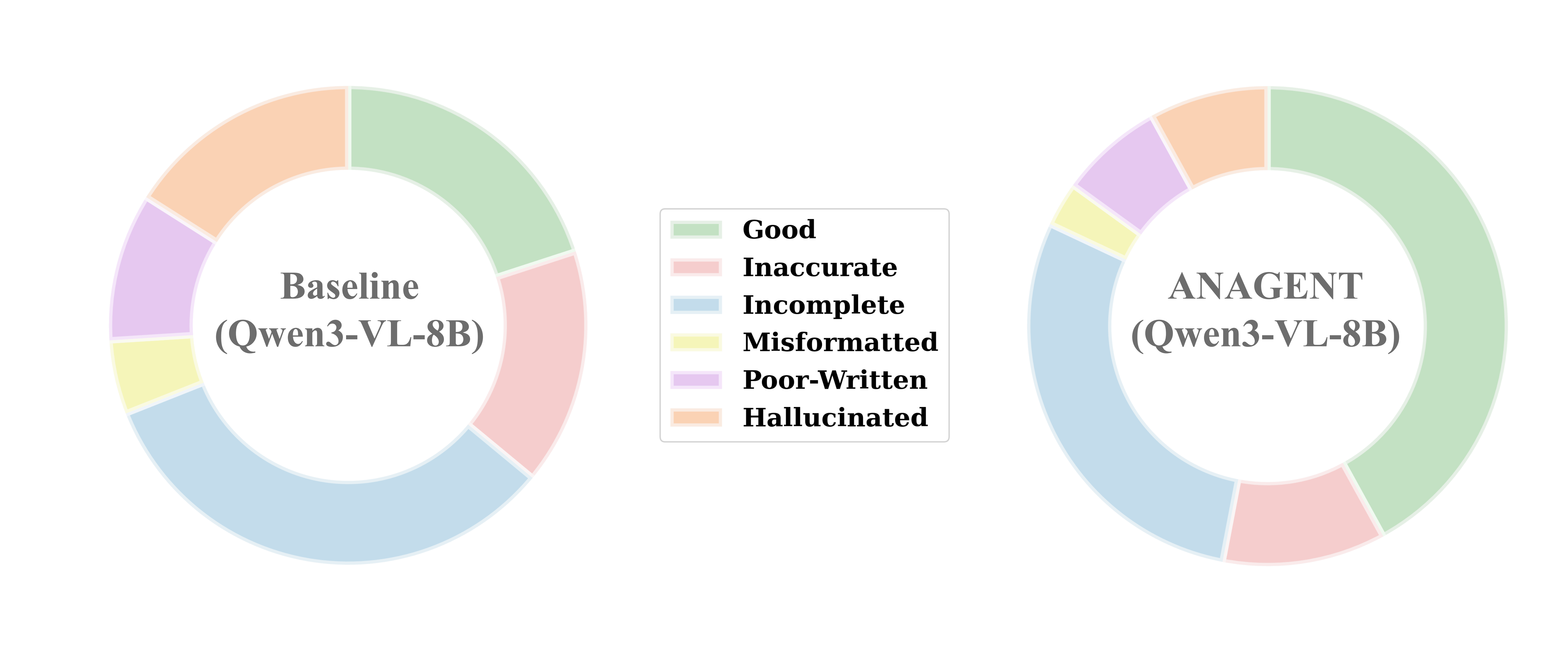}

\caption{\textbf{Domain Evaluation.} Assessing 100 samples manually for each model on each domain, we compare the evaluation results between baseline and \ours using the same base model (Qwen3-VL-8B).}

\vspace{12pt}

\label{fig:exp:human_domain_eval}
\end{figure*}

%% file: tables/human_eval.tex
\begin{table}[H]

\vspace{24pt}

\small
\centering
\renewcommand{\arraystretch}{1.3}
\setlength{\tabcolsep}{4pt}

\begin{tabular*}{\textwidth}{@{\extracolsep{\fill}}l cc ccc cc@{}}
\toprule
\multirow{2}{*}{\textbf{Dimension}} & \multicolumn{2}{c}{\textbf{MLLM Judge (\%)}} & \multicolumn{3}{c}{\textbf{Human Expert (\%)}} & \multicolumn{2}{c}{\textbf{Agreement \& Significance}} \\
\cmidrule(lr){2-3} \cmidrule(lr){4-6} \cmidrule(lr){7-8}
& \textbf{Baseline} & \ours & \textbf{Baseline} & \ours & \textbf{Gold} & \textbf{Krippendorff's} $\bm{\alpha}$ & \textbf{Wilcoxon} $\bm{p}$ \\

\midrule

\multicolumn{8}{c}{\cellcolor{baselinebg}\textbf{Five-Dimensional Evaluation Protocol}} \\
\addlinespace[3pt]

\textbf{Analysis Consistency} & 48.23 & 59.34 & 52.57 & 64.52 & 92.78 & 0.92 & $<0.001$ \\
\textbf{Query-Analysis Alignment} & 29.32 & 33.03 & 32.63 & 37.42 & 90.28 & 0.87 & $<0.001$ \\
\textbf{Knowledge Utilization} & 67.63 & 78.54 & 71.24 & 83.15 & 93.72 & 0.88 & $<0.001$ \\
\textbf{Format Correctness} & 70.22 & 76.36 & 72.18 & 80.60 & \textbf{99.37} & \textbf{0.97} & $<0.001$ \\
\textbf{Grounding Accuracy} & 55.37 & 67.91 & 58.52 & 72.23 & 92.74 & 0.82 & $<0.001$ \\

\midrule

\textbf{Overall} & \textbf{54.15} & \textbf{63.04} & \textbf{57.43} & \textbf{67.58} & \textbf{93.78} & \textbf{0.89} & $\bm{<0.001}$ \\

\bottomrule
\end{tabular*}

\vspace{0pt}

\caption{\textbf{Multi-Domain Human Expert Evaluation.} Three experts per domain rate agent-generated and gold analyses across six domains under the five-dimensional protocol (\S\ref{appendix:evaluation:five_dimension}). \textbf{Gold} shows expert ratings of \ourbench's author-written reference analyses under the identical protocol.}
\label{tab:appendix:human_expert_multidomain}

\vspace{12pt}

\end{table}

%% file: figures/exp_ablation_tool_call.tex
\begin{figure}[H]
\centering
\includegraphics[width=1.0\linewidth]{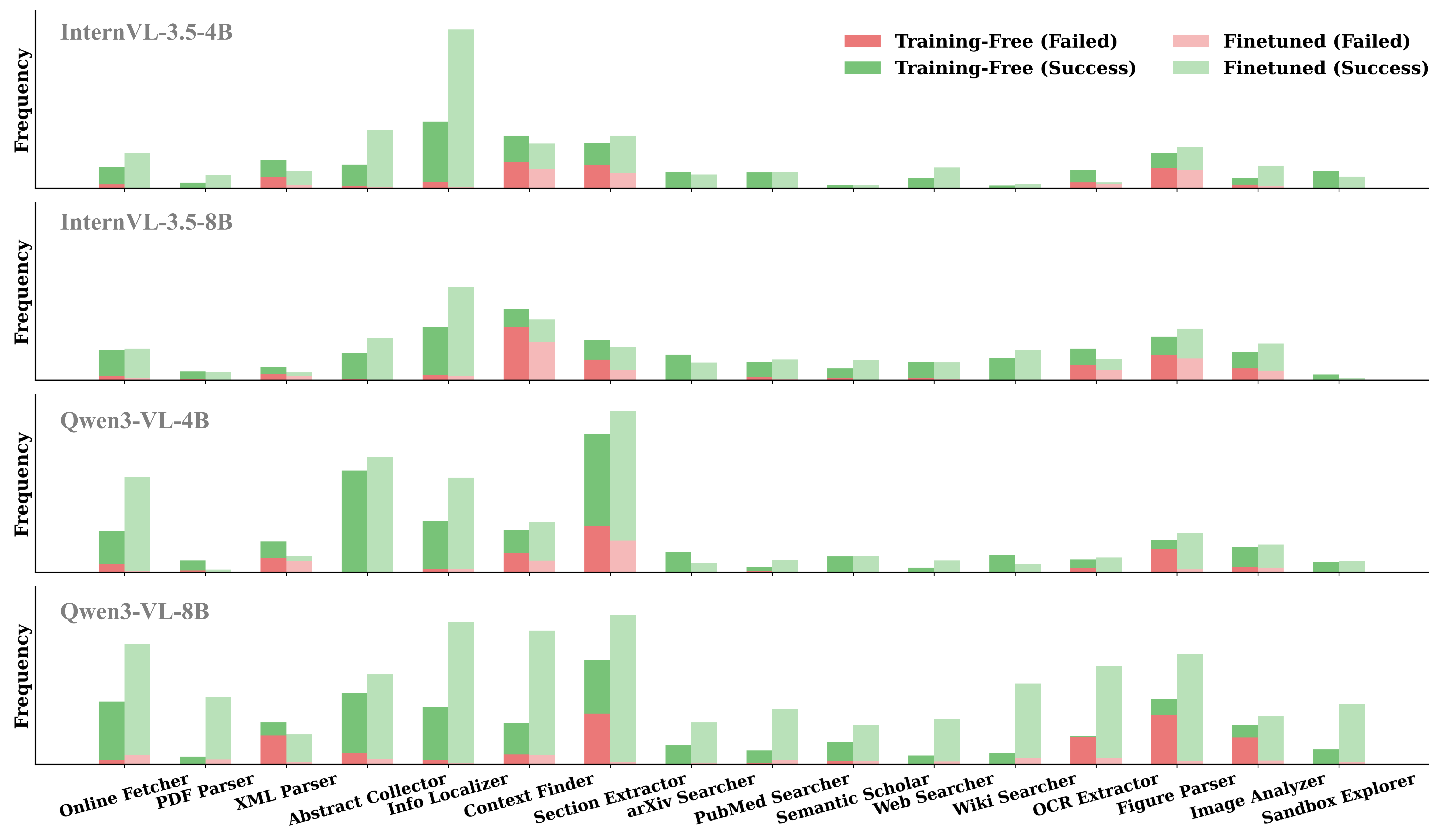}

\caption{\textbf{\ours Tool Utilization.} Investigation on tool utilization of training-free \ours and finetuned \ours.}

\vspace{12pt}

\label{fig:exp:ablation_tool_call}
\end{figure}

%% file: figures/exp_ablation_tool_strategy.tex
\begin{figure}[H]
\centering
\includegraphics[width=1.0\linewidth]{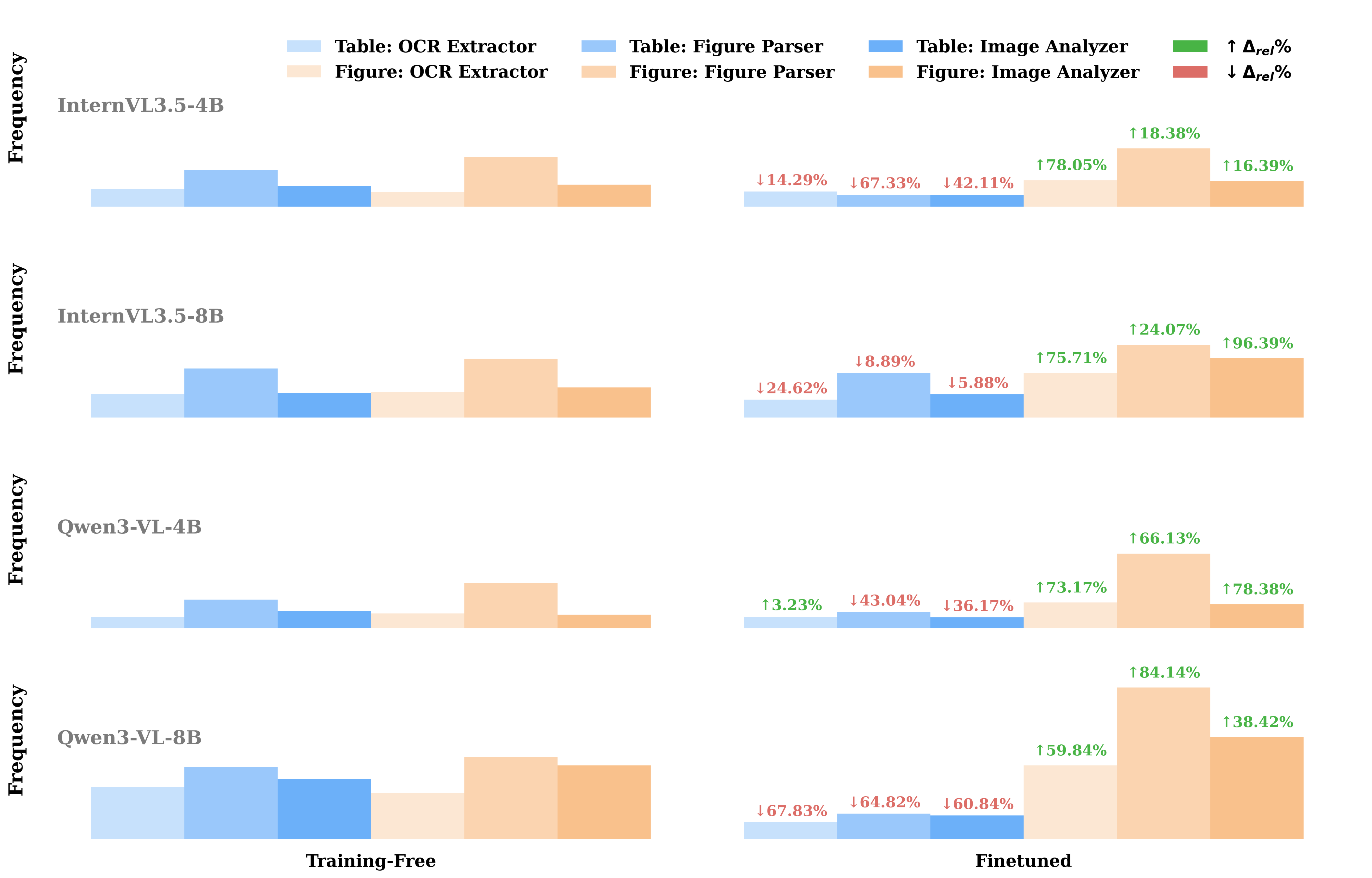}

\caption{\textbf{\ours Tool Utilization Strategy Optimization Through Finetuning.} Optimization of tool utilization strategy through finetuning.}

\vspace{12pt}

\label{fig:exp:ablation_tool_strategy}
\end{figure}

%% file: tables/controlled_exps.tex
\begin{table}[t]

\vspace{0pt}

\small
\centering
\renewcommand{\arraystretch}{1.3}
\setlength{\tabcolsep}{6pt}

\begin{tabular*}{\textwidth}{@{\extracolsep{\fill}}l ccc cc}
\toprule

\textbf{Method} & \textbf{MLLM Calls} & \textbf{Tokens/Call} & \textbf{Latency} & $\bm{S}_{\textbf{\textsc{Avg}}}$ & $\bm{\Delta}_{\textbf{\textsc{Rel}}}$ \\

\midrule

\multicolumn{6}{c}{\cellcolor{examplebg}\textbf{Single-Model Test-Time Scaling}} \\
\addlinespace[3pt]

Direct Generation & 1.00 & 1.00$\times$ & 1.00$\times$ & 26.76 & -- \\
Self-Consistency (SC@10) & 10.00 & 1.18$\times$ & 7.09$\times$ & 26.82 & \reldeltapct{26.76}{26.82} \\
Best-of-$N$ (BoN@10)$^{\dagger}$ & 10.00 & 1.07$\times$ & 7.11$\times$ & 27.03 & \reldeltapct{26.76}{27.03} \\
Self-Refine (SR@10) & 10.00 & 4.06$\times$ & 7.47$\times$ & 26.79 & \reldeltapct{26.76}{26.79} \\

\addlinespace[3pt]
\multicolumn{6}{c}{\cellcolor{oursbaselinebg}\textbf{Multi-Agent (Training-Free)}} \\
\addlinespace[3pt]

\textit{Omnion} & 5.48 & 4.87$\times$ & 7.26$\times$ & 22.87 & \reldeltapct{26.76}{22.87} \\
\ours & 6.64 & 4.52$\times$ & \textbf{6.24}$\times$ & 28.73 & \reldeltapct{26.76}{28.73} \\

\addlinespace[3pt]
\multicolumn{6}{c}{\cellcolor{ourstrainedbg}\textbf{Multi-Agent (Modular Training)}} \\
\addlinespace[3pt]

\ours (SFT) & 6.79 & 4.38$\times$ & 6.36$\times$ & \textbf{36.02} & \reldeltapctbf{26.76}{36.02} \\

\bottomrule
\end{tabular*}

\vspace{0pt}

\caption{\textbf{Controlled Test-Time Scaling \& Compute Accounting.} All experiments use the same Qwen3-VL-4B backbone and the same evaluation set (\S\ref{appendix:extended_experiment_analysis:context_significance}). MLLM calls, tokens per call, and latency are averaged per instance, with tokens per call and latency calculated relative to single-pass direct generation. \textit{Relative differences} (Eq.~\ref{eq:relative_delta}) are shown as \textit{positive} $\uparrow\Delta_{rel}\%$ or \textit{negative} $\downarrow\Delta_{rel}\%$. $^{\dagger}$BoN@10 selects the candidate with the highest $S_{\textsc{Avg}}$ against the reference analysis, and is therefore an \textit{oracle upper bound} rather than a deployable selection rule.}
\label{tab:appendix:controlled_test_time_scaling}

\vspace{18pt}

\end{table}

%% file: figures/good_analysis_examples.tex

\begin{figure}[H]
\centering
\includegraphics[width=1.0\linewidth]{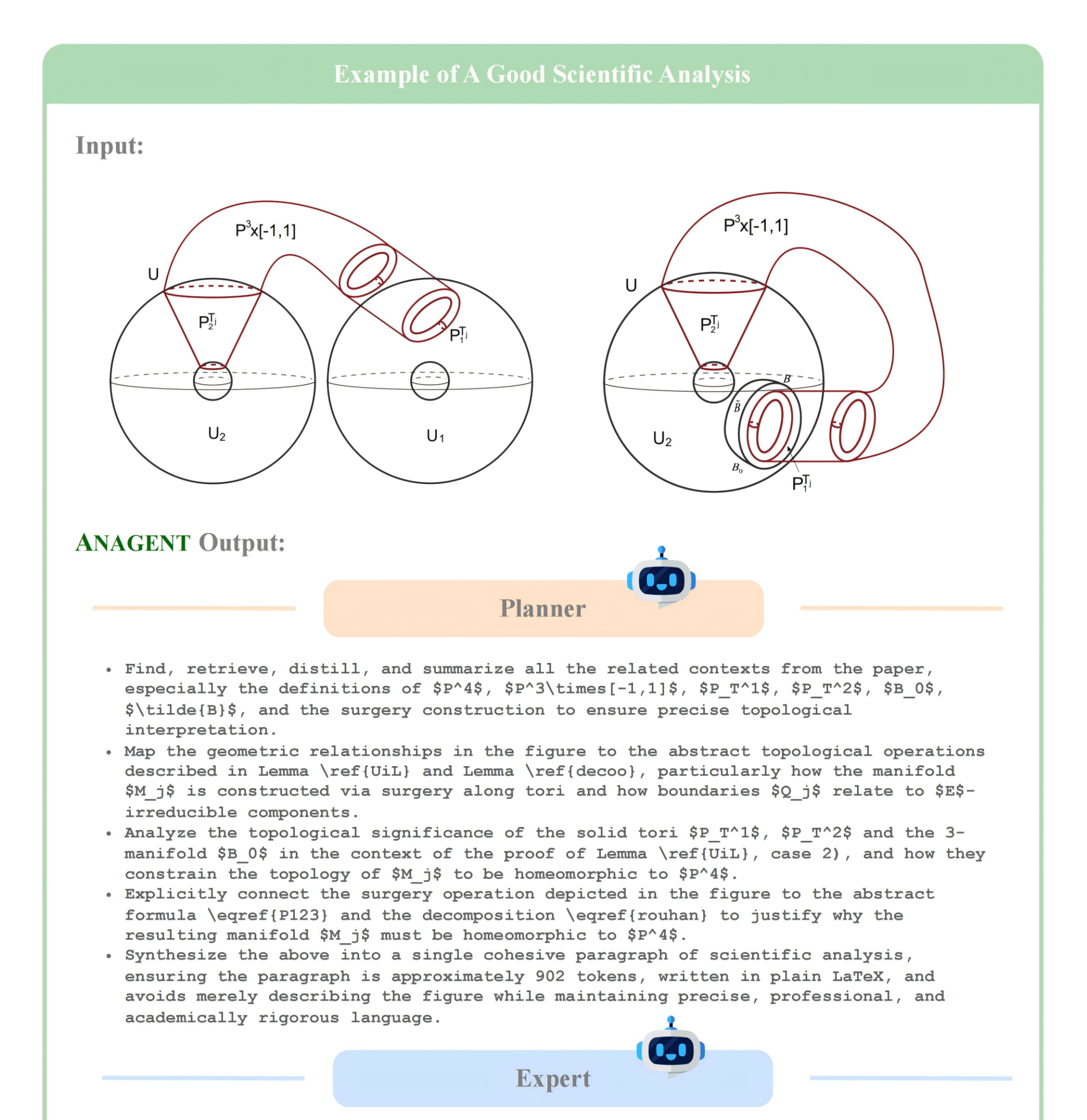}

\end{figure}

\begin{figure}[H]
\centering
\includegraphics[width=1.0\linewidth]{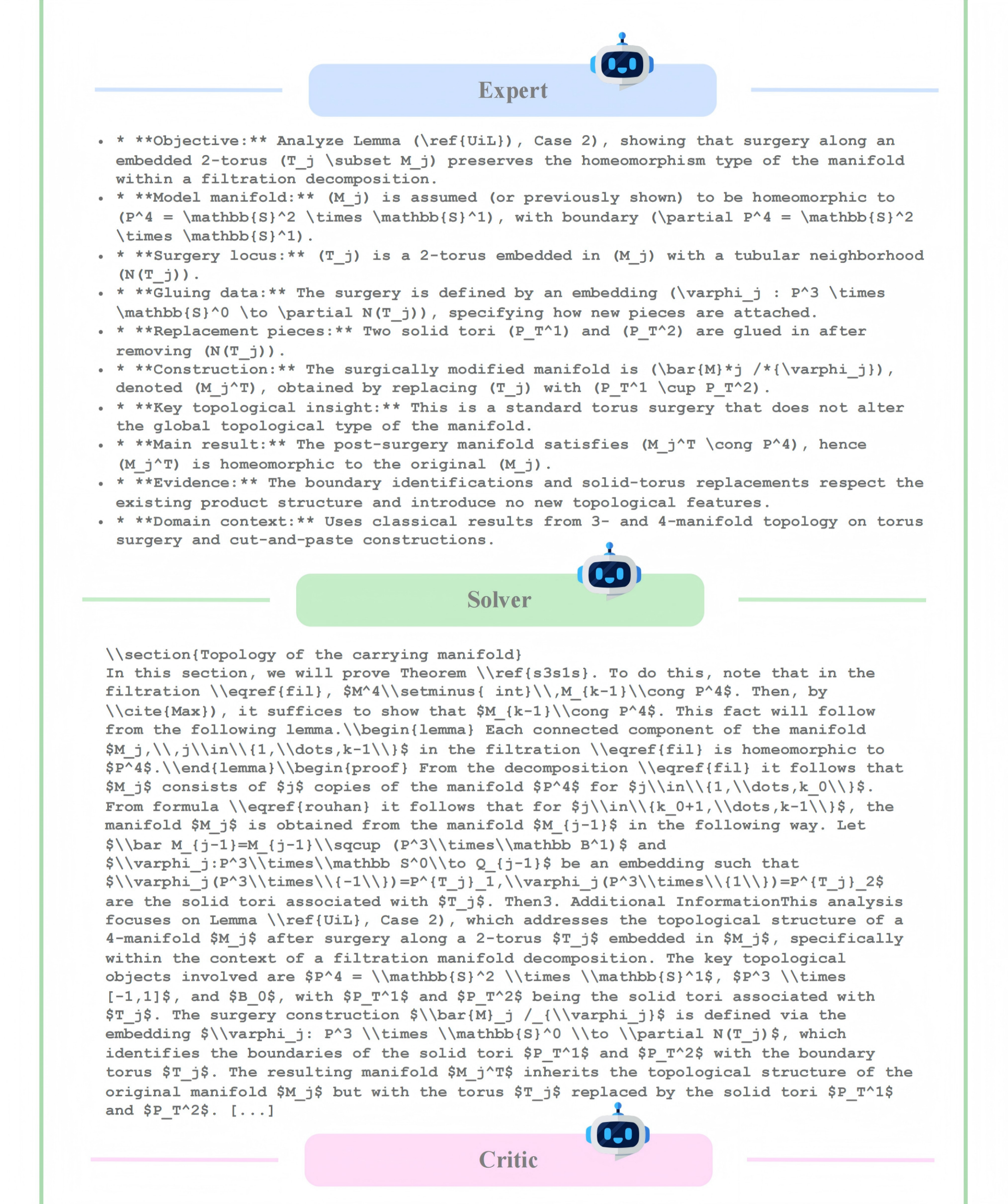}

\end{figure}

\begin{figure}[H]
\centering
\includegraphics[width=1.0\linewidth]{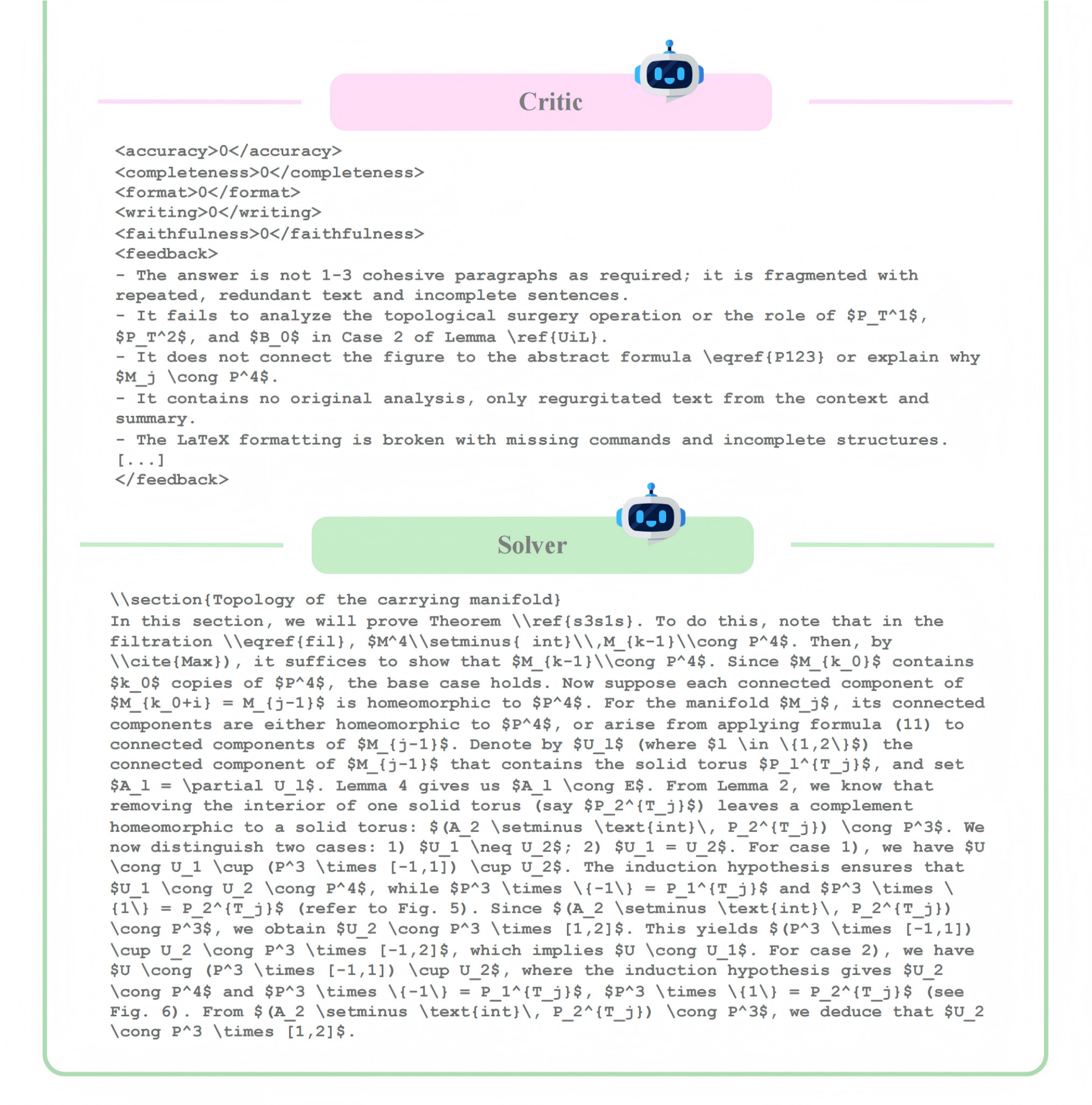}

\caption{\textbf{Example of \ours Scientific Analysis Writing.} This example ($S_{\textsc{AVG}}=72.41\%$) shows a complete end-to-end scientific analysis process of \ours, illustrating the full scientific analysis writing workflow from task planning, searching and executing, context-aware problem-solving, to iterative reflection and refinement. Input query and agent prompts are provided in \S\ref{appendix:anagent:collaboration} and omitted here for clarity.}
\label{fig:example:good_scientific_analysis}

\end{figure}

%% file: figures/failure_analysis_error_decrease.tex
\begin{figure}[H]
\centering
\includegraphics[width=1.0\linewidth]{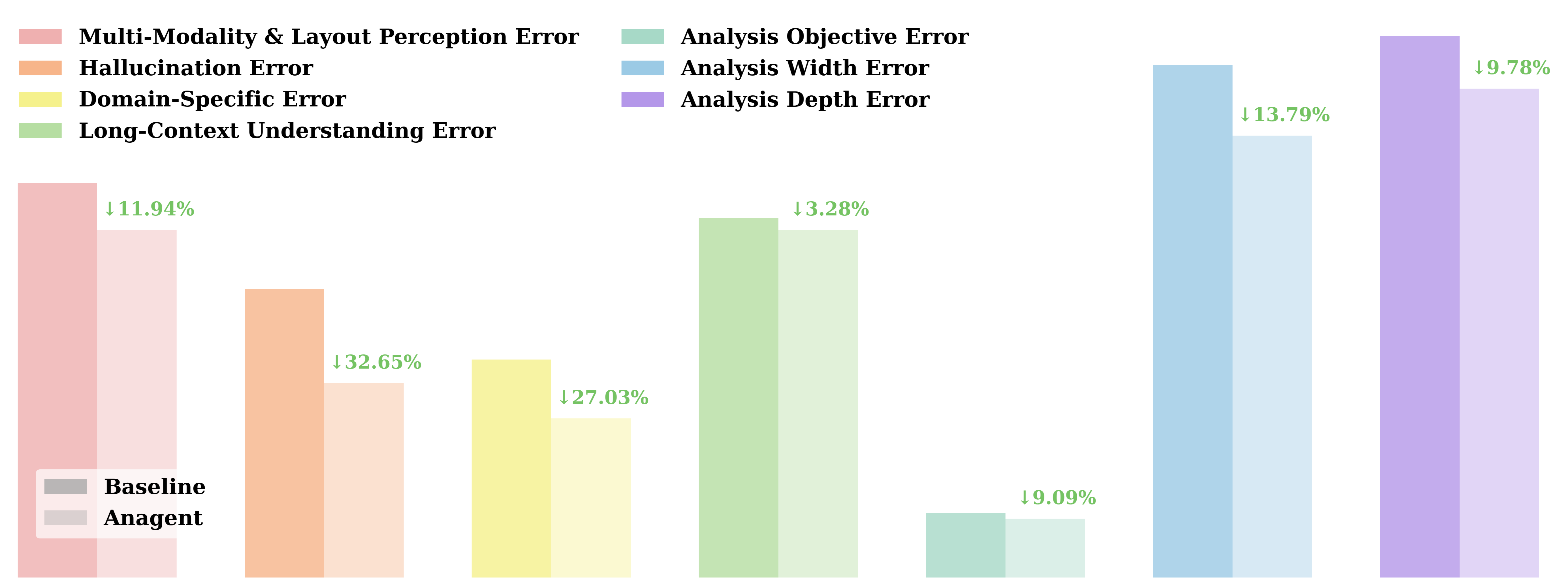}

\caption{\textbf{Examination On Error Distribution.} Comparison of the seven key error patterns (Fig.~\ref{fig:preliminary-error-analysis}) between baseline and \ours on the same set. Similar to Fig.~\ref{fig:preliminary-error-analysis}, the Y axis represents ``Error Rate (\%)" from low to high.}
\label{fig:exp:failure_analysis:error_distribution}
\end{figure}

%% file: sections/limitations.tex
\section{Limitations}
\label{sec:limitations}

By proposing \ourbench, we introduce the task of scientific table \& figure analysis, which is one of the core stages for scientific research. While \ours demonstrates promising capabilities in scientific table \& figure analysis, we acknowledge several limitations that we aim to address in our future work:

\paragraph{Computational Overhead.}
The multi-agent architecture with iterative execution and refinement introduces additional computational overhead as compared to single-pass generation. In practice, increasing the maximum number of iterations allowed for \ours may consume more computational resources. To mitigate this, we implement engine pre-checking during the initialization of \ours to avoid redundant MLLM engine initialization, and we plan to further optimize GPU utilization efficiency in future work.

\paragraph{Domain Coverage.}
Although \ours spans nine scientific domains across 170 fine-grained disciplines, it primarily focuses on arXiv and PubMed publications. The generalization of \ours to emerging scientific domains, non-English literature, and alternative dissertation platforms remains to be validated.

\paragraph{Tool Dependency.}
The effectiveness of \ours is closely tied to the quality and reliability of provided scientific toolkits. Unsuccessful tool calls, as restrained by backbone MLLM reasoning capabilities or truncated tool execution outputs due to MLLM context window limits, can directly impact analysis quality. These challenges motivates us to explore tool utilization and optimization in our future work.

\paragraph{Evaluation Challenges.}
While we employ multi-faceted evaluation including rule-based metrics, MLLM-as-Judge, and human expert assessment, the cost of MLLM-as-Judge and human expert assessment can be very large that substantially limits evaluation efficiency. We aim to investigate systematic evaluation with low costs in future research.

\noindent
Despite these limitations, we believe \ourbench and \ours represents a meaningful step toward more capable and reliable AI systems for scientific research, and we hope our work will inspire future investigations into scientific reasoning and analysis.

%% file: figures/failure_analysis_examples.tex

\begin{figure}[H]

\centering
\includegraphics[width=1.0\linewidth]{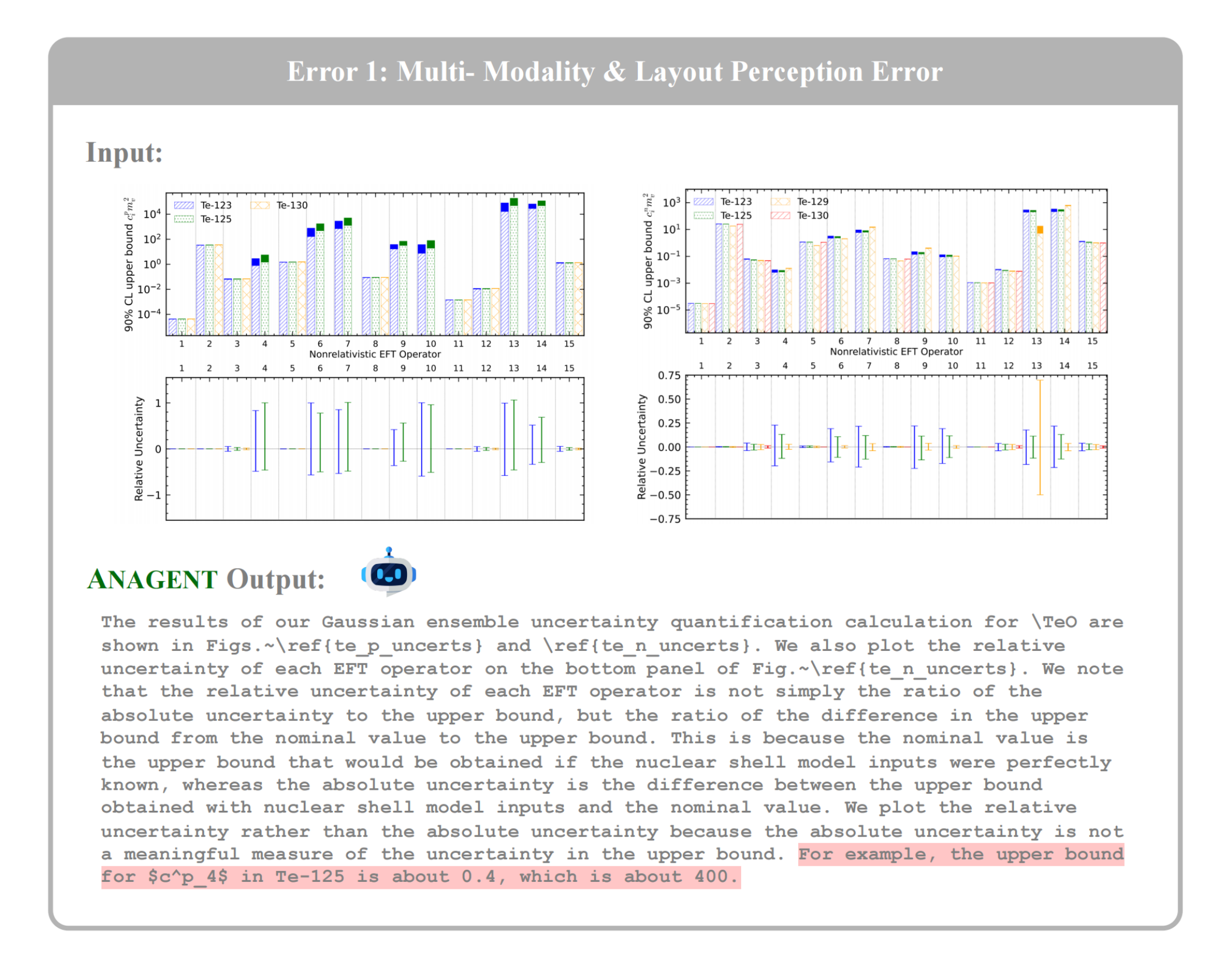}

\caption{\textbf{Example of Multi- Modality \& Layout Perception Error.} This example ($S_{\textsc{AVG}}=25.64\%$) illustrates a scientific analysis failure case with \textit{multi- modality \& layout perception errors} (\S\ref{appendix:subsec:failure_analysis:error1}). Input query and agent prompts are provided in \S\ref{appendix:anagent:collaboration} and omitted here for clarity. Other intermediate outputs and contexts are also omitted for clear presentation.}
\label{fig:failure_analysis:error1_perception_error}

\vspace{-6pt}
\end{figure}


\begin{figure}[H]

\centering
\includegraphics[width=1.0\linewidth]{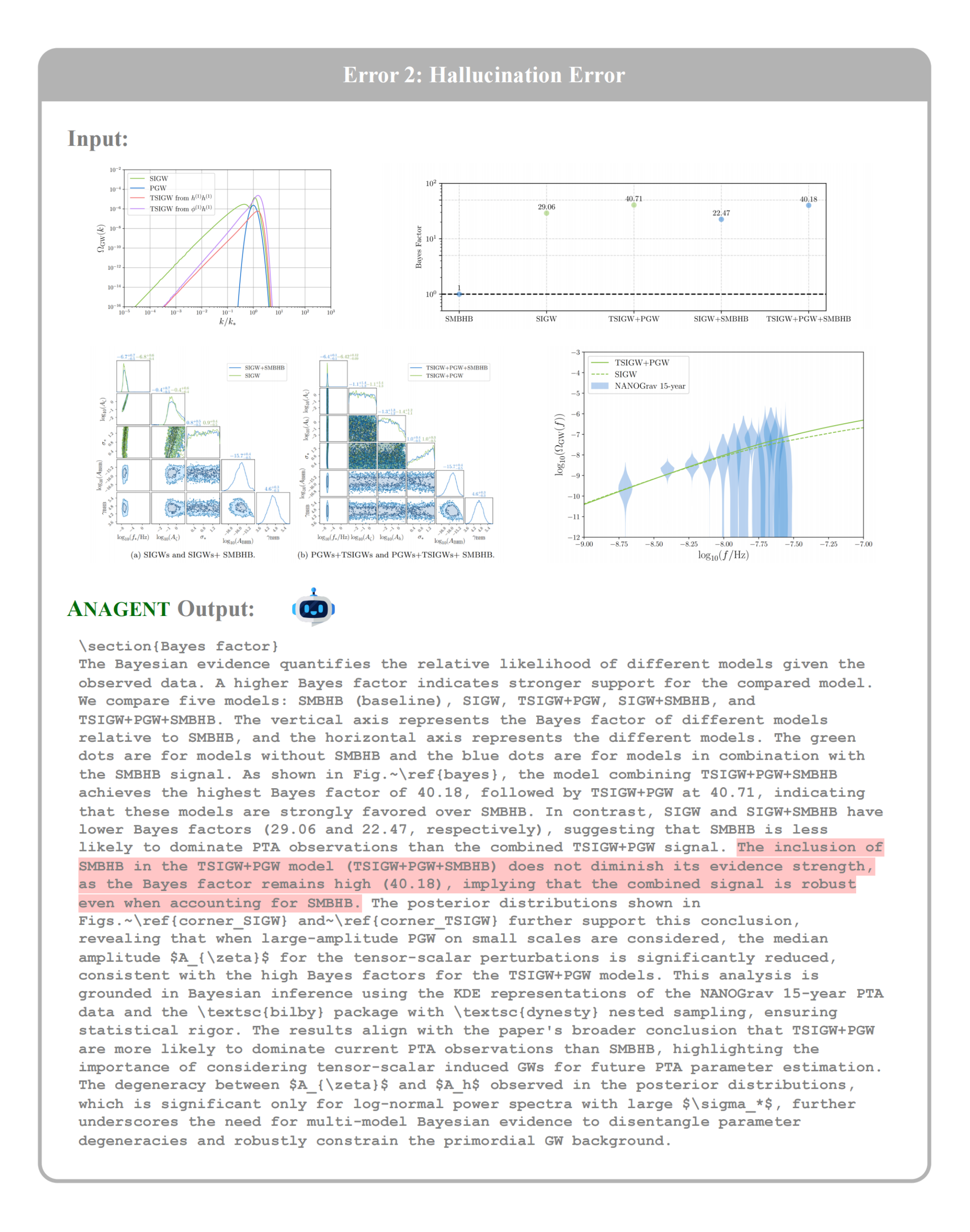}

\caption{\textbf{Example of Hallucination Error.} This example ($S_{\textsc{AVG}}=21.85\%$) illustrates a scientific analysis failure case with \textit{hallucination errors} (\S\ref{appendix:subsec:failure_analysis:error2}). Input query and agent prompts are provided in \S\ref{appendix:anagent:collaboration} and omitted here for clarity. Other intermediate outputs and contexts are also omitted for clearer presentation.}
\label{fig:failure_analysis:error2_hallucination_error}

\end{figure}


\begin{figure}[H]

\centering
\includegraphics[width=1.0\linewidth]{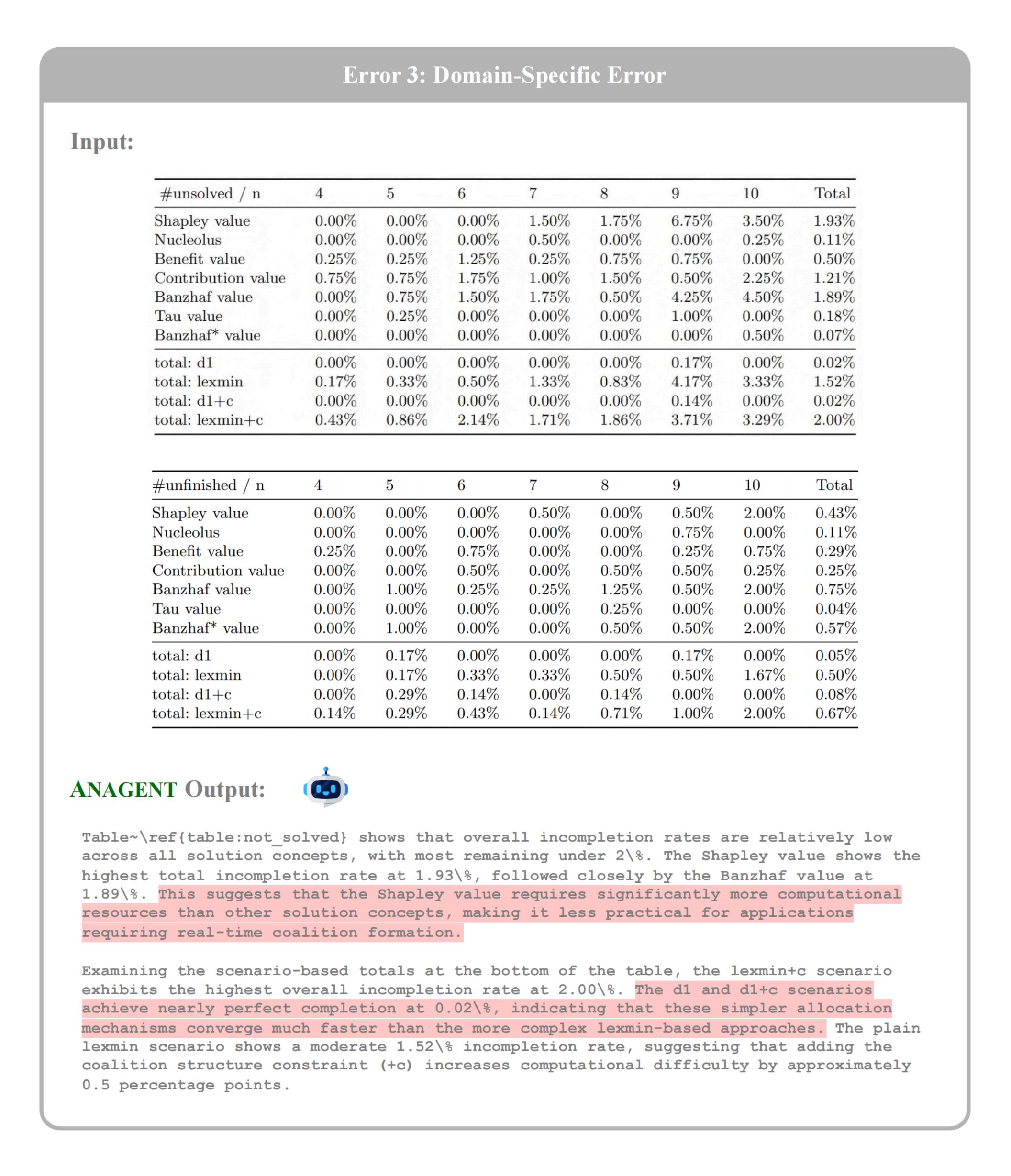}

\caption{\textbf{Example of Domain-Specific Error.} This example ($S_{\textsc{AVG}}=11.28\%$) illustrates a scientific analysis failure case with \textit{multi- modality \& layout perception errors} (\S\ref{appendix:subsec:failure_analysis:error3}). Input query and agent prompts are provided in \S\ref{appendix:anagent:collaboration} and omitted here for clarity. Other intermediate outputs and contexts are also omitted for clear presentation.}
\label{fig:failure_analysis:error3_domain_error}

\vspace{-6pt}
\end{figure}


\begin{figure}[H]

\centering
\includegraphics[width=1.0\linewidth]{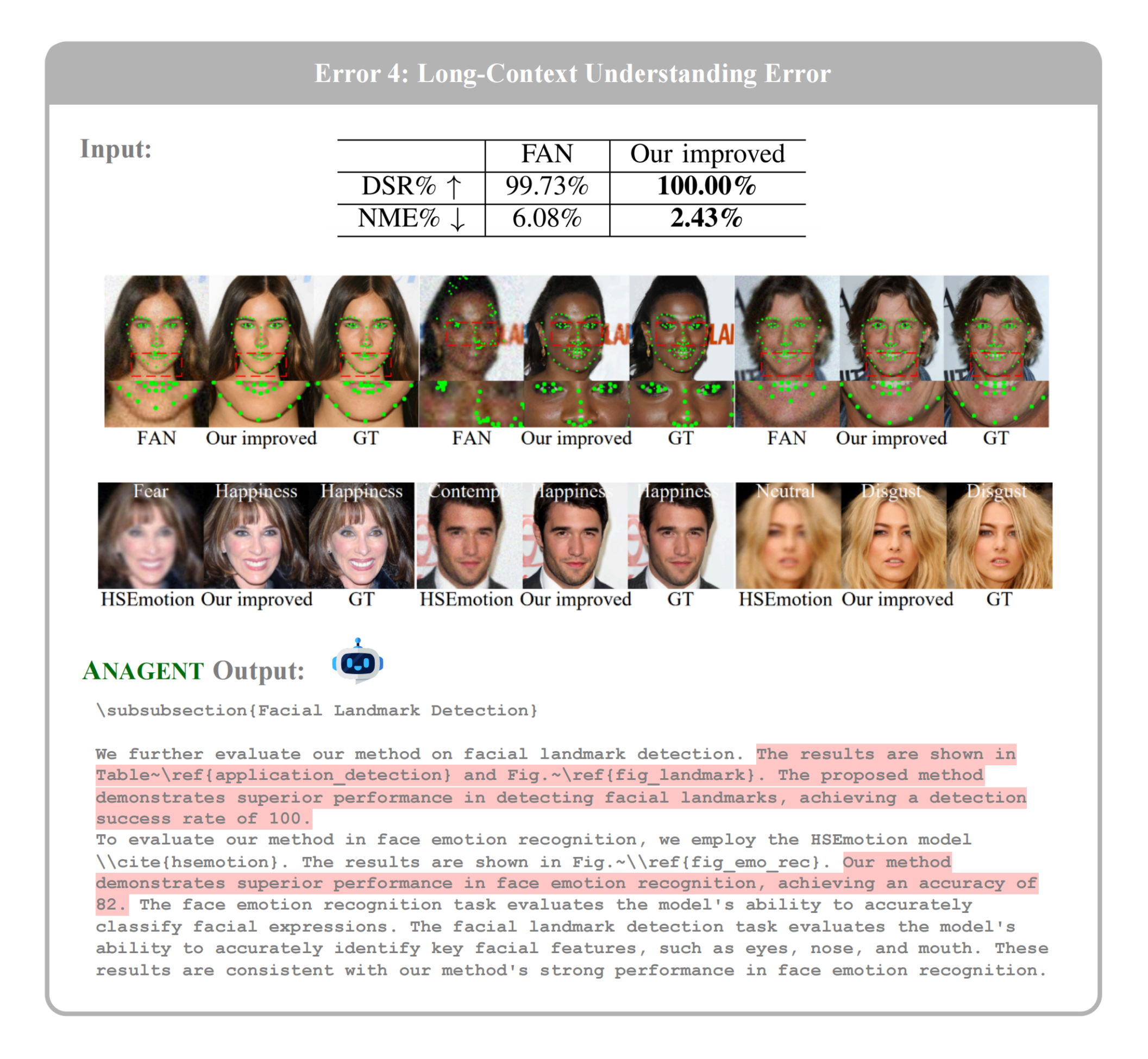}

\caption{\textbf{Example of Long-Context Understanding Error.} This example ($S_{\textsc{AVG}}=24.43\%$) illustrates a scientific analysis failure case due to \textit{long-context understanding errors} (\S\ref{appendix:subsec:failure_analysis:error4}). Input query and agent prompts are provided in \S\ref{appendix:anagent:collaboration} and omitted here for clarity. Other intermediate outputs and contexts are also omitted for clear presentation.}
\label{fig:failure_analysis:error4_long_context_error}

\vspace{-6pt}
\end{figure}


\begin{figure}[H]

\centering
\includegraphics[width=1.0\linewidth]{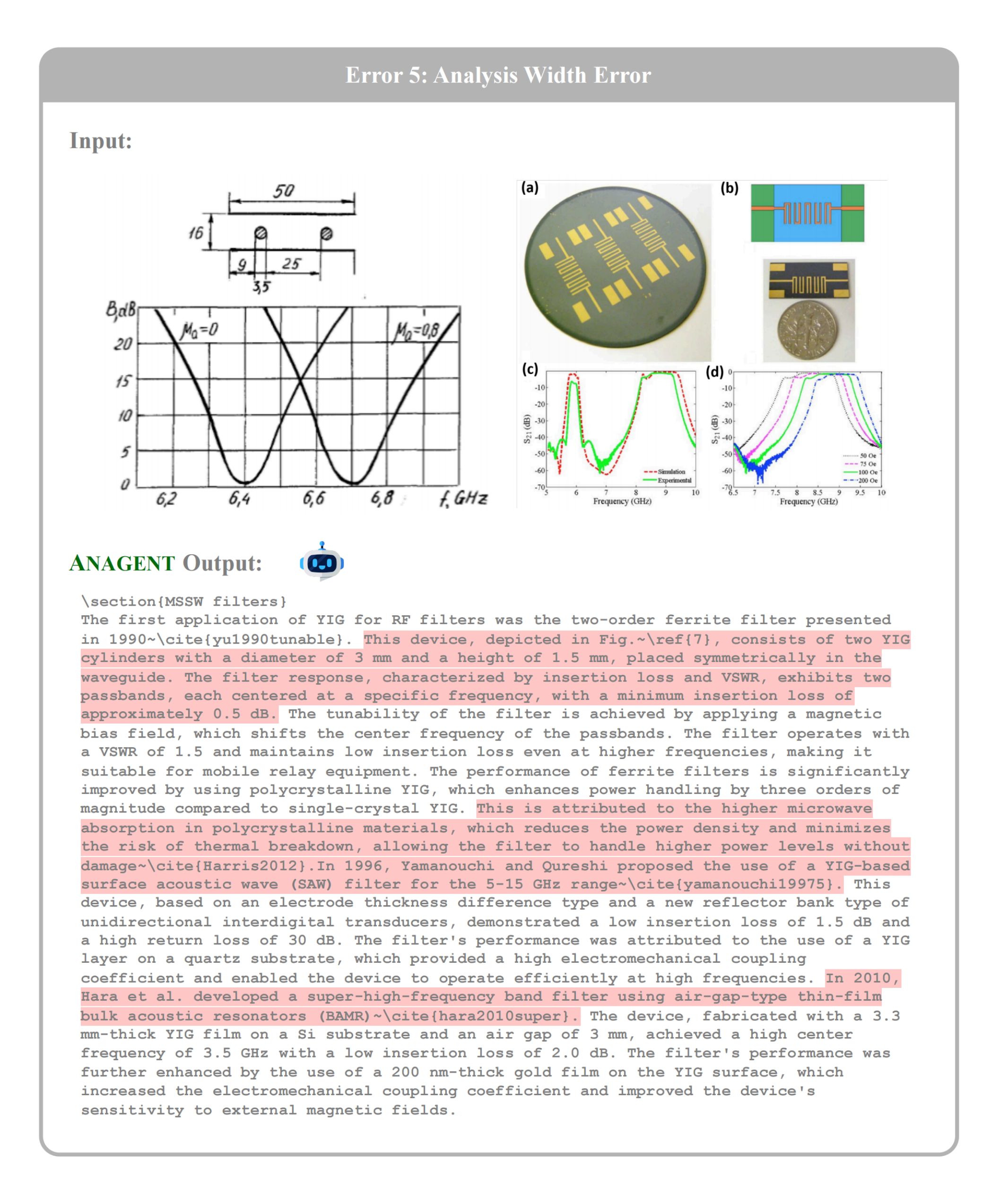}

\caption{\textbf{Example of Analysis Width Error.} This example ($S_{\textsc{AVG}}=25.96\%$) illustrates a scientific analysis failure case due to \textit{analysis width errors} (\S\ref{appendix:subsec:failure_analysis:error5}). Input query and agent prompts are provided in \S\ref{appendix:anagent:collaboration} and omitted here for clarity. Other intermediate outputs and contexts are also omitted for clear presentation.}
\label{fig:failure_analysis:error5_width_error}

\vspace{-6pt}
\end{figure}


\begin{figure}[H]

\centering
\includegraphics[width=1.0\linewidth]{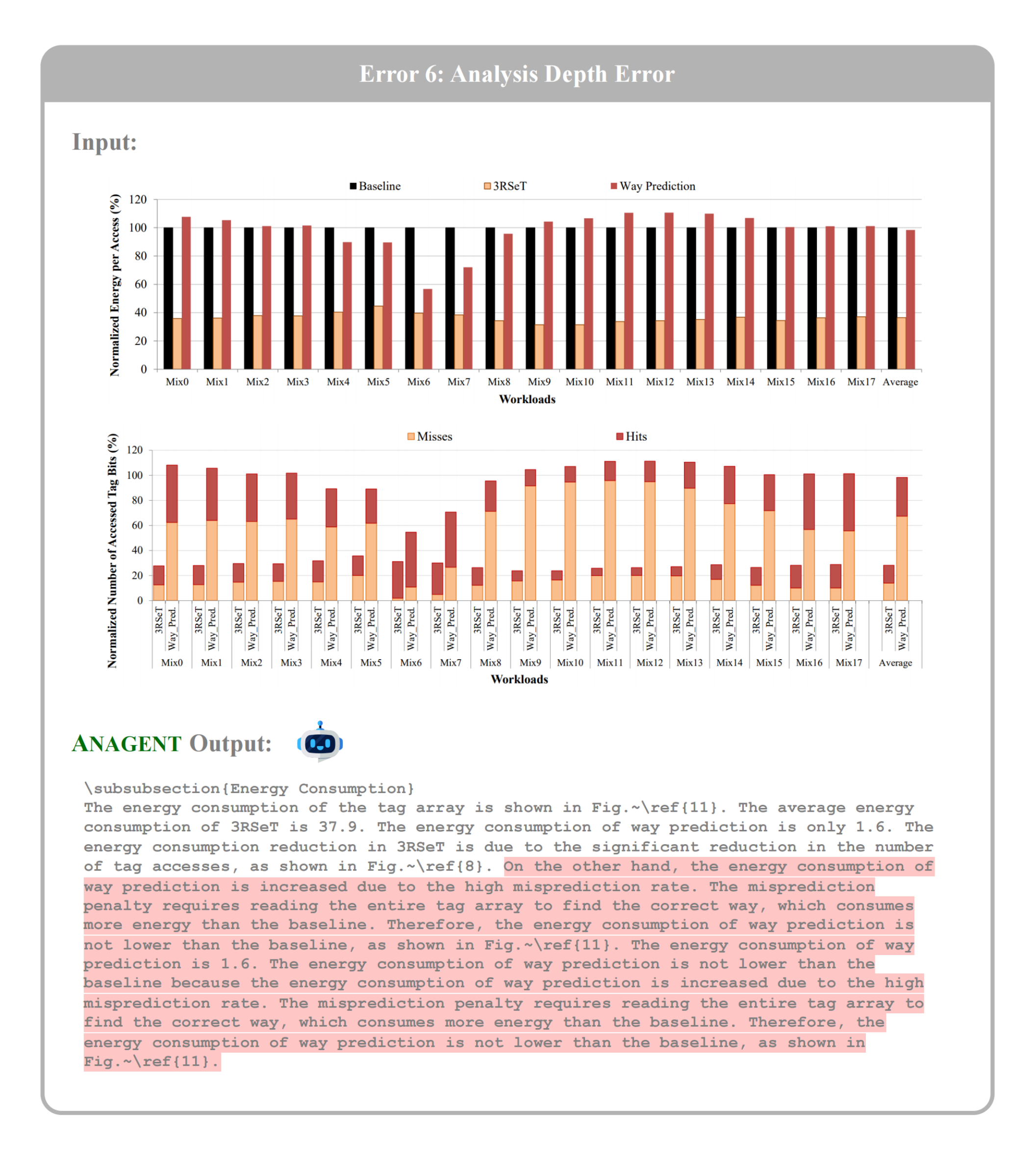}

\caption{\textbf{Example of Analysis Depth Error.} This example ($S_{\textsc{AVG}}=31.24\%$) illustrates a scientific analysis failure case due to \textit{analysis depth errors} (\S\ref{appendix:subsec:failure_analysis:error6}). Input query and agent prompts are provided in \S\ref{appendix:anagent:collaboration} and omitted here for clarity. Other intermediate outputs and contexts are also omitted for clear presentation.}
\label{fig:failure_analysis:error6_depth_error}

\vspace{-6pt}
\end{figure}


\begin{figure}[H]

\centering
\includegraphics[width=1.0\linewidth]{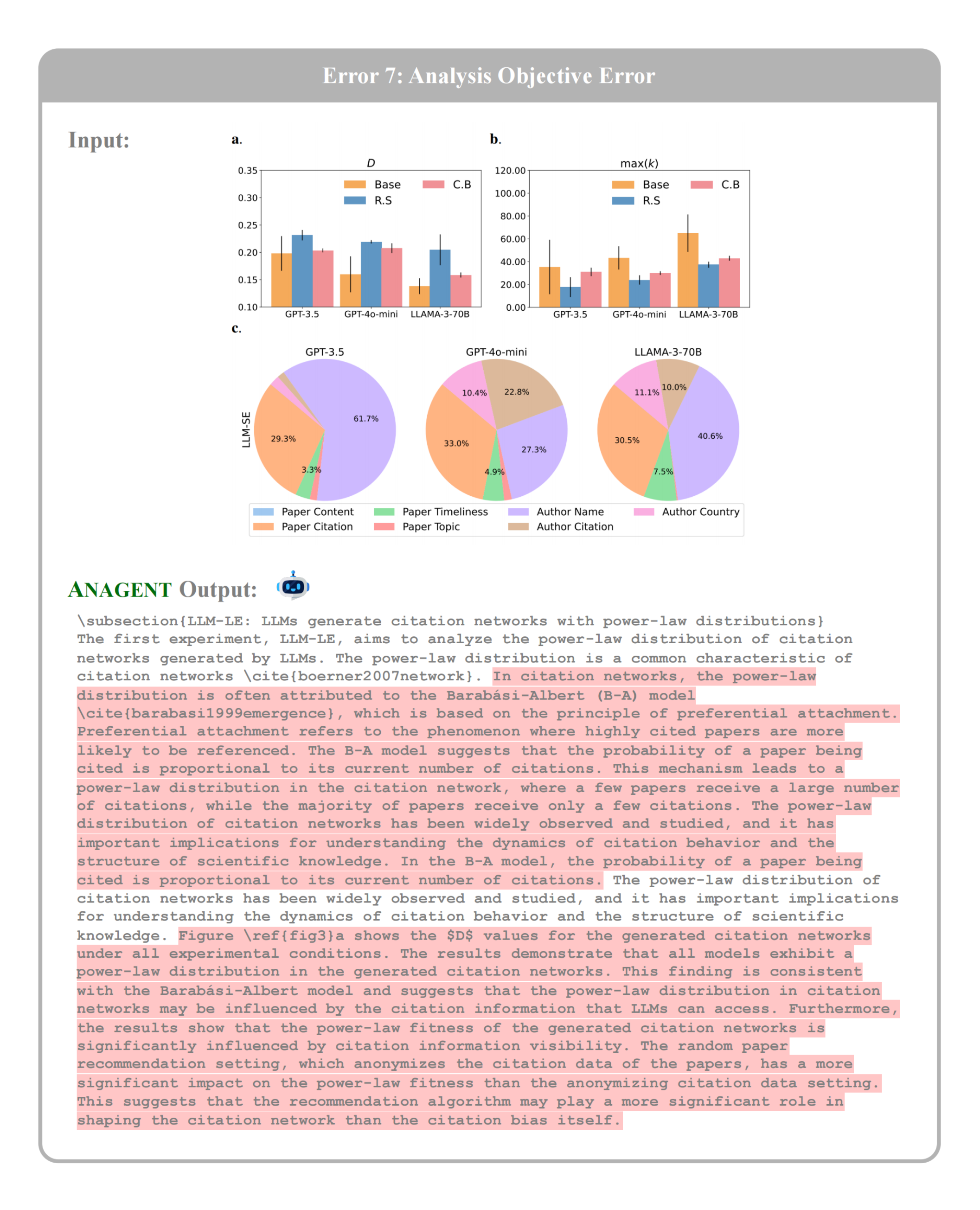}

\caption{\textbf{Example of Analysis Objective Error.} This example ($S_{\textsc{AVG}}=31.45\%$) illustrates a scientific analysis failure case due to \textit{analysis objective errors} (\S\ref{appendix:subsec:failure_analysis:error7}). Input query and agent prompts are provided in \S\ref{appendix:anagent:collaboration} and omitted here for clarity. Other intermediate outputs and contexts are also omitted for clear presentation.}
\label{fig:failure_analysis:error7_objective_error}

\vspace{-6pt}
\end{figure}


\begin{figure}[H]

\centering
\includegraphics[width=1.0\linewidth]{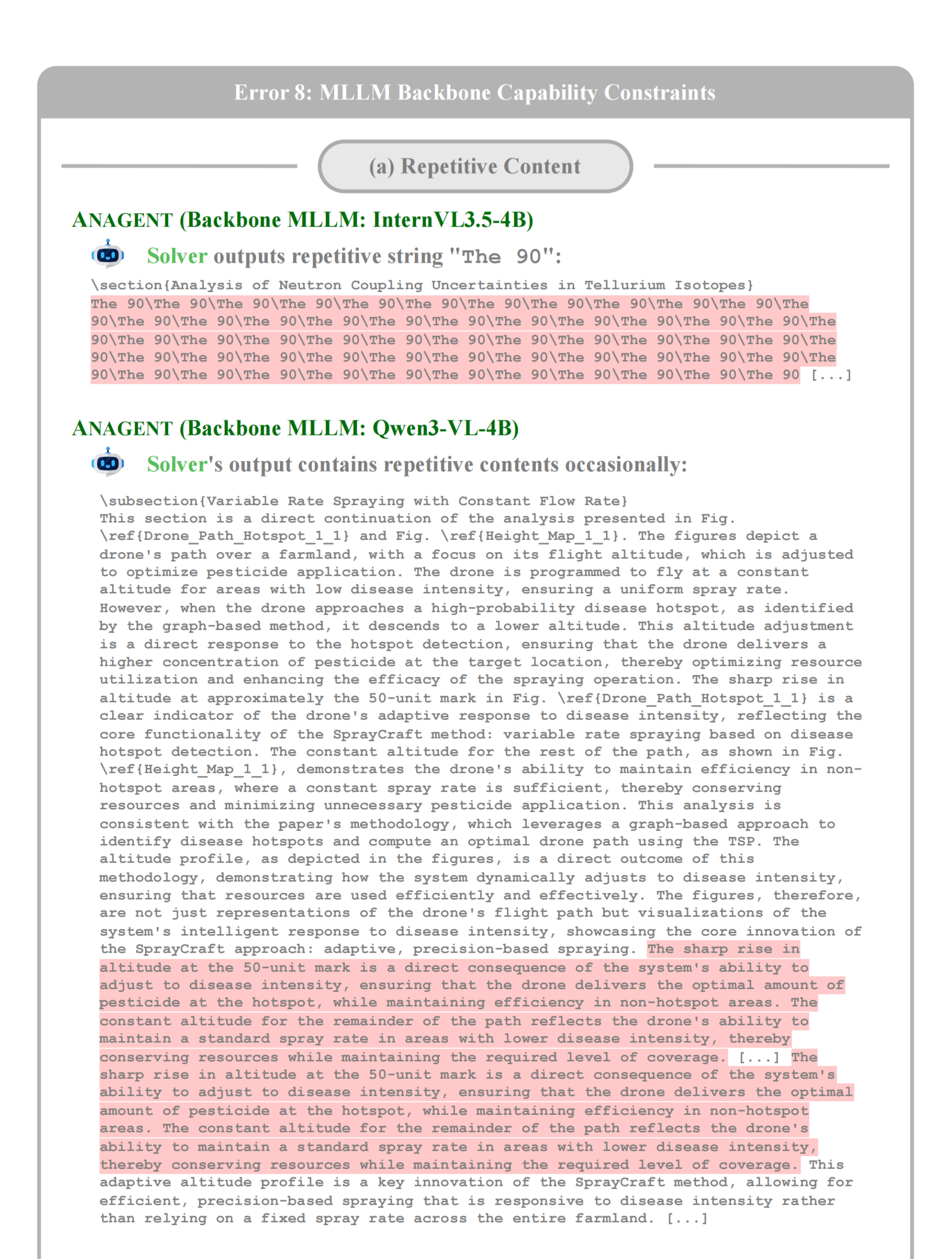}

\end{figure}

\begin{figure}[H]

\centering
\includegraphics[width=1.0\linewidth]{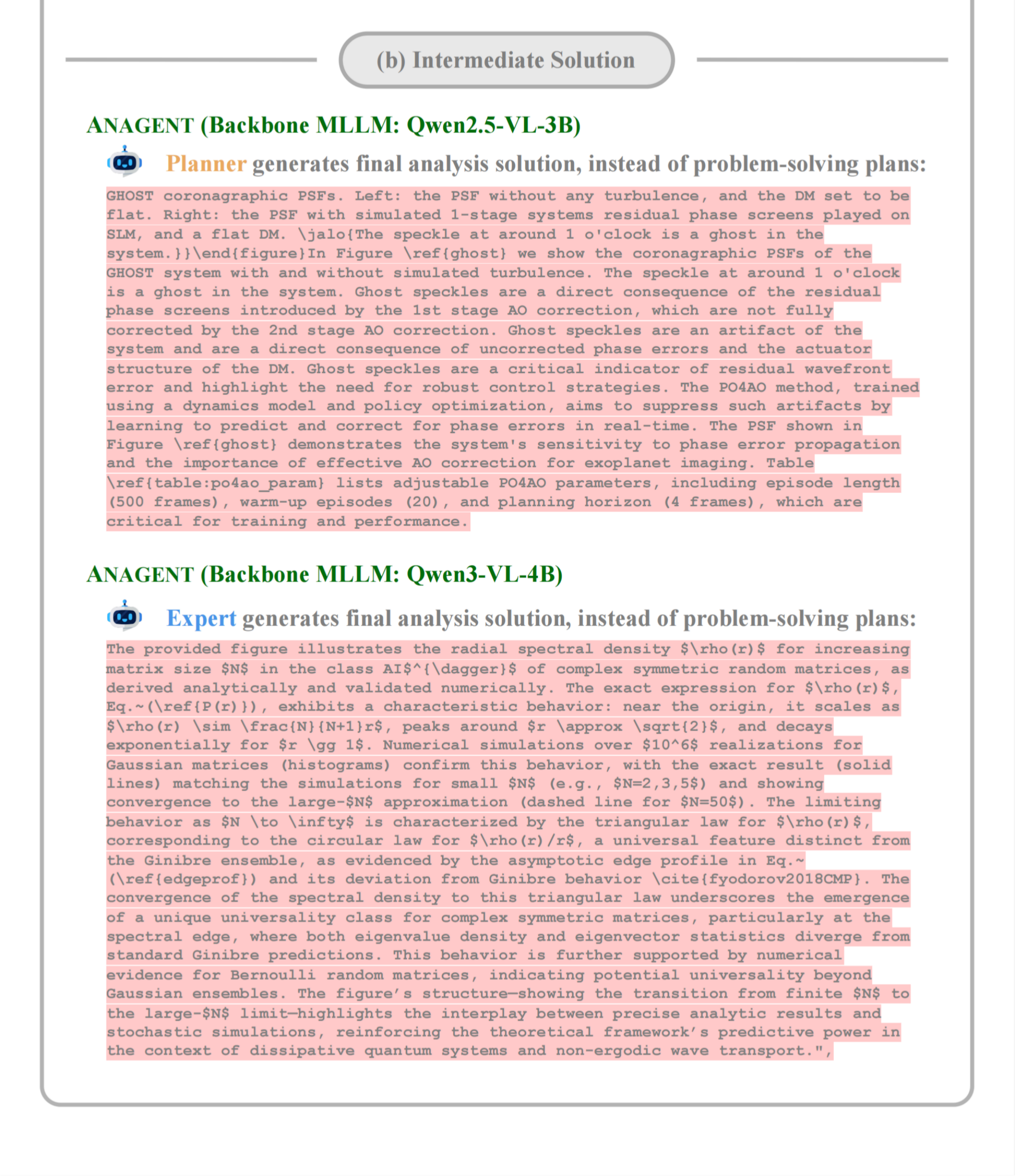}

\caption{\textbf{Example of Other Errors Due To MLLM Backbone Ability Constraints.} This figure shows two scientific analysis failure cases due to \textit{MLLM backbone ability constraints} (\S\ref{appendix:subsec:failure_analysis:error8}): (a) Repetitive Content, (b) Intermediate Solution. Input query and agent prompts are provided in \S\ref{appendix:anagent:collaboration} and omitted here for clarity. Some contents denoted as \texttt{[...]} are omitted for presentation brevity.}
\label{fig:failure_analysis:error8_other_errors}

\vspace{-6pt}
\end{figure}

%% file: figures/prompt_evaluation.tex
\begin{figure}[H]
\begin{promptbox}{Prompt for Five-Dimensional Evaluation Protocol}

You are an expert evaluator for scientific \code{\{data\_type\}} analysis writing tasks. Given a \code{\{data\_type\}} and its ground-truth analysis, evaluate whether a model-generated analysis is of high quality.

\vspace{6pt}

\textbf{Ground-truth Analysis}

\code{\{gt\_analysis\}}

\vspace{6pt}

\textbf{Model Analysis}

\code{\{model\_analysis\}}

\vspace{6pt}

\textbf{Evaluation Criteria}

1. \textbf{Content Accuracy}: Does the model analysis accurately and correctly reflect the explicitly presented information in the \code{\{data\_type\}}, such as reported values, trends, stated findings, and observed patterns, without distortion or omission?

2. \textbf{Analytical Completeness}: Does the model analysis sufficiently develop the derived analyses, interpretations, and conclusions implied by the \code{\{data\_type\}}, capturing key insights without missing important analytical findings?

3. \textbf{Format Correctness}: Are there any format errors in model analysis that violate the task requirements, such as incorrect structure, typographical errors, or improper formatting?

4. \textbf{Clarity \& Coherence}: Is the model analysis clearly articulated, properly structured, and logically coherent, with sound scientific writing throughout?

5. \textbf{Reliability \& Faithfulness}: Are there any hallucinated information in model analysis that is never covered, analyzed, and inferred in the ground-truth analysis? NOTE that both referred contents (e.g., referring to other tables/figures/sections/equations/etc., within the paper or referring to other papers) or reasonably inferred contents are considered as grounded (CAREFUL: analysis writing SHOULD have inferred contents, as long as they are reasonable and inferrable from the given information in such as way that are covered by ground-truth analysis)

\vspace{6pt}

\textbf{Instructions}

\begin{itemize}[nosep, topsep=0pt, partopsep=0pt, leftmargin=10pt]
\item Compare the \textbf{Model Analysis} against the \textbf{Ground-truth Analysis}
\item Focus on the quality of model analysis writing in the five dimensions of \textbf{Evaluation Criteria}
\item Consider semantic similarity and information coverage, NOT exact word matching
\item Rate the overall quality on a scale from 0-2 (where 2 is excellent)
\end{itemize}

\vspace{6pt}

\textbf{Response Format}

\begin{lstlisting}[basicstyle=\ttfamily\small, breaklines=true, columns=fullflexible]
<think>your evaluation reasoning</think>
<accuracy>grade 0-2 for **Content Accuracy**: 0 = mostly inaccurate; 1 = partially accurate; 2 = fully accurate</accuracy>
<completeness>grade 0-2 for **Analytical Completeness**: 0 = most key analyses missing; 1 = some key analyses missing; 2 = all key analyses present</completeness>
<format>grade 0-2 for **Format Correctness**: 0 = many format errors; 1 = some format errors; 2 = no format error</format>
<writing>grade 0-2 for **Clarity & Coherence**: 0 = poorly written; 1 = moderately clear; 2 = clear and coherent</writing>
<faithfulness>grade 0-2 for **Reliability & Faithfulness**: 0 = mostly hallucinated; 1 = partially hallucinated; 2 = fully reasonable</faithfulness>
\end{lstlisting}

\end{promptbox}

\caption{\textbf{Prompt For Five-Dimensional Evaluation Protocol.} During LLM-as-judge and human expert evaluation (\S\ref{subsec:anabench:evaluation_metrics}), the evaluator is prompted to assess the quality of each model-generated scientific analysis in five dimensions, including \textit{content accuracy}, \textit{analytical completeness}, \textit{format correctness}, \textit{clarity \& coherence}, and \textit{reliability \& faithfulness}.}
\label{fig:five_dimension_evaluation_prompt}

\end{figure}

%% file: figures/prompt_agent.tex

\begin{figure}[H]

\begin{agentbox}{Prompt for Planner Agent}

You are a Task Planning agent specialized in analyzing and task problems, decomposing tasks into concrete subtasks, and generating high-level plans that can guide resolving the given tasks.

\vspace{6pt}

\textbf{Task Problem}

\code{\{task\_problem\}}

\textbf{Additional Contexts}

\code{\{task\_context\}}

\vspace{6pt}

\textbf{What You Need To Do}

1. Understand the \textbf{Task Problem} and its requirements, and analyze the \textbf{analysis width, depth, and objective} of this task

2. Analyze what contextual information and domain-specific knowledge are needed to resolve this task

3. Decompose complex tasks into clear, actionable subtasks

4. Create a concise and instructive \textbf{Problem-Solving Plan} that can effectively guide resolving the \textbf{Task Problem}

\vspace{6pt}

\textbf{Requirements}

\begin{itemize}[nosep, topsep=0pt, partopsep=0pt, leftmargin=10pt]
\item Be systematic, thorough, and strategic in your planning
\item As contextual information and domain-specific knowledge are highly important, please ALWAYS highlight "\textit{Find, retrieve, distill, and summarize all the related contexts}" at the very beginning of your plan
\item Please specify finding and adding necessary citations and references to support the analysis completeness and format accuracy of the final answer
\item Your \textbf{Problem-Solving Plan} should be around \code{\{plan\_limit\}} characters, consisting several subtasks in bullet-point format using "*" WITHOUT indents
\item Provide both your reasoning and planning in the following format with proper enclosure:

\begin{lstlisting}[basicstyle=\ttfamily, breaklines=true, columns=fullflexible]
<think>your reasoning and analysis</think>
<plan>your **Problem-Solving Plan**</plan>
\end{lstlisting}

\end{itemize}

\end{agentbox}

\caption{\textbf{Prompt For \Planner Agent.} \Planner directs task-oriented planning, providing systematic guidance for improved analysis accuracy and completeness.}
\label{fig:planner_prompt}

\end{figure}


\begin{figure}[H]

\begin{agentbox}{Prompt for Expert Agent (PART I)}

You are an Expert Scientist equipped with various tools for information retrieval and knowledge collection.

Your job is to iteratively gather task-specific information by:

1. Analyze the given \textbf{Task Problem} and the corresponding \textbf{Problem-Solving Plan}

2. Decide which ONE tool to use in each turn

3. Receive and summarize tool execution results

4. Repeat until you have gathered sufficient information to solve the given task

You can ONLY use ONE tool per turn. Think carefully about which tool can provide the most useful information based on what you already know.
After gathering all the relevant contexts and domain knowledge by calling different tools, in the end, you should generate a concise task-specific \textbf{Knowledge Summary} with all the important information needed to answer the \textbf{Task Problem}.

\vspace{6pt}

\textbf{Task Problem}

\code{\{task\_problem\}}

\end{agentbox}

\end{figure}

\begin{figure}[H]

\begin{agentbox}{Prompt for Expert Agent (PART II)}

\textbf{Additional Contexts}

\code{\{task\_context\}}

\textbf{Problem-Solving Plan}

\code{\{planner\_plan\}}

\textbf{Available Tools}

\code{\{tool\_info\}}

\vspace{6pt}

\textbf{What You Need To Do}

You have \{max\_turns\} turns to gather information. This is TURN 1.
In each turn, you have two choices:

1. \textbf{Use A Tool} to gather more information:

\begin{itemize}[nosep, topsep=0pt, partopsep=0pt, leftmargin=20pt]
\item Analyze the task and current state to decide which tool to use. You can ONLY use ONE tool in each turn.
\item To call a tool, please follow these steps: (a) First, provide your analysis and reasoning, enclosed in: \code{<think>your analysis of what information is needed, which specific tool you choose for current turn, and why</think>}; (b) Second, specify your selected tool for this turn, enclosed in: \code{<tool>tool\_name</tool>}; (c) Third, provide your tool-specific query and options with proper enclosure as required in \textbf{Available Tools}.
\end{itemize}

2. \textbf{Stop and Summarize} if you believe your collected information is sufficient to resolve the \textbf{Task Problem}

\vspace{6pt}

\textbf{Instructions}

\begin{itemize}[nosep, topsep=0pt, partopsep=0pt, leftmargin=10pt]
\item Carefully review and analyze the \textbf{Task Problem} and its requirements, the \textbf{Problem-Solving Plan}, and all your collected information via tool calling
\item DO NOT ANSWER THE \textbf{Task Problem}: You ONLY need to analyze and summarize all your collected information based on your reasoning, analysis, observations, key findings, and any other highly important information that you believe can support solving the \textbf{Task Problem}
\item BE MINDFUL about the required depth of the \textbf{Task Problem}: Analyze carefully the \textbf{analysis width, depth, and objective} of this task, e.g., whether it asks about shallow description or in-depth analysis, academic writing or technical report, scientific discovery or creative generation, etc.
\item Your \textbf{Knowledge Summary} should: (1) Be both concise and informative, consisting of around \code{\{summary\_len\}} characters; (2) Use Markdown format but in free style that you believe best suits current task, e.g., plain text, bullet points, or a mix of both, etc.; (3) \textbf{Ensure The Accuracy and Quality of Your Summary:} Ensure the faithfulness of you \textbf{Knowledge Summary}, incorporating ONLY factual information of significance directly from the collected contexts and domain knowledge, avoiding self-generated, self-interpreted, self-inferred, or rephrased details; (4) Always use the correct format in line with the input table/figure. For example, if the input uses LaTeX, you SHOULD use Latex Bib and \code{\textbackslash cite} for reference citations, and specify Latex \code{\textbackslash ref} when referring to other tables/figures/sections/equations/etc.; (5) Organize a well-structured summary that includes both high-level insights and detailed information, analysis, observations, and key findings, etc. For example: \code{<think>your reasoning</think><summary>\{summary\_structure\}</summary>}
\end{itemize}

Please generate your reasoning and \textbf{Knowledge Summary} enclosed in:
\begin{lstlisting}[basicstyle=\ttfamily, breaklines=true, columns=fullflexible]
<think>your reasoning</think>
<summary>your knowledge summary</summary>
\end{lstlisting}

\end{agentbox}

\caption{\textbf{Prompt For \Expert Agent.} \Expert leads information searching and retrieval to supply task-specific contextual and domain knowledge.}
\label{fig:expert_prompt}

\end{figure}


\begin{figure}[H]

\begin{agentbox}{Prompt for Solver Agent (Initial Generation)}

You are a Expert Scientist that writes high-quality scientific analysis according to the given \textbf{Task Problem}.

\vspace{6pt}

\textbf{Task Problem}

\code{\{task\_problem\}}

\textbf{Additional Contexts}

\code{\{task\_context\}    \# Expert knowledge summary included}

\textbf{Problem-Solving Plan}

\code{\{planner\_plan\}}

\vspace{6pt}

\textbf{What You Need To Do}

1. Analyze the \textbf{Task Problem} and its requirements, and conclude the \textbf{analysis width, depth, and objective} of this task

2. Review the \textbf{Problem-Solving Plan} and \textbf{Additional Contexts}

3. Reason and analyze all the available information and knowledge systematically and thoroughly

4. Generate your solution enclosed in: \code{<answer>your solution</answer>}

\vspace{6pt}

\textbf{Requirements}

\begin{itemize}[nosep, topsep=0pt, partopsep=0pt, leftmargin=10pt]
\item Please ensure the accuracy, precision, completeness, profession, and quality of your scientific analysis
\item Please carefully review all the given information and contexts, and generate semantically coherent, logically structured, and scientifically grounded analysis
\item Always use the correct format in line with the input table/figure. For example, if the input uses LaTeX, you SHOULD use Latex Bib and \code{\textbackslash cite} for reference citations, and specify Latex \code{\textbackslash ref} when referring to other tables/figures/sections/equations/etc.
\item Provide your reasoning and solution in the following format with proper enclosure:
\begin{lstlisting}[basicstyle=\ttfamily, breaklines=true, columns=fullflexible]
<think>your reasoning</think>
<answer>your scientific analysis</answer>
\end{lstlisting}
\end{itemize}

\end{agentbox}

\caption{\textbf{Prompt for the \Solver Agent.} As supported by \Planner and \Expert, \Solver integrates important contextual information and domain knowledge into task inputs to generate initial scientific analysis, and collaborates with the \Critic to iteratively reflect on, correct, and refine the analysis. \Critic feedback is omitted here for prompt presentation clarity, and is added to the prompt during iterative reflection and refinement.}
\label{fig:solver_prompt}

\end{figure}


\begin{figure}[H]

\begin{agentbox}{Prompt for Critic Agent}

You are a CRITIC ADVISOR responsible for evaluating and improving STUDENT's \textbf{STUDENT Answer} to the given \textbf{Task Problem} based on available \textbf{Additional Contexts}.

\vspace{3pt}

\textbf{Task Problem}

\code{\{task\_problem\}}

\textbf{Additional Contexts}

\code{\{task\_context\}}

\textbf{STUDENT Problem-Solving Plan}

\code{\{planner\_plan\}}

\textbf{STUDENT Answer}

\code{\{solver\_solution\}}

\vspace{3pt}

\textbf{Evaluation Criteria}

1. \textbf{Content Accuracy}: Does the model analysis accurately and correctly reflect the explicitly presented information in the \code{\{data\_type\}}, such as reported values, trends, stated findings, and observed patterns, without distortion or omission?

2. \textbf{Analytical Completeness}: Does the model analysis sufficiently develop the derived analyses, interpretations, and conclusions implied by the \code{\{data\_type\}}, capturing key insights without missing important analytical findings?

3. \textbf{Format Correctness}: Are there any format errors in model analysis that violate the task requirements, such as incorrect structure, typographical errors, or improper formatting?

4. \textbf{Clarity \& Coherence}: Is the model analysis clearly articulated, properly structured, and logically coherent, with sound scientific writing throughout?

5. \textbf{Reliability \& Faithfulness}: Are there any hallucinated information in model analysis that is never covered, analyzed, and inferred in the ground-truth analysis? NOTE that both referred contents (e.g., referring to other tables/figures/sections/equations/etc., within the paper or referring to other papers) or reasonably inferred contents are considered as grounded (CAREFUL: analysis writing SHOULD have inferred contents, as long as they are reasonable and inferrable from the given information in such as way that are covered by ground-truth analysis)

\vspace{3pt}

\textbf{What You Need To Do}

1. Analyze the \textbf{Task Problem} and its requirements, and conclude the \textbf{analysis width, depth, and objective} of this task

2. Review the \textbf{Additional Contexts}, reasoning and analyzing all the available information and knowledge systematically and thoroughly

3. Evaluate STUDENT's \textbf{STUDENT Answer} according to the five \textbf{Evaluation Criteria}

4. Generate your detailed improvement guidance feedback to STUDENT, which should at least cover the five evaluation dimensions above, and less than \code{\{feedback\_limit\}} characters

\vspace{3pt}

\textbf{Response Format}

\begin{lstlisting}[basicstyle=\ttfamily\small, breaklines=true, columns=fullflexible, belowskip=0pt]
<think>your evaluation reasoning</think>
<accuracy>grade 0-2 for **Content Accuracy**: 0 = mostly inaccurate; 1 = partially accurate; 2 = fully accurate</accuracy>
<completeness>grade 0-2 for **Analytical Completeness**: 0 = most key analyses missing; 1 = some key analyses missing; 2 = all key analyses present</completeness>
<format>grade 0-2 for **Format Correctness**: 0 = many format errors; 1 = some format errors; 2 = no format error</format>
<writing>grade 0-2 for **Clarity & Coherence**: 0 = poorly written; 1 = moderately clear; 2 = clear and coherent</writing>
<faithfulness>grade 0-2 for **Reliability & Faithfulness**: 0 = mostly hallucinated; 1 = partially hallucinated; 2 = fully reasonable</faithfulness>
<feedback>your feedback</feedback>
\end{lstlisting}

\end{agentbox}

\caption{\textbf{Prompt For \Critic Agent.} Cooperating with \Solver, \Critic also employs the five-dimensional evaluation protocol (Fig.~\ref{fig:five_dimension_evaluation_prompt} \& \S\ref{appendix:preliminary:method}) for reflective correction and refinement.}
\label{fig:critic_prompt}

\end{figure}